\RequirePackage{fix-cm}

\RequirePackage{snapshot}

\documentclass[twocolumn]{svjour3}          
\smartqed  
\usepackage{graphicx}
\usepackage{amsmath}
\usepackage{amssymb}
\newcommand{\minisection}[1]{\vspace{0.04in} \noindent {\bf #1}\ \ }
\usepackage{subcaption}
\usepackage{multirow}
\usepackage{multicol}
\usepackage[table]{xcolor}

\usepackage{soul, xcolor}
\setstcolor{red}

\usepackage{array}
\makeatletter
\newcommand*{\@rowstyle}{}
\newcommand*{\rowstyle}[1]{
  \gdef\@rowstyle{#1}%
  \@rowstyle\ignorespaces%
}

\newcolumntype{=}{
  >{\gdef\@rowstyle{}}%
}

\newcolumntype{+}{
  >{\@rowstyle}%
}

\makeatother

\usepackage[round]{natbib}
\usepackage[final, colorlinks=true,allcolors=blue,breaklinks=true]{hyperref} 

\newcommand\crule[3][black]{\textcolor{#1}{\rule{#2}{#3}}}

\definecolor{f_trail}{RGB}{170,170,170}
\definecolor{f_grass}{RGB}{0,255,0}
\definecolor{f_vegetation}{RGB}{102,102,51}
\definecolor{f_sky}{RGB}{0,120,255}
\definecolor{f_obstacle}{RGB}{0,0,0}
\definecolor{u_Bed}{RGB}{0,0,255}
\definecolor{u_Books}{RGB}{233, 89, 48}
\definecolor{u_Ceiling}{RGB}{0, 218, 0}
\definecolor{u_Chair}{RGB}{149, 0, 240}
\definecolor{u_Floor}{RGB}{222, 241, 24}
\definecolor{u_Furniture}{RGB}{255, 206, 206}
\definecolor{u_object}{RGB}{0, 224, 229}
\definecolor{u_Picture}{RGB}{106, 136, 204}
\definecolor{u_Sofa}{RGB}{117, 29, 41}
\definecolor{u_Table}{RGB}{240, 35, 235}
\definecolor{u_Tv}{RGB}{0, 167, 156}
\definecolor{u_wall}{RGB}{250, 139, 0}
\definecolor{u_Window}{RGB}{225, 229, 195}

\begin{document}
\sloppy
\title{Mix and match networks: cross-modal alignment for  zero-pair image-to-image translation}

\author{Yaxing Wang         \and
        Luis Herranz \and  Joost van de Weijer
}


\institute{Yaxing Wang, Luis Herranz, Joost van de Weijer  \at
                the Computer Vision Center Barcelona, Edifici O, Campus UAB, 08193, Bellaterra, Spain. \\
              \email{\{yaxing, lherranz, joost\}@cvc.uab.es} }

\date{Received: date / Accepted: date}

\maketitle

\begin{abstract}
This paper addresses the problem of inferring unseen cross-modal image-to-image translations between multiple modalities. We assume that only some of the pairwise translations have been seen (i.e. trained) and infer the remaining unseen translations (where training pairs are not available). We propose mix and match networks, an approach where multiple encoders and decoders are aligned in such a way that the desired translation can be obtained by simply cascading the source encoder and the target decoder, even when they have not interacted during the training stage (i.e. unseen). The main challenge lies in the alignment of the latent representations at the bottlenecks of encoder-decoder pairs. We propose an architecture with several tools to encourage alignment, including autoencoders and robust side information and latent consistency losses.
We show the benefits of our approach in terms of effectiveness and scalability compared with other pairwise image-to-image translation approaches.
We also propose zero-pair cross-modal image translation, a challenging setting where the objective is inferring semantic segmentation from depth (and vice-versa) without explicit segmentation-depth pairs, and only from two (disjoint) segmentation-RGB and depth-RGB training sets. We observe that a certain part of the shared information between unseen modalities might not be reachable, so we further propose a variant that leverages pseudo-pairs which allows us to exploit this shared information between the unseen modalities.

\end{abstract}

\section{Introduction}
\label{intro}
For many computer vision applications, the task is to estimate a mapping between an input image and an output image. This family of methods is often known as image-to-image translations (image translations hereinafter). They include transformations between different modalities, such as from RGB to depth~\citep{liu2016learning}, or domains, such as luminance to color images~\citep{zhang2016colorful}, or editing operations such as artistic style changes~\citep{gatys2016image}. These mappings can also include other 2D representations such as semantic segmentations~\citep{long2015fully} or surface normals~\citep{eigen2015predicting}. One drawback of the initial research on image translations is that the methods required paired data to train the mapping between the domains~\citep{long2015fully,eigen2015predicting,isola2016image}. Another class of algorithms, based on cycle consistency, address the problem of mapping between unpaired domains~\citep{kim2017learning,yi2017dualgan,zhu2017unpaired}. These methods are based on the observation that translating from one domain to another and translating back to the original domain should result in recovering the original input image.

\begin{figure}
    \centering   
   \includegraphics[width=0.8\columnwidth]{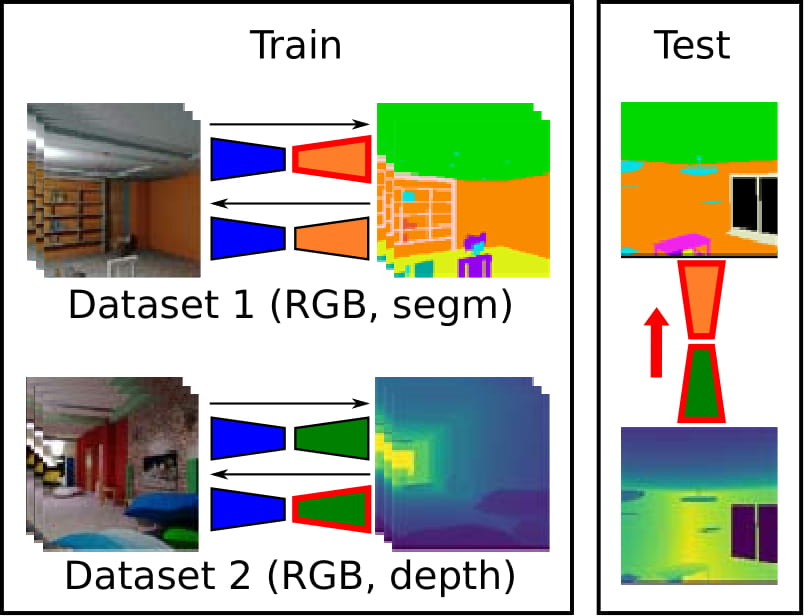}
   \caption{Overview of mix and match networks (M\&MNets) and zero-pair translation. Two disjoint datasets are used to train seen translations between RGB and segmentation and between RGB and depth (and vice versa). We want to infer the unseen depth-to-segmentation translation (i.e. \textit{Zero-pair translation}). The M\&MNets approach builds the unseen translator by simply cascading the source encoder and target decoder (i.e. depth and segmentation, respectively). Best viewed in color.}\label{fig:introduction}
\end{figure}

The discussed approaches consider translations between two domains which are either paired or unpaired. However, for many real-world applications there exist both paired and unpaired domains simultaneously. 
Consider the case of image translation between multiple modalities, where for some of them we have access to aligned data pairs but not for all modalities. The aim would then be to exploit the knowledge from the paired modalities to obtain an improved mapping for the unpaired modalities. An example of such a translation setting is the following: you have access to a set of RGB images and their semantic segmentation, and a (different) set of RGB images and their corresponding depth maps, but you are interested in obtaining a mapping from depth to semantic segmentation (see Figure~\ref{fig:introduction}). We call this the \textit{unseen} translation because we do not have pairs for this translation, and we refer to this setting as \textit{zero-pair translation}. Zero-pair translation is typically desired when we extend an experimental setup with an additional camera in another modality. We now would like to immediately exploit this new sensor without the cost of labelling new data. In this paper, we provide a new approach to address the zero-pair translation problem.

We propose a new method, which we call \textit{mix and match networks}, which addresses the problem of learning a mapping between unpaired modalities by seeking alignment between encoders and decoders via their latent spaces\footnote{The code is available online at http://github.com/yaxingwang/Mix-and-match-networks.}. The translation between unseen modalities is performed by simply concatenating the source modality encoder and the target modality decoder (see Figure~\ref{fig:introduction}). The success of the method depends on the alignment of the encoder and decoder for the unseen translation. We study several techniques that contribute to achieve alignment, including the usage of autoencoders, latent space consistency losses and the usage of robust side information to guide the reconstruction of spatial structure. 

We evaluate our approach in a challenging cross-modal task, where we perform zero-pair depth to semantic segmentation translation (or semantic segmentation to depth  translation), using only RGB-depth and RGB-semantic segmentation pairs during training. Furthermore, we show that the results can be further improved by using pseudo-pairs between the unseen modalities that allow the network to exploit unseen shared information. We also show that our approach can  be used for cross-modal translation and with unpaired data. In particular, we show that mix and match networks scale better with the number of modalities, since they are not required to learn all pairwise image translation networks (i.e. they scale linearly instead of quadratically).

This article is an extended version of a previous conference publication~\citep{wang2018mix}. We have included more analysis and insight about how mix and match networks exploit the information shared between modalities, and propose an improved mix and match networks framework with pseudo-pairs which allows us to access previously unexploited shared information between unseen  modalities (see Section~\ref{sec:pseudo-pairs}). This was found to significantly improve performance. In addition, \cite{wang2018mix} only report results on a synthetic dataset. Here we also provide results on real images (SUN RGB-D dataset~\citep{song2015sun}) and four modalities (Freiburg Forest dataset~\citep{valada2016deep}). Furthermore, we have added more insights on how the alignments between encoders and decoders evolve during training.

\section{Related work}
\label{sec:1}
In this section we discuss the literature of related research areas. 

\begin{figure*}[t]
      \centering
      \begin{subfigure}[b]{0.2\textwidth}
              \centering
              \includegraphics[scale=0.12]{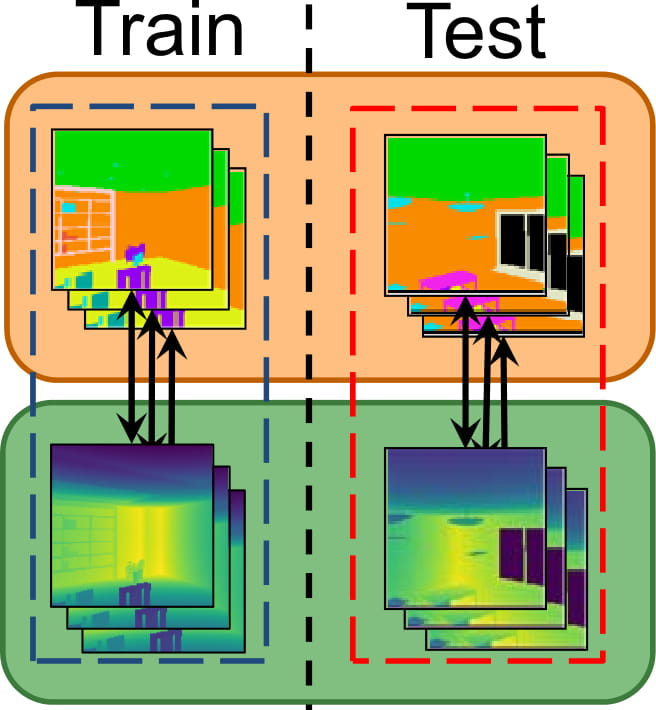}
              \caption{Paired translation}
      \end{subfigure}
      \!
      \begin{subfigure}[b]{0.2\textwidth}
              \centering
              \includegraphics[scale=0.12]{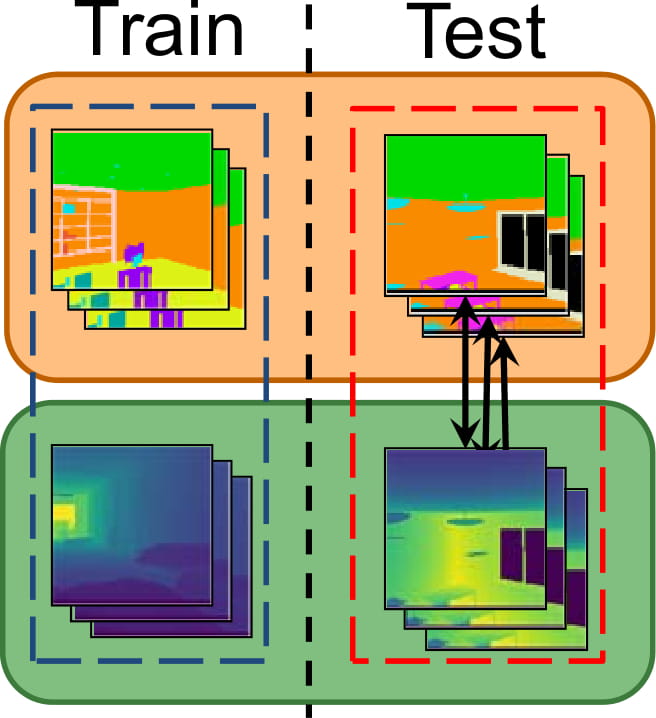}
              \caption{Unpaired translation}
      \end{subfigure}
      \!
      \begin{subfigure}[b]{0.28\textwidth}
              \centering
              \includegraphics[scale=0.12]{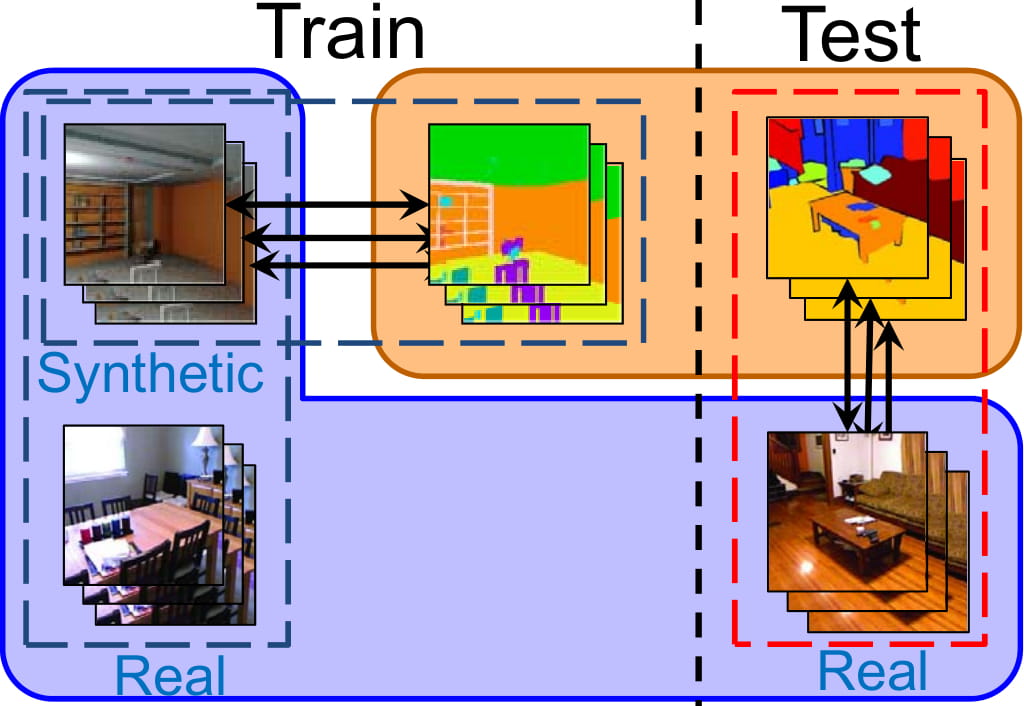}
              \caption{Unsupervised domain adapt.}
      \end{subfigure}
      \!
      \begin{subfigure}[b]{0.28\textwidth}
              \centering
              \includegraphics[scale=0.12]{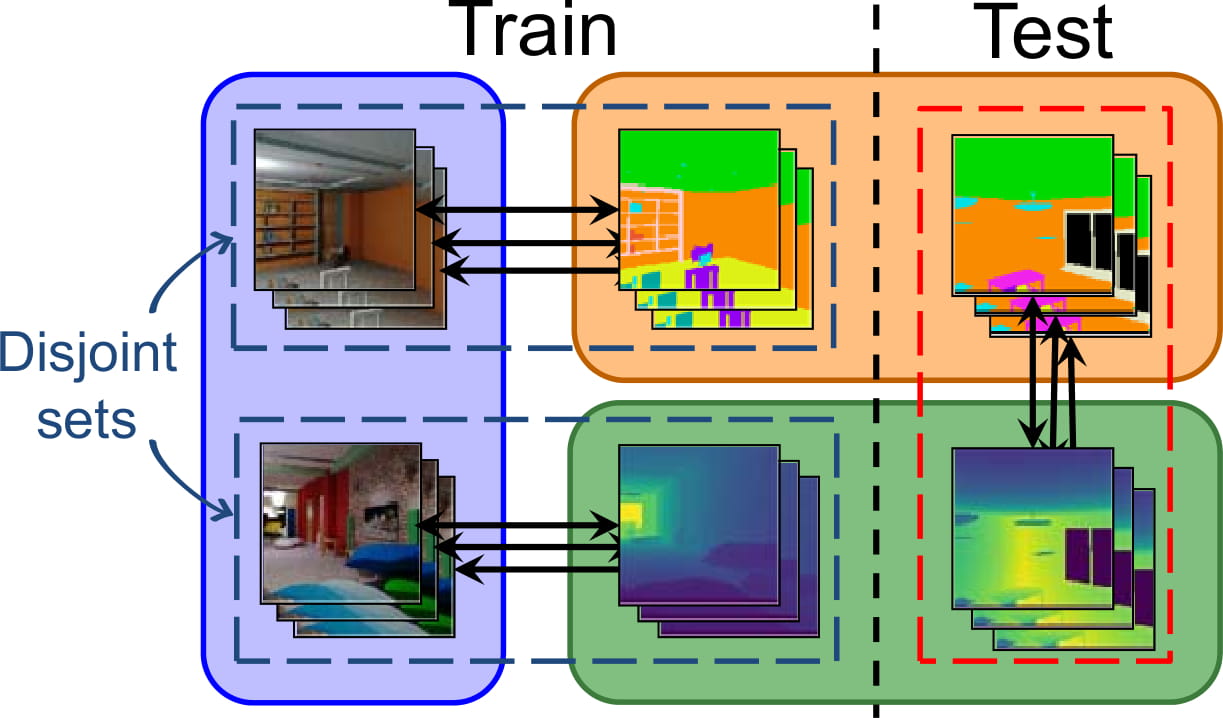}
              \caption{Zero-pair translation}
      \end{subfigure}
        \caption{\label{fig:cross-modal-settings} cross-modal translation train and test settings: (a) paired translation, (b) unpaired translation, (c) unsupervised domain adaptation for segmentation (two modalities and two domains in the RGB modality), (d) zero-paired translation (three modalities). Best viewed in color.}
\end{figure*}

\subsection{Image-to-image translation}
\minisection{Paired translations}
Generic encoder-decoder architectures have achieved impressive results in a wide range of transformations between images. \cite{isola2016image} proposed \textit{pix2pix}, which is a conditional generative adversarial network (conditional GAN)~\citep{goodfellow2014generative,mirza2014conditional} trained with pairs of input and output images to learn a variety of image translations. Those translations include cross-domain image translations such as colorization and style transfer. \cite{gonzalez2018image} disentangle the information of the domains in the latent space, which allows to do cross-domain retrieval as well as perform one-to-many translations. The ability of GANs to generate realistic images also enables \textit{pix2pix} to address effectively challenging cross-modal translations, such as semantic segmentation to RGB image. In this case, recent multi-scale architectures~\citep{chen2017photographic,wang2018high} achieve better results in higher resolution images.

\minisection{Unpaired translations}
Various works extended image translation to the case where no explicit input-output image pairs are available (\textit{unpaired} image translation), using the idea of cyclic consistency~\citep{kim2017learning,yi2017dualgan,zhu2017unpaired,lin2018conditional} or consistency between certain extracted features~\citep{taigman2016unsupervised}. To avoid accidental artifacts and improve learning, \cite{mejjati2018unsupervised} integrate an attention mechanism to help translations focus on semantically meaningful regions. \cite{liu2017unsupervised} show that unsupervised mappings can be learned by imposing a joint latent space between the encoder and the decoder. Both TransGaGa~\citep{Wu_2019_CVPR} and TraVeLGAN~\citep{Amodio_2019_CVPR} address the issues of image translation across large geometry variations. The former disentangles image space in a Cartesian product of the appearance and the geometry latent spaces, and the latter considers a Siamese network to replace the cycle-consistency constraint. 

In this work, we consider the case where paired data is available between some  modalities and not available between others (i.e. zero-pair), and how the knowledge can be transferred to those unseen translations. Whereas previous work has focused on unpaired domains of the same modality, we show results for unpaired domains of different modalities. 

\minisection{Diversity in translations}
Given an input image (e.g. an edge image or a grayscale image) there are often multiple possible solutions (e.g. different plausible colorizations). The paired translation framework was extended to one-to-many translations in the work of ~\cite{zhu2017toward}. DRIT~\citep{lee2018diverse},  MUNIT~\citep{huang2018multimodal} and Augmented CycleGAN~\citep{almahairi2018augmented} can learn one-to-many translations in unpaired settings. In general, disentangled representations allow achieving diversity by keeping the content component and sampling the style component of the latent representation~\citep{mathieu2016disentangling,gonzalez2018image,lee2018diverse}. \cite{Cho_2019_CVPR} propose a novel group-wise deep whitening-and-coloring method to improve computational efficiency. \cite{alharbi2019latent} scale the latent filter to avoid a complicated network framework to perform one-to-many translations. 

\minisection{Multi-domain translations}
We also consider the case of multiple domains (and modalities). In concurrent work, \cite{choi2017stargan} also address scaling to multiple domains by using a single encoder-decoder model, which was previously explored by ~\cite{perarnau2016invertible}. \cite{chen2019homomorphic} effectively disentangle the intermediate states between source and target domains.  \cite{wang2019sdit} perform diverse and scalable image transfer by a single model. These works focus on faces and changing relatively superficial and localized attributes such as make-up, hair color, gender, etc., always within the RGB modality. In contrast, our approach uses multiple cross-aligned modality-specific encoders and decoders, which are necessary to address the deeper structural changes required by our cross-modal setting. \cite{Anoosheh_2018} also use multiple encoders-decoders but focus on the easier cross-domain task of style transfer.

\subsection{Semantic segmentation and depth estimation}
Semantic image segmentation aims at assigning each pixel to an object class. \cite{long2015fully} propose fully convolutional networks (FCN), following an encoder-decoder structure.  
Since the FCN shows outstanding performance, this paradigm has been adopted in many current methods for semantic segmentation~\citep{badrinarayanan2015segnet,ronneberger2015u,yu2015multi,chen2018deeplab,zhao2017pyramid}. 
Of particular interest is   SegNet~\citep{badrinarayanan2015segnet}, which we adapt in our method. SegNet introduces the use of pooling indices instead of copying encoder features (i.e. skip connections, as in U-Net~\citep{ronneberger2015u}). We also consider pooling indices in our architecture for zero-pair image translation because we found them to be more robust and invariant under unseen translations.
%

Depth estimation aims at estimating the depth structure of an RGB image, usually represented as a depth map encoding the distance of each pixel to the camera. Most depth estimation methods are formalized as regression problems, where the aim is to minimize the mean squared error (MSE) with respect to a ground truth depth map. 
In general, an encoder-decoder architecture is used, often incorporating multiscale networks and skip connections~\citep{liu2016learning,wang2015towards,roy2016monocular,eigen2015predicting,kim2016unified,kuznietsov2017semi,laina2016deeper}. 

\minisection{Multi-modal encoder-decoders} 
With the development of multi-sensor cameras and datasets~\citep{lai2011large,silberman2012indoor,song2015sun}, encoder-decoder architectures have been adapted to multi-modal inputs~\citep{ngiam2011multimodal}, where different modalities (e.g. RGB, depth, infrared, surface normals) are encoded and combined prior to the decoding. The network is trained to perform tasks such as multi-modal object recognition~\citep{eitel2015multimodal,cheng2016semi,song2015sun}, scene recognition~\citep{song2017depth,song2015sun}, object detection~\citep{Gupta_2016_CVPR} (with simple classifiers or regressors as decoders in these cases) and semantic segmentation~\citep{silberman2012indoor,kendall2017multi,wang2018depth}. Similarly, multi-task learning can be applied to reconstruct multiple modalities~\citep{eigen2015predicting,kendall2017multi}. For instance ~\cite{eigen2015predicting} estimate depth, surface normals and semantic segmentation from a single RGB image, which can be seen as cross-modal translation.

Training a multi-task multimodal encoder-decoder network was recently studied by~\cite{Kuga_2017_ICCV}. They use a joint latent representation space for the various modalities. In our work we consider the alignment and transferability of pairwise image translations to unseen translations, rather than joint encoder-decoder architectures. Another multimodal encoder-decoder network was studied by~\citet{cadena2016multi}. They show that multi-modal autoencoders can address the depth estimation and semantic segmentation tasks simultaneously, even in the absence of some of the input modalities. All these works do not consider the zero-pair image translation problem addressed in this paper. 

\subsection{Zero-shot recognition} In conventional supervised image recognition, the objective is to predict the class label that is provided during training. However, this poses limitations in scalability to new classes, since new training data and annotations are required. In zero-shot learning~\citep{lampert2014attribute,fu2017recent,xian2018zero,xian2018feature,akata2016label}, the objective is to predict an unknown class for which there is no image available, but a description of the class (i.e. \textit{class prototype}) or any other source of semantic similarity with seen classes. This description can be a set of attributes (e.g. has wings, blue, four legs, indoor)~\citep{lampert2014attribute,jayaraman2014zero}, concept ontologies~\citep{fergus2010semantic,rohrbach2011evaluating} or textual descriptions~\citep{reed2016learning}. In general, an intermediate semantic space is leveraged as a bridge between the visual features from seen classes and class description from unseen ones.
In contrast to zero-shot recognition, we focus on unseen translations (unseen input-output pairs rather than simply unseen class labels).

\subsection{Zero-pair language translation} Evaluating models on unseen language pairs has been studied recently in machine translation~\citep{johnson2016google,chen2017teacher,zheng2017maximum,firat2016multi}. \citet{johnson2016google} proposed a neural language model that can translate between multiple languages, even pairs of language where no explicit paired sentences where provided\footnote{Note that~\citet{johnson2016google} refers to this as \textit{zero-shot} translation. In this paper we refer to this setting as zero-pair to emphasize that what is unseen is paired data and avoid ambiguities with traditional zero-shot recognition which typically refers to unseen samples.}. In their method, the encoder, decoder and attention are shared. In our method we focus on images, which are essentially a radically different type of data, with two dimensional structure in contrast to the sequential structure of language.

\subsection{Domain adaptation}
A related line of research is unsupervised domain adaptation. In that case the task is to transfer knowledge from a supervised source domain to an unsupervised target domain (see Figure~\ref{fig:cross-modal-settings}c). This problem has been addressed by finding domain invariant feature spaces~\citep{gong2012geodesic,ganin2015unsupervised,tsai2018learning}, using image translation models to map between source and target domain~\citep{wu2018dcan}, and exploiting pseudo-labels~\citep{saito2017asymmetric,zou2018domain}. Knowledge can also be transferred across modalities~\citep{Gupta_2016_CVPR,castrejon2016learning,hoffman2016cross,hoffman2016learning}. For instance, \citet{Gupta_2016_CVPR} use cross-modal distillation to learn depth models for classification by distilling RGB features (from pretrained model trained on a much larger RGB dataset), through a large set of unlabeled RGB-D pairs. Modality adaptation can also be achieved using cross-modal translation\citep{xu2017learning,zhang2019synthetic}.

When comparing this line of research with the setting we consider in this paper (i.e. zero-pair translation) there are some important differences.
The unsupervised domain adaptation setting (see Figure~\ref{fig:cross-modal-settings}c) typically involves two modalities (e.g. RGB and segmentation), and two domains within the RGB modality (e.g. synthetic and real). Paired data is available only for the synthetic-segmentation while the synthetic-real translation is unpaired, and the unseen translation is real-segmentation (with test paired data). In contrast, our setting (see Figure~\ref{fig:cross-modal-settings}d) is more challenging involving three modalities, with one disjoint paired training set for each seen translation. In comparison, using paired data to tackle domain shift allows us to reach much larger and challenging domain shifts and even modality shifts, a setting which, to the best of our knowledge, is not considered in the domain adaptation literature.

\begin{figure}
    \centering   
   \begin{subfigure}[b]{0.35\columnwidth}
          \centering
          \includegraphics[scale=0.18]{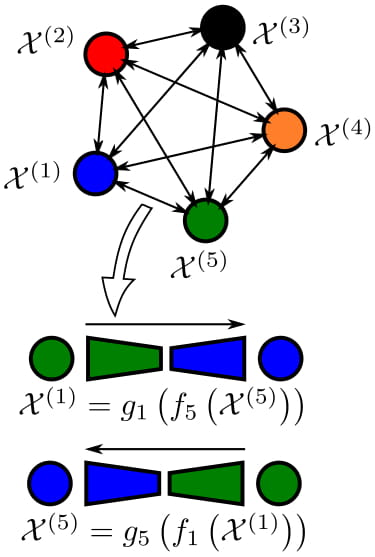}
          \caption{All seen}\label{fig:overview_all_seen}
      \end{subfigure}
      \begin{subfigure}[b]{0.63\columnwidth}
          \centering
          \includegraphics[scale=0.18]{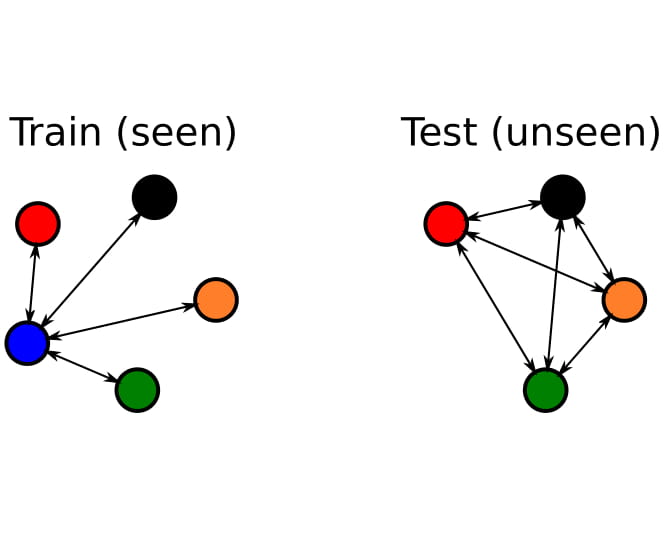}
          \caption{Seen and unseen}\label{fig:overview_seen_unseen}
      \end{subfigure}
   \caption{   Multi-domain image translation using pairwise translations: (a) all translations are seen during training, and (b) our setting: some translations are seen, then test on unseen. Best viewed in color.
   }
\end{figure}

\section{Multi-modal image translations}
We consider the problem of image translation between multiple modalities.
In particular, a translation from a source modality $\mathcal{X}^{(i)}$ to a target modality $\mathcal{X}^{(j)}$ is a mapping $T_{ij} \colon x^{(i)} \mapsto x^{(j)}$. This mapping is implemented as an encoder-decoder chain $x^{(j)}=T_{ij}\left(x^{(i)}\right)=g_j\left(f_i\left(x^{(i)}\right)\right)$ with source  encoder $f_i$ and target decoder $g_j$. Translations between modalities connected during training are all learned jointly, and in both directions. Note that the encoder and decoder of translation $T_{ij}$ are different from those of $T_{ji}$. In order to perform any arbitrary translation between modalities, all pairwise translations must be trained (i.e. seen) during the training stage (see Figure~\ref{fig:overview_all_seen}).

In this article we address the case where only a subset of the translations are seen during training, while the rest remain unseen (see Figure~\ref{fig:overview_seen_unseen}). Our objective is to be able to infer these unseen translations during test time.

\subsection{Inferring unseen translations}
In the case where some of the translations are unseen during training, we could still try to infer them by reusing the available networks. Here we discuss two possible ways: \textit{cascading translators}, which we use as baseline, and the proposed \textit{mix and match networks} approach.
\begin{figure}
    \centering
      \begin{subfigure}[b]{0.42\columnwidth}
          \centering
          \includegraphics[scale=0.2]{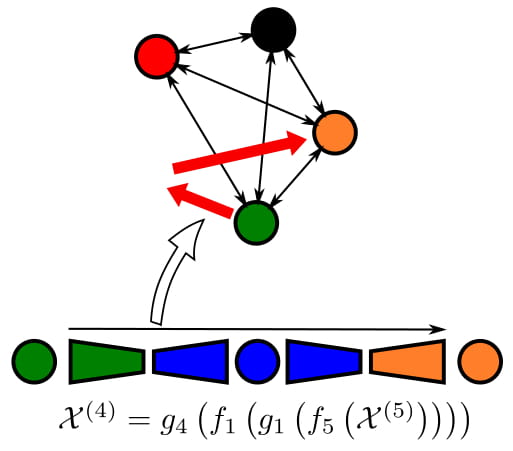}
          \caption{Cascade}\label{fig:overview_cascade}
      \end{subfigure}
      \begin{subfigure}[b]{0.33\columnwidth}
          \centering
          \includegraphics[scale=0.2]{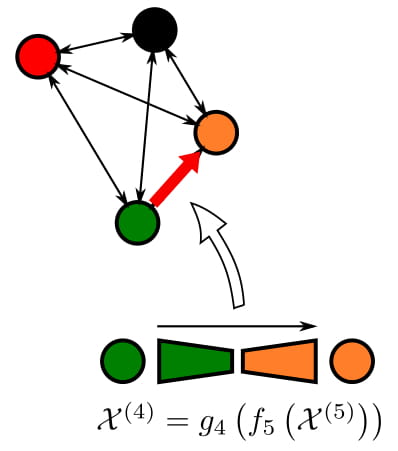}
          \caption{Mix\&match}\label{fig:overview_mixmatch}
      \end{subfigure}
      \begin{subfigure}[b]{0.215\columnwidth}
          \centering
          \includegraphics[scale=0.2]{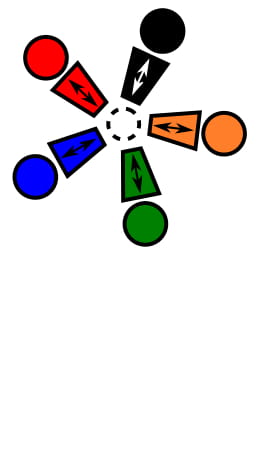}
          \caption{Ideal}\label{fig:overview_ideal}
      \end{subfigure}
  
   \caption{Inferring unseen translations: (a) cascading translators, (b) mix and match networks~(M\&MNets), and (c) ideal case of encoders-decoders with aligned representations. Best viewed in color.}
\end{figure}

\minisection{Cascaded translators}
Assuming there is a path of seen translations between the source modality and the target modality via intermediate modalities (see Figure~\ref{fig:overview_seen_unseen}), a possible solution is simply concatenating the seen translators across this path. This will result in a mapping from the source to the target modality by reconstructing images in the intermediate modalities (see Figure~\ref{fig:overview_cascade}). However, the success of this approach depends on the effectiveness of the intermediate translators.

\minisection{Unpaired translators}
An alternative is to frame the problem as unpaired translation between the source and target modalities and disregard the other modalities, learning a translation using methods based on cycle consistency~\citep{zhu2017unpaired,kim2017learning,yi2017dualgan,liu2017unsupervised}. This approach requires training an unpaired translator per unseen translation. In general, unpaired translation can be effective when the translation is within the same modality and involves a relatively small shift between source and target domains (e.g. body texture in horse-to-zebra), but struggles in the more challenging case of cross-modal translations.

\minisection{Mix and match networks (M\&MNets)}
We propose to obtain the unseen translator by simply concatenating the encoder of the source modality and the decoder of the target modality (see Figure~\ref{fig:overview_mixmatch}). The problem is that these two networks have not directly interacted during training, and therefore, for this approach to be successful, the two latent spaces must be aligned. 

\subsection{Aligning for unseen translations}\label{sec:i2imultiway}
The key challenge in M\&MNets is to ensure that the latent representation from the encoders can be decoded by all decoders, including those unseen (see Figure~\ref{fig:overview_ideal}). In order to address this challenge, encoders and decoders must be aligned in their latent representations. In addition, the encoder-decoder pair should be able to preserve the spatial structure, even in unseen translations.

In the following we describe the different techniques we use to enforce feature alignment between unseen encoder-decoder pairs.

\minisection{Shared encoders and decoders} Sharing encoders and decoders is a basic requirement to reuse latent representations and reduce the number of networks.

\minisection{Autoencoders} We jointly train modality-specific autoencoders with the image translation networks. By sharing the weights between the auto-encoders and the image translation encoder-decoder pairs the latent space is forced to align.

\minisection{Robust side information} 
In general, image translation tasks result in output images with similar spatial structure as the input ones, such as scene layouts, shapes and contours that are preserved across the translation.
In fact, this spatial structure available in the input image is critical to simplify the problem and achieve good results, especially in cross-modal translations.
Successful image translation methods often use multi-scale intermediate representations from the encoder as side information to guide the decoder in the upsampling process. Examples of side information are skip connections~\citep{he2016deep,ronneberger2015u} and pooling indices~\citep{badrinarayanan2015segnet,li2018closed}. We exploit side information in cross-modal translation (see discussion in Section~\ref{sec:side-information}). 

\minisection{Latent space consistency (only in paired settings)}
When paired data between some modalities is available, we can enforce consistency in the latent representations of each direction of the translations. \cite{taigman2016unsupervised} use L2 distance between a latent representation and the reconstructed after another decoding and encoding cycle. Here we enforce the representations $f_i\left(x^{(i)}\right)$ and $f_j\left(x^{(j)}\right)$ of two paired samples  $\left(x^{(i)},x^{(j)}\right)$,  to be aligned, since both images represent the same content (just in two different modalities). This is done by introducing  a  latent space consistency loss which is defined as  $\left\|f_i\left(x^{(i)}\right) - f_j\left(x^{(j)}\right) \right\|_{2}$.
We exploit this constraint in zero-pair image translation (see Section~\ref{sec:zero-pair}).

\minisection{Adding noise to latent space} 
The latent space consistency we apply is based on reducing the difference between the $f_i\left(x^{(i)}\right)$ and $f_j\left(x^{(j)}\right)$. The network can minimize this loss by aligning the representations of $f_i\left(x^{(i)}\right)$ and $f_j\left(x^{(j)}\right)$, but it could also minimize it by just reducing the signal   $\left\|f_i\left(x^{(i)}\right) \right\|$ and $\left\|f_j\left(x^{(j)}\right) \right\|$. This would reduce the latent space consistency loss but not improve the alignment.   Adding noise to the output of  each encoder  prevents this problem, since reducing the signal would then hurt the translation and auto-encoder losses. In practice,  we found that adding noise helps to train the networks and improves the results during test. 

\subsection{Scalable image translation with M\&MNets}\label{sec:scale}
As the number of modalities increases, the number of pairwise translations grows quadratically. Training encoder-decoder pairs for all pairwise translations in $N$ modalities would require $N\times(N-1)/2$ encoders and $N\times(N-1)/2$ decoders (see Figure~\ref{fig:overview_all_seen}). One of the advantages of M\&MNets is their better \textit{scalability}, since many of those translations can be inferred without explicitly training them (see Figure~\ref{fig:overview_seen_unseen}). It requires that each encoder and decoder should be involved in at least one translation pair during training in order to be aligned with the others, thereby reducing complexity from quadratic to linear with the number of modalities (i.e. $N$ encoders and $N$ decoders). 

\subsection{Translating domains instead of modalities}\label{sec:domains}
Although we described the proposed framework for cross-modal translation, the same framework can be easily adapted to cross-domain image translation. In that case, the modality is the same (typically RGB) and the translation is arguably less complex since the network does not need to learn to change the modality, just the domain. It can be learned sometimes with unpaired data (e.g. style transfer, face attributes and expressions). 

\begin{figure*}[t]
      \centering
      \begin{subfigure}[b]{0.24\textwidth}
              \centering
              \includegraphics[width=\columnwidth]{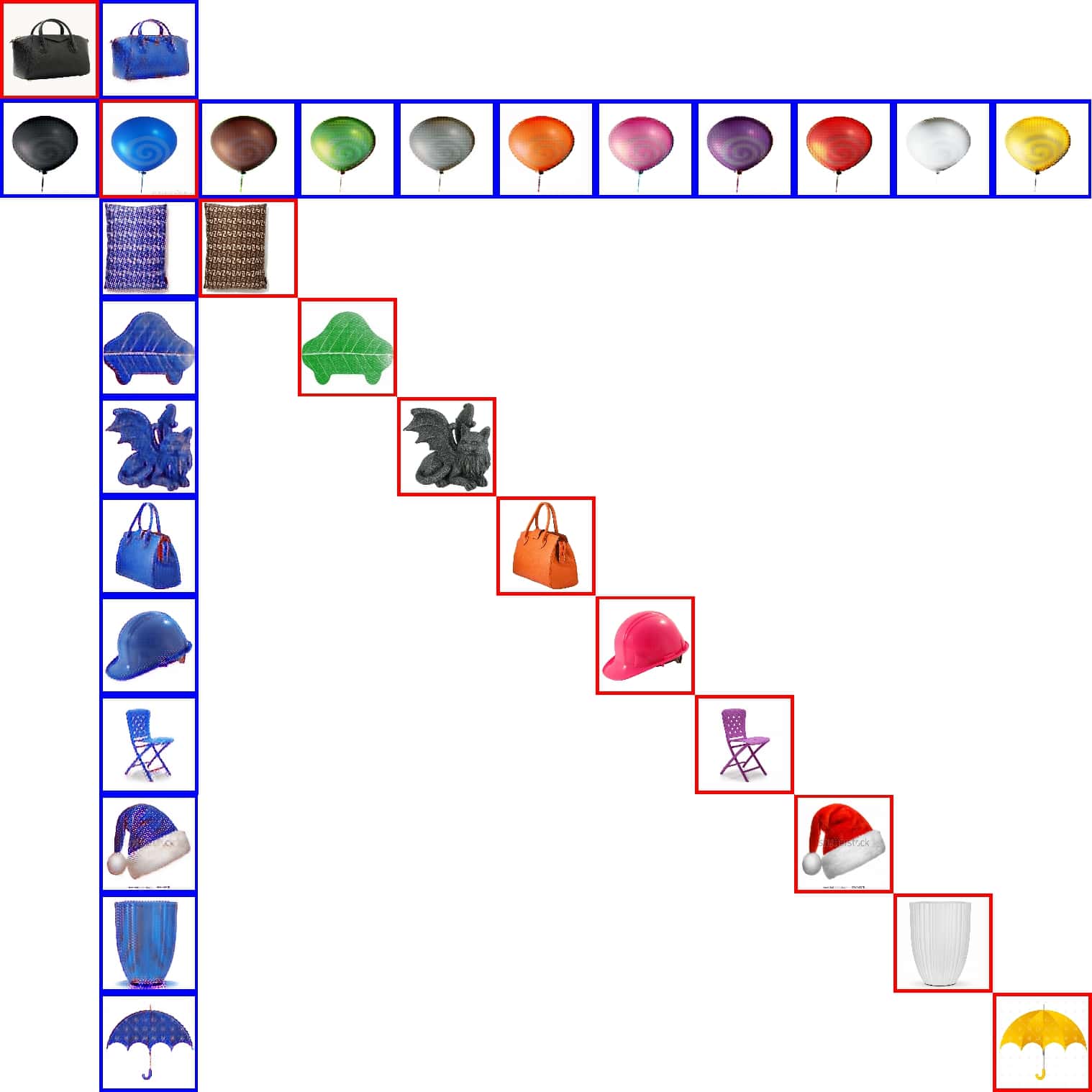}
              \caption{Input+seen}
      \end{subfigure}
      \!
      \begin{subfigure}[b]{0.24\textwidth}
              \centering
              \includegraphics[width=\columnwidth]{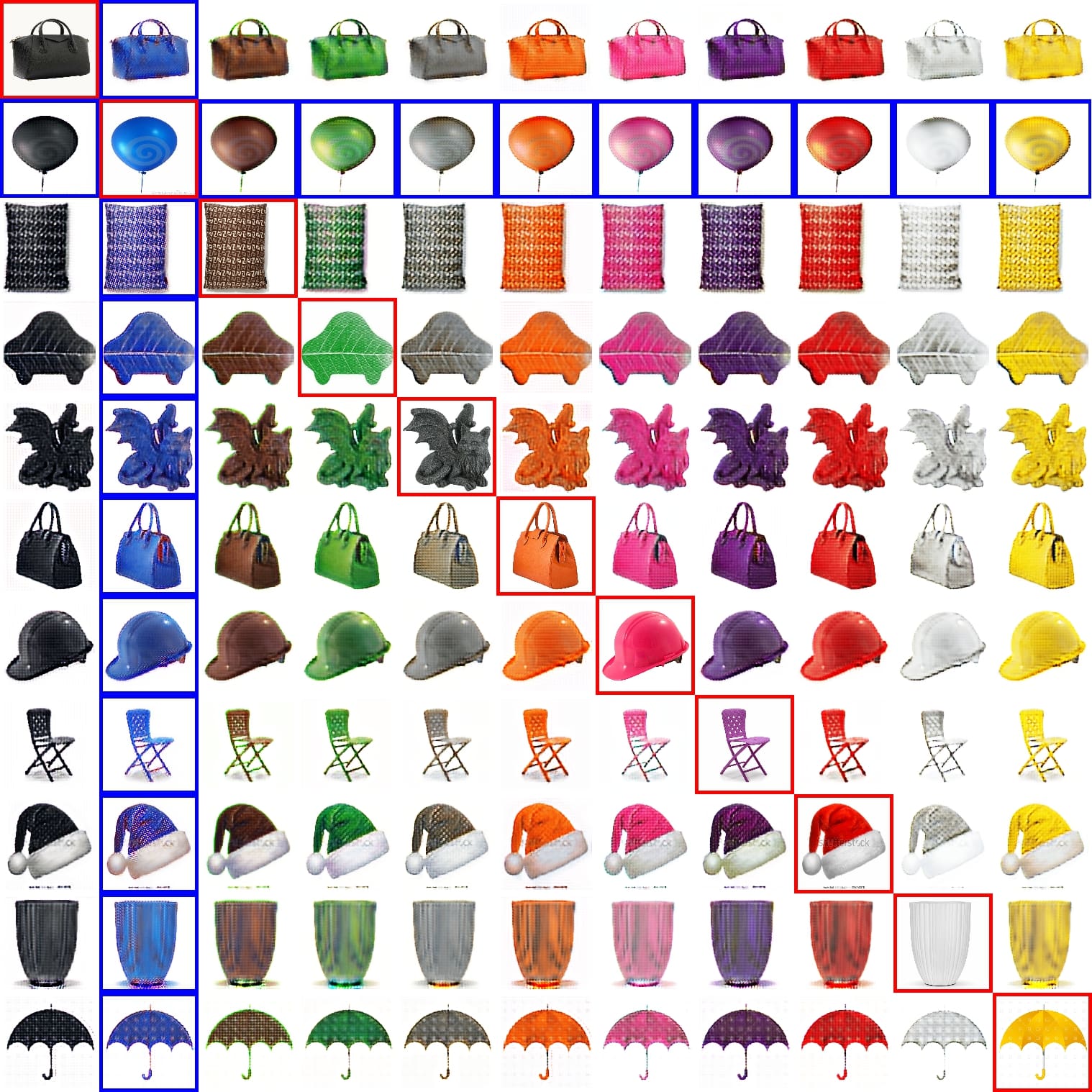}
              \caption{Input+seen+unseen}
      \end{subfigure}
      \hfill
      \begin{subfigure}[b]{0.24\textwidth}
              \centering
              \includegraphics[width=\columnwidth]{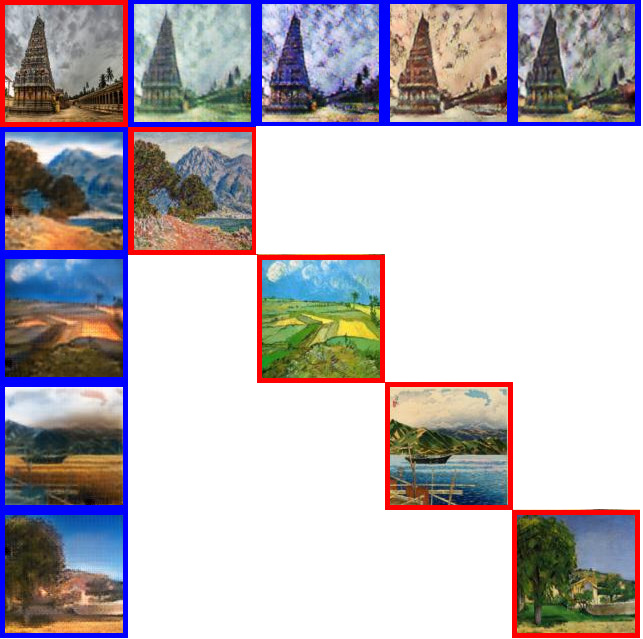}
              \caption{Input+seen}
      \end{subfigure}
      \!
      \begin{subfigure}[b]{0.24\textwidth}
              \centering
              \includegraphics[width=\columnwidth]{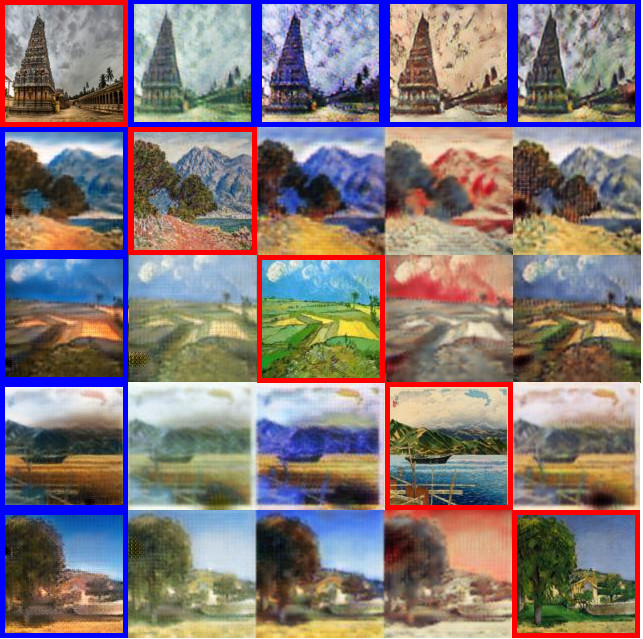}
              \caption{Input+seen+unseen}
      \end{subfigure}
        \caption{\label{fig:scalable_multidomain_translation} Two examples of scalable inference of multi-domain translations with M\&MNets. Color transfer (a-b): only transformations from blue or to blue (anchor domain) are seen. Style transfer (c-d): trained on four styles + photo (anchor) with data from \cite{zhu2017unpaired}. From left to right: photo, Monet, van Gogh, Ukiyo-e and Cezanne. Input images are highlighted in red and seen translations in blue. Best viewed in color.}
\end{figure*}

Here we use cross-domain image translation to illustrate the scalability of M\&MNets. The datasets (color and artworks) and the network architectures are provided in Appendix~\ref{Appendix_color_artworks}. Figure~\ref{fig:scalable_multidomain_translation} shows two examples involving multi-domain unpaired image translation. Figure~\ref{fig:scalable_multidomain_translation}a-b shows an image recoloring application with eleven domains ($N=11$). Images are objects in the colored objects dataset~\citep{yu2018beyond}, where we use colors as domains. A naive solution is training all pairwise recoloring combinations with CycleGANs, which requires training a total of $N\left(N-1\right)/2=55$ encoders (and decoders). In contrast, M\&MNets only require to train eleven encoders and eleven decoders, while still successfully addressing the recoloring task. In particular, all translations from or to the blue domain are trained, while the remaining translations not involving blue are unseen. The input images (framed in red) and the resulting seen translations (framed in blue) are shown in Figure~\ref{fig:scalable_multidomain_translation}a. The additional images in Figure~\ref{fig:scalable_multidomain_translation}b correspond to the remaining unseen translations.

We also illustrate M\&MNets in a style transfer setting with five domains. They include photo (used as anchor domain) and four artistic styles with data from \cite{zhu2017unpaired}). M\&MNets can reasonably infer unseen translations between styles (see Figure~\ref{fig:scalable_multidomain_translation}d) using only five encoders and five decoders (for a total of twenty possible translations). Note that the purpose of these examples is to illustrate the scalability aspect of M\&MNets in multiple domains, not to compete with state-of-the-art recoloring or style transfer methods. 

\section{Zero-pair cross-modal translation} \label{sec:zero-pair}
Well aligned M\&MNets can be applied to a variety of problems. Here, we apply them to a challenging setting we call \textit{zero-pair cross-modal translation}, which involves three modalities\footnote{For simplicity, we will refer to the output semantic segmentation maps and depth as modalities rather than tasks, as done in some works.}. Note that cross-modal translations usually  require modality-specific architectures and losses.

\subsection{Problem definition}
We consider the problem of jointly learning two seen cross-modal translations: RGB-to-segmentation translation $y=T_{RS}\left(x\right)$ (and $x=T_{SR}\left(y\right)$)  and RGB-to-depth translation $z=T_{RD}\left(x\right)$ (and $x=T_{DR}\left(z\right)$)  and evaluating on the unseen depth-to-segmentation transformations $y=T_{DS}\left(z\right)$ and $z=T_{SD}\left(y\right)$ (see Figures~\ref{fig:introduction} and \ref{fig:cross-modal-settings}c). In contrast to the conventional unpaired translation setting, here seen translations have paired data (cross-modal translation is difficult to learn without paired samples). In particular, we consider the case where the former translations are learned from a semantic segmentation dataset $\mathcal{D}_{RS}$ with pairs $\left(x,y\right)\in \mathcal{D}_{RS}$ of RGB images and segmentation maps, and the second from a disjoint RGB-D dataset $\mathcal{D}_{RD}$ with pairs of RGB and depth images $\left(x,z\right)\in \mathcal{D}_{RD}$. Therefore no pairs with matching depth images and segmentation maps are available to the system. 
The system is evaluated on a third dataset $\mathcal{D}_{DS}$ with paired depth images and segmentation maps.

\subsection{Mix and match networks architecture}\label{sec:basic_framework}
\begin{figure*}
    \centering
   \includegraphics[width=\textwidth]{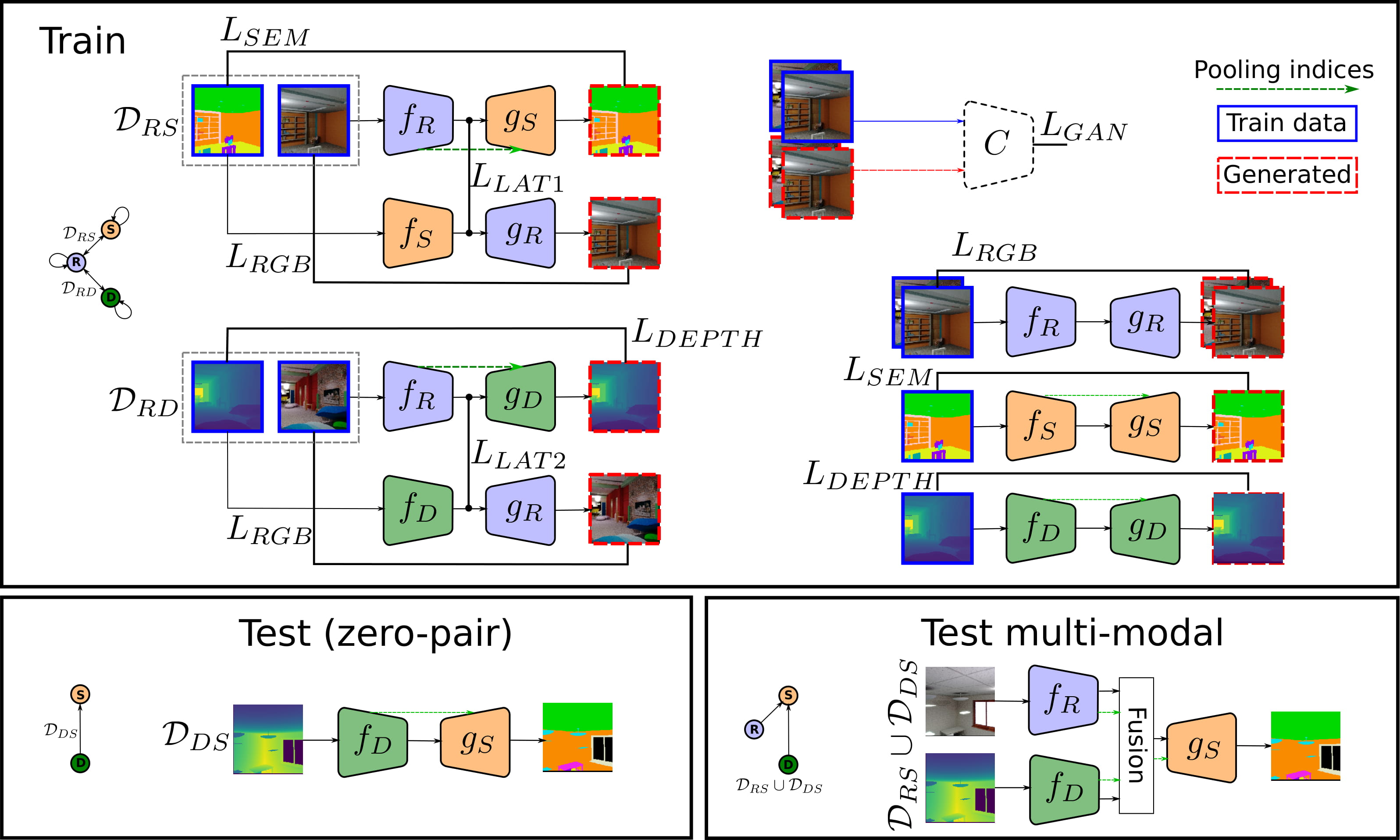}
   \caption{Zero-pair cross-modal and multi-modalimage translation with M\&MNets. Two disjoint sets $\mathcal{D}_{RS}$ and $\mathcal{D}_{RD}$ are seen during training, containing (RGB,depth) pairs and (RGB,segmentation) pairs, respectively. The system is tested on the unseen translation depth-to-segmentation (zero-pair) and (RGB+depth)-to-segmentation (multimodal), using a third unseen set $\mathcal{D}_{DS}$. Encoders and decoders with the same color share weights.  Note that we do not apply pooling indices for RGB decoders. Best viewed in color.}\label{fig:basic_framework}
\end{figure*}
The overview of the framework is shown in Figure~\ref{fig:basic_framework}. As basic building blocks we use three modality-specific encoders $f_R\left(x\right)$, $f_D\left(z\right)$ and $f_S\left(y\right)$ (RGB, depth and semantic segmentation, respectively), and the corresponding three modality-specific decoders $g_R\left(h\right)$, $g_D\left(h\right)$ and $g_S\left(h\right)$, where $h$ is the latent representation in the shared space. The required translations are implemented as $y=T_{RS}\left(x\right)=g_S\left(f_R\left(x\right)\right)$, $z=T_{RD}\left(x\right)=g_D\left(f_R\left(x\right)\right)$ and $y=T_{DS}\left(z\right)=g_S\left(f_D\left(z\right)\right)$.

Encoder and decoder weights are shared across the different translations involving same modalities (same color in Figure~\ref{fig:basic_framework}). To enforce better alignment between encoders and decoders of the same modality, we also include self-translations using the corresponding three autoencoders $T_{RR}(x)=g_R\left(f_R\left(x\right)\right)$, $T_{DD}(y)=g_D\left(f_D\left(y\right)\right)$ and $T_{SS}(z)=g_S\left(f_S\left(z\right)\right)$.

We base our encoders and decoders on the SegNet architecture~\citep{badrinarayanan2015segnet}. The encoder of SegNet itself is based on the 13 convolutional layers of the VGG-16 architecture~\citep{simonyan2014very}. The decoder mirrors the encoder architecture with 13 deconvolutional layers. Weights in encoders and decoders are  randomly initialized following a standard Gaussian distribution except for the RGB encoder which is pretrained on ImageNet~\citep{imagenet_cvpr09}.

As in SegNet, pooling indices at each downsampling layer of the encoder are provided to the corresponding upsampling layer of the (seen or unseen) decoder\footnote{The RGB decoder does not use pooling indices, since in our experiments we observed undesired grid-like artifacts in the RGB output when we use them.}. These pooling indices seem to be relatively similar across the three modalities and effective to transfer spatial structure information that help to obtain better depth and segmentation boundaries in higher resolutions. Thus, they provide relatively modality-independent side information. We also experimented with skip connections and no side information at all.

\subsection{Loss functions}
As we mentioned before, for a correct cross-alignment between encoders and decoders, training is critical for zero-pair translation. The final loss combines a number of modality-specific losses for both cross-modal translation and self-translation (i.e. autoencoders) and alignment constraints in the latent space
\begin{equation*}
L = \lambda_R L_{RGB}+\lambda_S L_{SEG}+\lambda_D L_{DEPTH}+\lambda_A L_{LAT}
\end{equation*}
where $ \lambda_R$, $\lambda_S$, $\lambda_D$ and $\lambda_A$  are weights which balance the losses.

\minisection{RGB} We use a combination of pixelwise L2 distance and adversarial loss $L_{RGB}=\lambda_{L2} L_{L2}+ L_{GAN}$. L2 distance is used to compare the ground truth RGB image and the output RGB image of the translation from a corresponding depth or segmentation image. It is also used in the RGB autoencoder
\begin{eqnarray}
L_{L2} &=&  \mathbb{E}_{(x,y)\sim \mathcal{D}_{RS}}\left[ \left\| T_{SR}\left(y\right) - x  \right\|_2 \right] \\
       &+& \mathbb{E}_{(x,z)\sim \mathcal{D}_{RD}}\left[ \left\| T_{DR}\left(z\right) -x \right\|_2 \right] \\
      &+& \mathbb{E}_{x\sim \mathcal{D}_{RS}\bigcup\mathcal{D}_{RD}}\left[ \left\| T_{RR}\left(x\right) -x \right\|_2 \right]
\end{eqnarray}
In addition, we also include the least squares adversarial loss \citep{mao2016multi,isola2016image} on the output of the RGB decoder
\begin{equation*}
L_{GAN} = \mathbb{E}_{x\sim \mathcal{D}_{RS}\bigcup\mathcal{D}_{RD}}\left[ \left( C\left(x\right) -1 \right)^2 \right]+ \mathbb{E}_{\hat{x}\sim \hat{p}(x)}\left[ \left( C\left(\hat{x}\right) \right)^2 \right]
\end{equation*} 
where $\hat{p}(x)$ is the resulting distribution of the combined images $\hat{x}$ generated by $\hat{x}=T_{SR}\left(y\right)$, $\hat{x}=T_{DR}\left(z\right)$ and $\hat{x}=T_{RR}\left(x\right)$. Note that the RGB autoencoder and the discriminator $C\left(x\right)$ are both trained with the combined RGB data $\mathcal{X}$.

\minisection{Depth}
For depth we use the Berhu loss~\citep{laina2016deeper} in both RGB-to-depth translation and in the depth autoencoder
\begin{eqnarray}
L_{DEPTH} &=& \mathbb{E}_{(x,z)\sim \mathcal{D}_{RD}}\left[ \mathcal{B}\left( T_{RD}\left(x\right) -z \right) \right] \\
      	  &+& \mathbb{E}_{(x,z)\sim \mathcal{D}_{RD}}\left[ \mathcal{B}\left( T_{DD}\left(z\right) -z \right) \right]\label{eq_berhu}
\end{eqnarray}
where $\mathcal{B}\left(z\right)$ is the average Berhu loss,  which is given by 
\begin{eqnarray}
\mathcal{B}\left({z}' -z \right) &=&  \left\{\begin{matrix}
\left |\left( {z}' -z \right)  \right | & \left | {z}' -z  \right |\leqslant c & \\ 
\frac{ \left ({z}' -z \right) ^{2} + c^{2}}{2c}& \left | {z}' -z  \right | > c & 
\end{matrix}\right.
\end{eqnarray}
where ${z}' = T_{RD}\left(x\right)$, and $c=\frac{1}{5}{\max}_{i}\left ( \left | {z}'_{i}-z_{i} \right | \right )$, where $i$ indexes the pixels of each image.

\minisection{Semantic segmentation} For segmentation we use the average cross-entropy loss $\mathcal{CE}\left(\hat{y},y\right)$ in both RGB-to-segmentation translation and the segmentation autoencoder
\begin{eqnarray}
L_{SEM} &=& \mathbb{E}_{ (x,y)\sim \mathcal{D}_{RS}}\left[ \mathcal{CE}\left( T_{RS}\left(x\right),y \right) \right] \\
        &+& \mathbb{E}_{(x,y)\sim \mathcal{D}_{RS}}\left[ \mathcal{CE}\left( T_{SS}\left(y\right),y \right) \right].
\end{eqnarray}

\minisection{Latent space consistency} We enforce latent representations to remain close, independently of the encoder that generated them. In our case we have two latent space consistency losses

\begin{eqnarray}
L_{LAT} &=& L_{LAT1}+ L_{LAT2} \\
L_{LAT1} &=& \mathbb{E}_{(x,y)\sim \mathcal{D}_{RS}}\left[ \left\| f_R\left(x\right) - f_S\left(y\right) \right\|_2 \right] \\
		L_{LAT2} &=& \mathbb{E}_{(x,z)\sim \mathcal{D}_{RD}}\left[ \left\| f_R\left(x\right) - f_D\left(z\right) \right\|_2 \right]        
\end{eqnarray}

 \subsection{The role of side information}\label{sec:side-information}
Spatial side information plays an important role in image translation, especially in cross-modal translation (e.g. semantic segmentation). Reconstructing images requires reconstructing spatial details. Side information from a particular encoder layer can provide helpful hints to the decoder about how to reconstruct the spatial structure at a specific scale and level of abstraction. 

\minisection{Skip connections} Perhaps the most common type of side information connecting encoders and decoders comes in the form of \textit{skip connections}, where the feature from a particular layer is copied and concatenated with another feature further in the processing chain. U-Net~\citep{ronneberger2015u} introduced a widely used architecture in image segmentation and image translation where convolutional layers in encoder and decoder are mirrored and the feature of a particular encoding layer is concatenated with the feature with the corresponding layer at the decoder.
It is important to observe that skip connections make the decoder heavily condition on the particular features of the encoder. This is not a problem in general because translations are usually seen during training and therefore latent representations are aligned. However, in our setting with unseen translations that conditioning is simply catastrophic, because the target decoder is only aware of the features in encoders from modalities seen during training. Otherwise, as in the case of an unseen encoder, the result is largely unpredictable.

\minisection{Pooling indices} The SegNet architecture~\citep{badrinarayanan2015segnet} includes unpooling layers that leverage pooling indices from the mirror layers of the encoder.
Pooling indices capture the locations of the maximum values in the input feature map of a max pooling layer.
These locations are then used to guide the corresponding unpooling operation in the decoder, helping to preserve finer details. Note that pooling indices are more compact descriptors than encoder features from skip connections, and since the unpooling operation is not learned, pooling indices are less dependent on the particular encoder and therefore more robust for unseen translations.

\section{Shared information between unseen modalities}\label{sec:pseudo-pairs}
\subsection{Shared and modality-specific information}
The information conveyed by the latent representation is key to perform image translation. Encoders extract this information from the input image and decoders use it to reconstruct the output image. In general, this latent representation can contain information shared between the source and target modalities, and information specific to each modality.
In a setting where the same latent representation is used across multiple encoders and decoders, the latent representation must capture information about all input and output modalities.

\begin{figure}
    \centering
    \begin{subfigure}[b]{0.46\columnwidth}
   \includegraphics[width=\columnwidth]{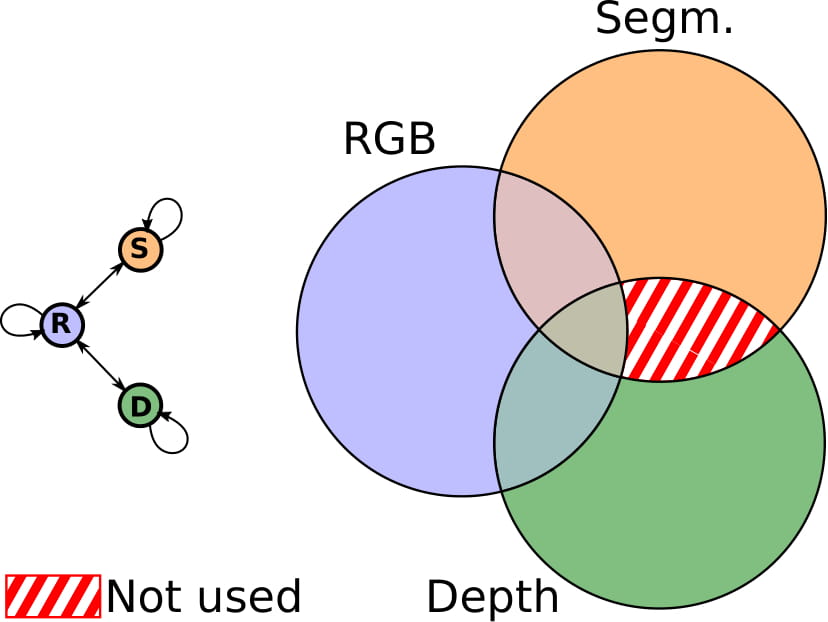}
   \caption{Seen shared information\\}
   \end{subfigure}
   \hspace*{\fill}
   \begin{subfigure}[b]{0.46\columnwidth}
   \includegraphics[width=\columnwidth]{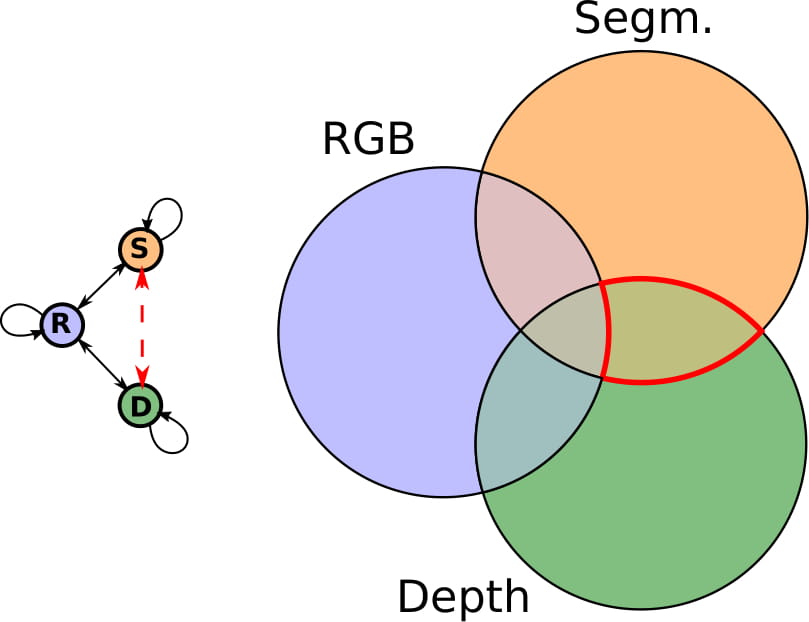}
   \caption{Seen+unseen shared information}
   \end{subfigure}
   \begin{subfigure}[b]{0.55\columnwidth}
   \vspace{3mm}
   \includegraphics[width=\columnwidth]{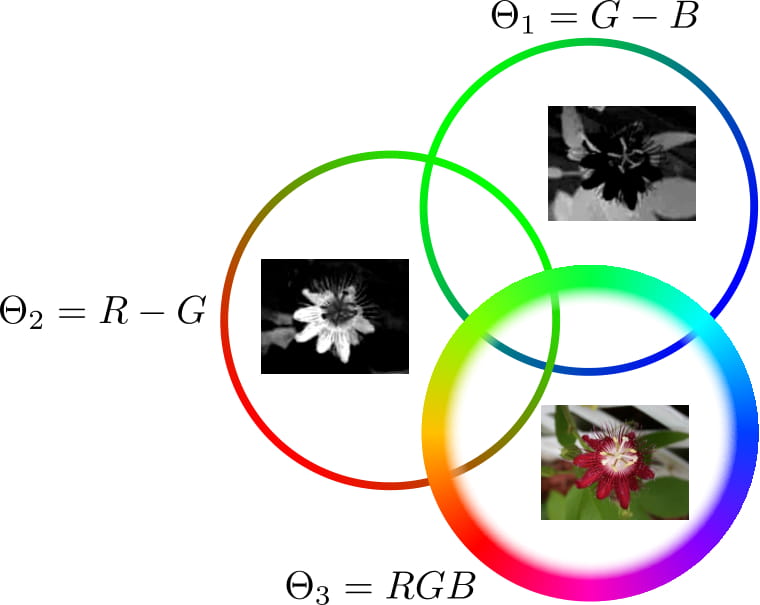}
   \caption{Color opponents example}\label{fig:circles_flowers}
   \end{subfigure}
   \caption{Specific and shared information: (a) basic mix and match nets (see Fig~\ref{fig:basic_framework}) ignore depth-segmentation shared information, (b) extended mix and match net exploiting depth-segmentation shared information (unpaired information in our case), and (c) illustration using color opponents (trained on $(\Theta_{1}$,$\Theta_{2})$ and $(\Theta_{1}$,$\Theta_{3})$, and evaluated on unseen translation $(\Theta_{2}$,$\Theta_{3})$). Best viewed in color.}\label{fig:circles}
\end{figure}

We can represent modalities as circles, whose intersections represent shared information between them. Figure~\ref{fig:circles}a represents the particular case of zero-pair cross-modal translation with three modalities (described in the previous section). Note that translators and autoencoders force the latent representation to capture both shared and modality-specific information. However, the better the information shared between modalities is captured in the latent representation, the more effective cross-modal translations are.

The framework described in Section~\ref{sec:basic_framework} enables the inference of unseen translations via the anchor modality RGB, whose encoder and decoder are shared across the two seen translations. That is the only component that indirectly enforces alignment of depth and segmentation encoders and decoders. Therefore, the latent information used in the unseen translation is the one shared by the three modalities.

In contrast, the information shared between depth and segmentation that is not shared with RGB (the dashed region in Figure~\ref{fig:circles}a) is not exploited during training by depth and segmentation encoders and decoders, because it is of no use to solve any of the seen translations. This makes inferred translations less effective because depth and segmentation encoders are ignoring potentially useful information that could improve translation to segmentation and depth, respectively. In this section, we propose an extension of our basic framework that aims at explicitly enforcing alignment between unseen modalities in order to  exploit all shared information between unseen modalities (see the highlighted region in Figure~\ref{fig:circles}b). Since no training pairs between those modalities are available, that alignment requires to be between unpaired samples.

\subsection{Exploiting shared information between unseen modalities}
\begin{figure}
    \centering
    \includegraphics[width=0.9\columnwidth]{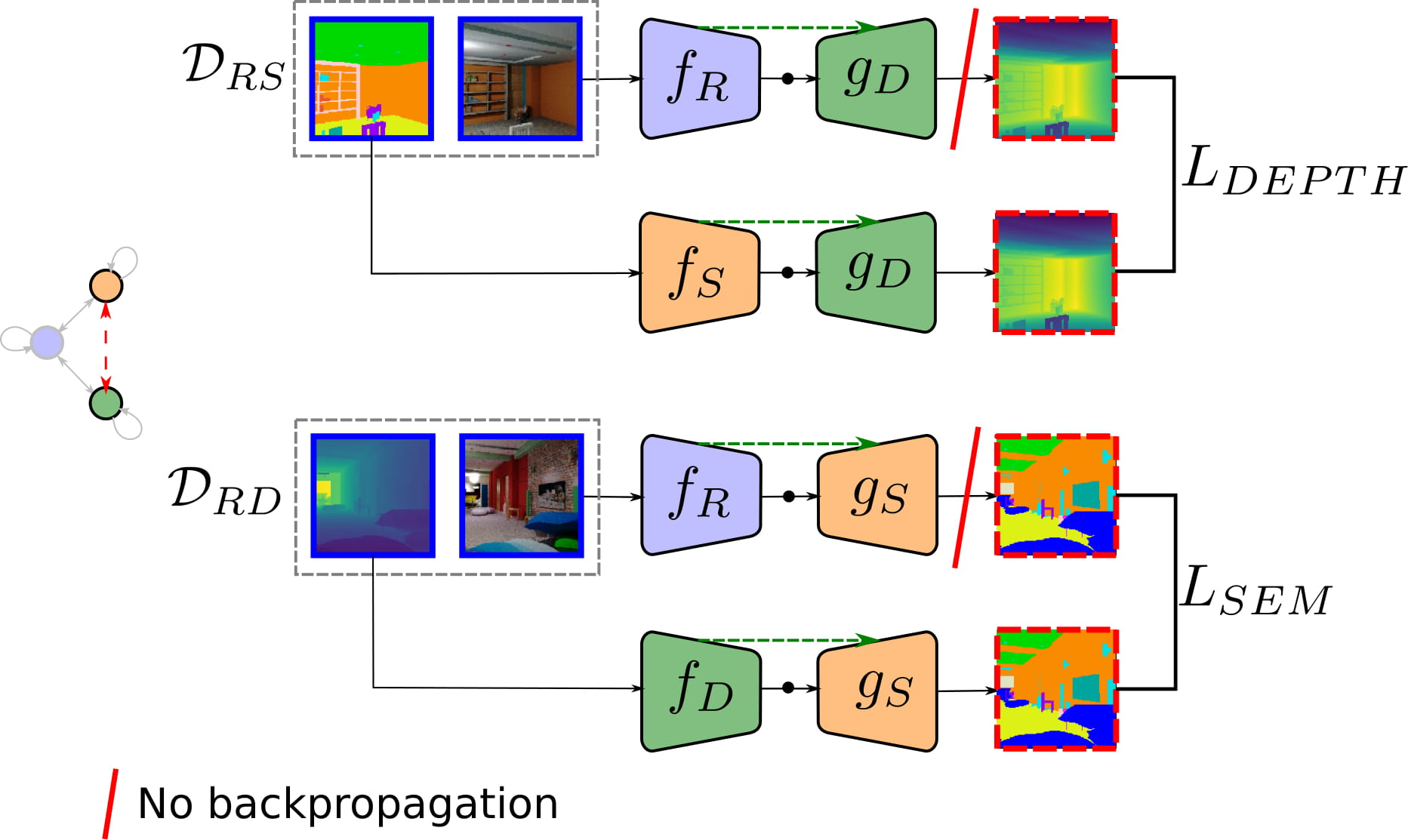}
   \caption{Pseudo-pairs pipeline on the unseen translation. This pipeline is combined with the basic cross-modal M\&MNets of Fig~\ref{fig:basic_framework}. }\label{fig:frameworks_pseudo}
\end{figure}

We adapt the idea of pseudo-labels, used previously in unsupervised domain adaptation~\citep{saito2017asymmetric,zou2018domain}, to our zero-pair cross-modal setting. The main idea is that we would also like to train directly the encoder-decoder between the unseen modalities. However, since we have no paired data between these modalities, we propose to use pseudo-pairs. 

In our specific zero-pair cross-modal setting, recall we use $x$, $y$, and $z$ to respectively indicate data from the the RGB, semantic segmentation and depth modality. We  use the encoder-decoder networks between the seen modalities to form the pseudo-pairs $(T_{RD}(x),y)$ and  $(T_{RS}(x),z)$. Now we can also train encoders and decoders between the unseen modalities depth and segmentation (see Figure~\ref{fig:frameworks_pseudo}) using the following loss:
 \begin{eqnarray}
 L_{PP} &=& \mathbb{E}_{(x,y)\sim \mathcal{D}_{RS}}\left[ \mathcal{B}\left( 
 T_{RD}\left(x\right) -T_{SD}\left(y\right) \right) \right] \\
 &+& \mathbb{E}_{(x,z)\sim \mathcal{D}_{RD}}\left[ \mathcal{CE}\left( T_{RS}\left(x\right),T_{DS}\left(z\right)\right) \right]
 \end{eqnarray}
where $\mathcal{B}$ is the average Berhu loss~\citep{laina2016deeper}, and $\mathcal{CE}$ is the cross-entropy loss. The direct training of the encoder-decoder between the unseen modality allows us to exploit  correlation between features in these modalities for which no evidence exists in the RGB modality (dashed region in Figure~\ref{fig:circles}a). In practice we first train the network without the pseudo-labels. After convergence we add $L_{PP}$ and train further with all losses until final convergence.

Note that this additional term encourages the segmentation-to-depth and depth-to-segmentation translators to exploit this shared information between the unseen modalities, including the previously ignored one, in order to improve the translation to match the one obtained from RGB. The latter is more accurate since it has been trained with paired samples. A problem with this approach is that this new loss can harm the training of seen translations from RGB, since pseudo-labels are less reliable than true labels. For this reason we do not update the weights of the translators involving RGB with the pseudo-pairs (this is indicated with the red line in Figure~\ref{fig:frameworks_pseudo}).

\subsection{Pseudo-pair example}
To illustrate the potential of pseudo-pairs we consider a cross-domain image translation example where the not-used part between the unpaired domains (striped region in Figure~\ref{fig:circles}) is expected to be substantial. We consider the task of estimating an RGB image from a single channel. In particular, we consider the following three domains\footnote{We choose the opponent channels because they are less correlated than the R, G and B channels~\citep{geusebroek2001color}.}
\begin{equation}
\begin{array}{l}
 \Theta _1  = R - G \\ 
 \Theta _2  = G - B \\
 \Theta _3 = \left( {R,G,B} \right) \\ 
 \end{array}
\end{equation}
where $\Theta _1$ and $ \Theta _2$ are scalar images and $\Theta_3$ is a three channel RGB image (see Figure~\ref{fig:circles_flowers}). Both domains $\Theta_1$ and $\Theta_2$ contain relevant and complementary information on estimating the RGB image. 

For this experiment we use the ten most frequent classes of the Flower dataset~\citep{nilsback2008automated} which are \textit{passionflower}, \textit{petunia}, \textit{rose}, \textit{wallflower}, \textit{watercress}, \textit{waterlily}, \textit{cyclamen}, \textit{foxglove}, \textit{frangipani}, \textit{hibiscus}.  For training we have pairs ($\Theta_{1}$, $\Theta_{2}$) and ($\Theta_{1}$, $\Theta_{3}$) of non-overlapping images. For test we use a separate test set. To evaluate the quality of the computed RGB images, we apply a flower classification algorithm on them and report the classification accuracy (See Appendix~\ref{Appendix_flower}).

\begin{table}
\centering
\setlength{\tabcolsep}{1.5pt}
\resizebox{\columnwidth}{!}{
\begin{tabular}{ccc}
\hline
Type & Method & Accuracy (\%) \\
\hline
\multirow{2}{*}{Seen} & \textbf{Paired} &  \\
& M\&MNets $\Theta_{1} \rightarrow \Theta_{3} $ &{75.0}         \\
\hline
\multirow{3}{*}{Unseen} & \textbf{Zero-pair}
& \\
& M\&MNets $\Theta_{2} \rightarrow \Theta_{3} $ &{36.5}         \\
& M\&MNets+PP  $\Theta_{2} \rightarrow \Theta_{3} $ &\textbf{57.5}         \\
\hline
\multirow{3}{*}{Seen/unseen} & \textbf{Multi-modal}&  \\
& M\&MNets $(\Theta_{1},\Theta_{2}) \rightarrow \Theta_{3} $
&{77.5}    \\
& M\&MNets + PP $(\Theta_{1},\Theta_{2}) \rightarrow \Theta_{3} $
&\textbf{80.5}    \\

\hline
\end{tabular}
}
  \caption{Flower classification accuracy obtained on $\Theta_{3}$ computed for various image translation models. The importance of pseudo-pairs can be clearly seen.}\label{table:flower accuracy}%
\end{table}

The results are presented in Table~\ref{table:flower accuracy}. In the first two rows the result of M\&MNets with and without pseudo-pairs are compared. The usage of pseudo-pairs results in a huge absolute performance gain of 21\%. This shows that, for domains which have considerable amounts of complementary information, pseudo-pairs can significantly improve performance. In the next two rows, we have also included the multi-modal results. In this case the pseudo-pairs double the performance gain with respect to the paired domain (last row) from $77.5-75=2.5\%$ to $80.5-75=5.5\%$.  

\begin{figure}[tb]
\centering
\includegraphics[width=\columnwidth]{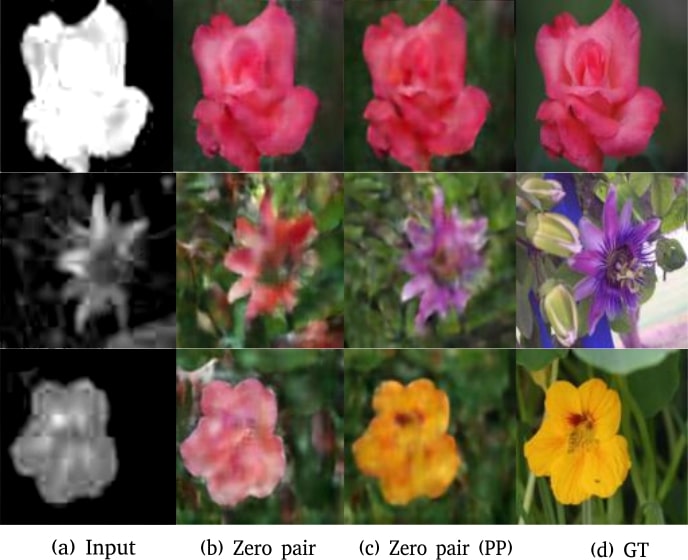}
\caption{Visualization of RGB image estimation in Flowers dataset. (a) input image from $\Theta_2$ (via seen translation), (b) zero pair translation without pseudo-pairs~\citep{wang2018mix}, (c)  zero pair with the pseudo-pairs (PP), (d) ground truth. }\label{fig:cvpr_and_pseudo_flower} 
\end{figure}

The qualitative results are provided in Figure~\ref{fig:cvpr_and_pseudo_flower}. The results show the effectiveness of the pseudo-pairs. The method without the pseudo-pairs can only exploit information which is shared between the three domains. The domain $\Theta_1$ contains information about the red-green color axes, and the mix and match nets (without pseudo-pairs) approach  does partially manage to reconstruct that part (see first row Figure~\ref{fig:cvpr_and_pseudo_flower}). However, $\Theta_1$ has no access to the blue-yellow information which is encoded in the $\Theta_2$. Adding the pseudo-pairs allows to exploit this information and the reconstructed RGB images are closer to the ground truth image (see second and third row Figure~\ref{fig:cvpr_and_pseudo_flower}).

\section{Experiments}
In this section we demonstrate the effectiveness of M\&MNets and their variants to address unseen translations in the challenging cross-modal translation setting involving the modalities RGB, depth and segmentation.

\subsection{Datasets and experimental settings}
We use two RGB-D datasets annotated with segmentation maps, one with synthetic images and the other with real captured images. A third dataset also includes near infrared (NIR) as a fourth modality.

\minisection{SceneNet RGB-D}  The SceneNet RGB-D dataset~\citep{McCormac:etal:ICCV2017} consists of 16865 synthesized training videos and 1000 test videos. Each of them contains 300 frames representing the same scene in a multi-modal triplet (RGB, depth and segmentation), with a size of 320x240 pixels. We collected 150K triplets for our training set, 10K triplets for our validation set and 10K triplets for our test set.
The triplets are sampled uniformly from the first frame to the last frame every 30 frames. The triplets for the validation set are collected from the remaining training videos and the test set is taken from the test dataset.

In order to evaluate zero-pair translation, we divided the training set (and validation set) into two equal non-overlapping splits from different videos (to avoid covering the same scenes). We discard depth images in one set and segmentation maps in the other, thus creating two disjoint training sets with paired instances,  $\mathcal{D}_{RS}$ and $\mathcal{D}_{RD}$ respectively, to train our model.

\minisection{SUN RGB-D}  The SUN RGBD dataset~\citep{song2015sun} contains 10335 real RGB-D images of room scenes. Each RGB image has a corresponding depth and segmentation map. We collected two sets: 10K triplets for the training set and 335 triplets for test set. For the training set, we split it into two disjoint subsets, one containing (RGB, segmentation) pairs, and the other containing (RGB, depth) pairs, each of them consisting of 5K pairs. 

\minisection{Freiburg Forest} The Freiburg Forest dataset~\citep{valada2016deep} consists of images of 1024$\times$768. We crop images (RGB, depth, NIR and semantic segmentation) to $256 \times 256$. We consider five different semantic classes: \textit{Sky}, \textit{Trail}, \textit{Grass}, \textit{Vegetation} and \textit{Obstacle}. Note we combine the \textit{tree} and \textit{vegatation} into an single  class (\textit{Vegetation}) as suggested in~\citep{valada2016deep}. We use the training and test datasets splits provided by the authors.

\minisection{Network training} We use  Adam~\citep{kingma2014adam} with a batch size of 6, using a learning rate of 0.0002. We set $\lambda_R = 1$, $\lambda_S = 100$, $\lambda_D = 10$, $\lambda_A =1$, $\lambda_{L2} =1$. We initially train the mix and match framework without autoencoders, without latent consistency losses, and without adding noise during the first 200K iterations. Then we freeze the RGB encoder, add the autoencoders, latent consistency losses and noise to the latent space, and for the following 200K iterations we use $\lambda_R = 10$, $\lambda_A =10$, $\lambda_{L2} =100$. We found that the network converges faster using a larger $\lambda_A$ for the second stage.
The noise is sampled from a Gaussian distribution with zero mean and a standard deviation of 0.5. For the variant with pseudo-pairs, in a third stage we include the pseudo-pair pipeline and the corresponding loss and train for another additional 100K iterations, using $\lambda_{PP} = 100$ and  learning rate 0.00002.  We experimentally found that the above setting also achieves outstanding performance on the Freiburg Forest dataset. The network information is displayed in Appendix~\ref{Appendix_rgbd}.

\minisection{Evaluation metrics} Following common practice, for the segmentation modality we compute the intersection over union (IoU) and per-class average (mIoU), and the global scores, which gives the percentage of correctly classified pixels. For the depth modality we also include quantitative evaluation, following the standard error metrics for depth estimation~\citep{eigen2015predicting}:

\begin{equation}
\begin{array}{l}
\begin{split}
\delta < & = 
\frac1{|y|}\sum_{y_i\in y}\left[ \delta(y_i,y'_i) < \nu \right]
\\
\textrm{RMSE (linear)} & =  
\sqrt{\frac1{|y|}\sum_{y_i\in y}\left \|y_i - y'_i\right \|^2}
\\
\textrm{RMSE (log)} & =
\sqrt{\frac1{|y|}\sum_{y_i\in y} \left \| \log y_i - \log y'_i \right \| ^2}
\end{split}
\end{array}
\end{equation}
where $y$ and $y'$ are the predicted and ground truth depth images, $\delta(u,v)=\max(\frac{u}{v},\frac{v}{u})$ and $\left[ P \right]$ is the Iverson bracket which is 1 when $P$ is true and 0 otherwise.

\subsection{Experiments on SceneNet RGB-D}
\subsubsection{Ablation study}
We first performed an ablation study on the impact of several design elements on the overall performance of the system. We use a smaller subset of SceneNet RGB-D based on 51K triplets from the first 1000 videos (selecting 50 frames from the first 1000 videos for training, and the first frame from another 1000 videos for test).

\minisection{Side information} 
We first evaluate the usage of side information from the encoder to guide the upsampling process in the decoder. We consider three variants: no side information, skip connections~\citep{ronneberger2015u} and pooling indices~\citep{badrinarayanan2015segnet}. The results in Table~\ref{table:side_information} show that skip connections obtain worse results than no side information at all. This is caused by the fact that side information makes the decoder(s) conditioned on the \textit{seen} encoder(s). This is problematic for \textit{unseen} translations because the features passed through skip connections are different from those seen by the decoder during training, resulting in a drop in performance.
In contrast, pooling indices provide a significant boost over no side information. Although the decoder is still conditioned to the particular seen encoders, pooling indices seem to provide helpful spatial hints to recover finer details, while being more invariant to the particular input-output combination, and even generalizing to unseen ones.

\begin{figure}[t]
\centering
\includegraphics[width=1\columnwidth]{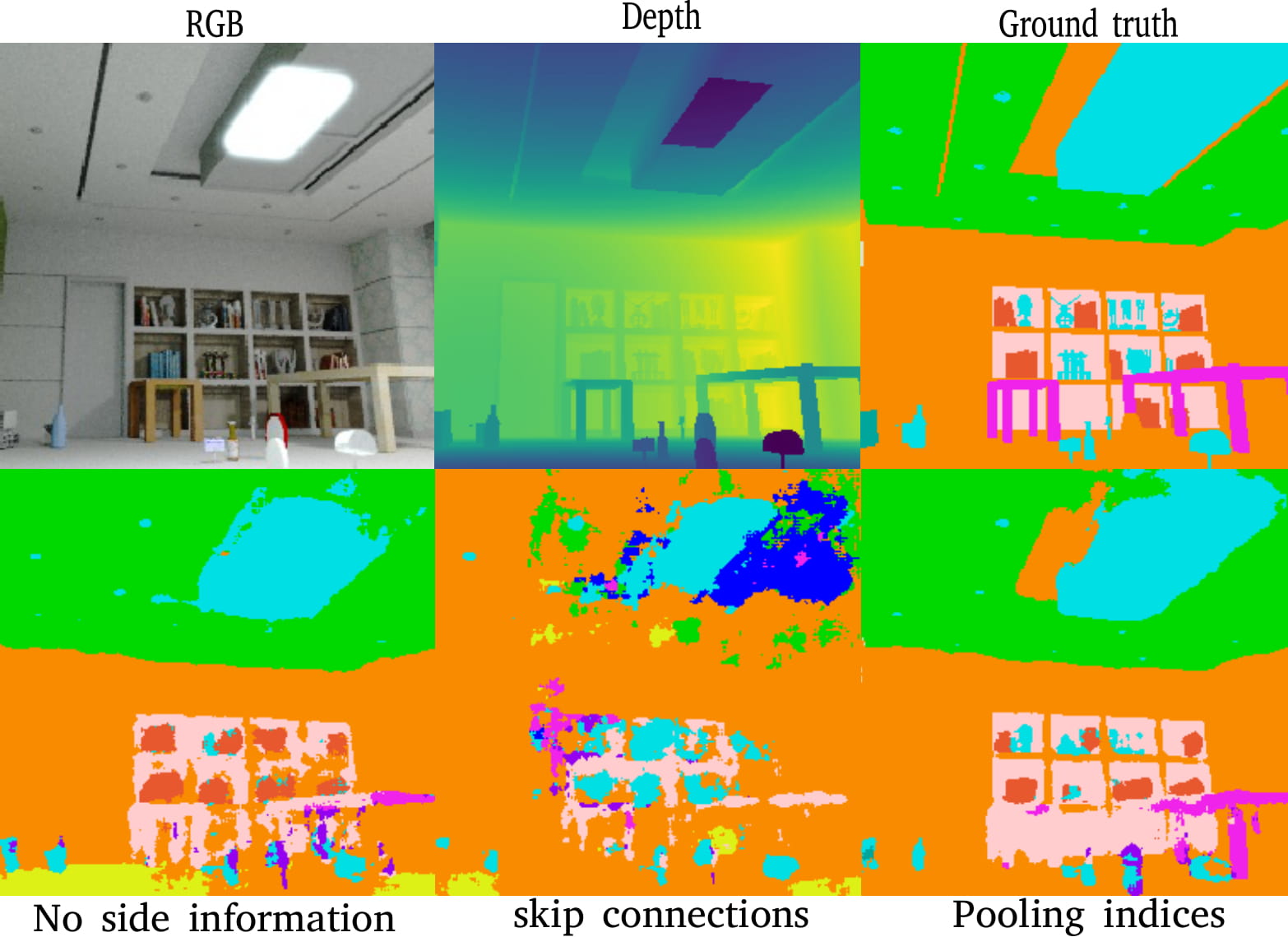}
\caption{\label{fig:example_depth-to-segm} Role of side information in unseen depth-to-segmentation translation in SceneNet RGB-D.}%
\end{figure}

Figure~\ref{fig:example_depth-to-segm} illustrates the differences between these three variants in depth-to-segmentation translation. Without side information the network is able to reconstruct a coarse segmentation, but without further guidance it is not able to refine it properly. Skip connections completely confuse the decoder by providing unseen encoding features. Pooling indices are able to provide helpful hints about spatial structure that allows the unseen decoder to recover finer segmentation maps. 

\begin{table}[tb]
\setlength{\tabcolsep}{2.5pt}
\centering
\resizebox{0.8\columnwidth}{!}{
\begin{tabular}{cc|cc}
\hline
Side information  &Pretrained  &mIoU  &Global  \\ 
\hline  
-  & N  &{29.8\%}    &{61.6\%} \\ 
Skip connections  & N     &{12.7\%}    &{50.1\%} \\
Pooling indices  & N  &{43.2\%}    &{73.5\%} \\
Pooling indices  & Y    &\textbf{46.7\%}    &\textbf{78.4\%}    \\
\hline
\end{tabular}
}
\caption{Influence of side information and RGB encoder pretraining on the final results. The task is zero-shot depth-to-semantic segmentation in SceneNet RGB-D (51K).}%
\label{table:side_information}
\end{table}

\minisection{RGB pretraining} We also compare training the RGB encoder from scratch and initializing with pretrained weights from ImageNet. Table~\ref{table:side_information} shows an additional gain of around 4\% in mIoU when using the pretrained weights.

Given these results we perform all the remaining experiments initializing the RGB encoder with pretrained weights and use pooling indices as side information.

\minisection{Latent space consistency, noise and autoencoders} We evaluate these three factors in Table~\ref{table:ablation}. The results show that latent space consistency and the usage of autoencoders lead to significant performance gains; for both, the performance (in mIoU) is more than doubled. Adding noise to the output of the encoder results in a small performance gain.  The results in Table~\ref{table:ablation} do not apply pooling indices for the RGB decoder (as also shown in Fig.~\ref{fig:basic_framework}). When we add pooling indices to our approach without pseudo-pairs,
results drop from 46.7\% to 42.4\% in mIoU. This could be  because we focus on unseen translations to depth or segmentation modalities, which do not include reconstructing the RGB modality. We believe that forcing the RGB decoder to use pooling indices to reconstruct RGB images lowers the efficiency of the latent representation to reconstruct depth or segmentation. Hence, we sacrifice some of the performance translating to the RGB modality to improve the results for depth and semantic segmentation.

\minisection{Pseudo-pairs} We also evaluate the impact of using pseudo-pairs to exploit shared information between unseen modalities. Table~\ref{table:ablation} shows a significant gain of almost 3\% in mIoU and a more moderate gain in global accuracy.

\begin{table}[tb]
\centering
\setlength{\arrayrulewidth}{1\arrayrulewidth}
\begin{tabular}{cccc|cc}
\hline
AutoEnc   &Latent   &Noise &PP &mIoU  &Global
\\ 
\hline  
 N  & N  & N  & N   &{6.48\%}    &{15.7\%}   \\
 Y  & N  & N  & N   &{20.3\%}    &{49.4\%}   \\
 Y  & Y  & N  & N   &{45.8\%}    &{76.9\%}  \\
 Y  & Y  & Y  & N   &{46.7\%}    &{78.4\%}  \\
 Y  & Y  & Y  & Y   &\textbf{49.2\%}    &\textbf{80.5\%}  \\

\hline
\end{tabular}
\caption{Impact of several components (autoencoder, latent space consistency loss, noise and pseudo-pairs) in the performance. The task is zero-pair depth-to-segmentation in SceneNet RGB-D (51K). PP: pseudo-pairs.}
\label{table:ablation}
\end{table}

In the following sections we use the SceneNet RGB-D dataset with 170K triplets. 

\subsubsection{Monitoring alignment}
\begin{figure}[tb]
\centering
\includegraphics[width=\columnwidth]{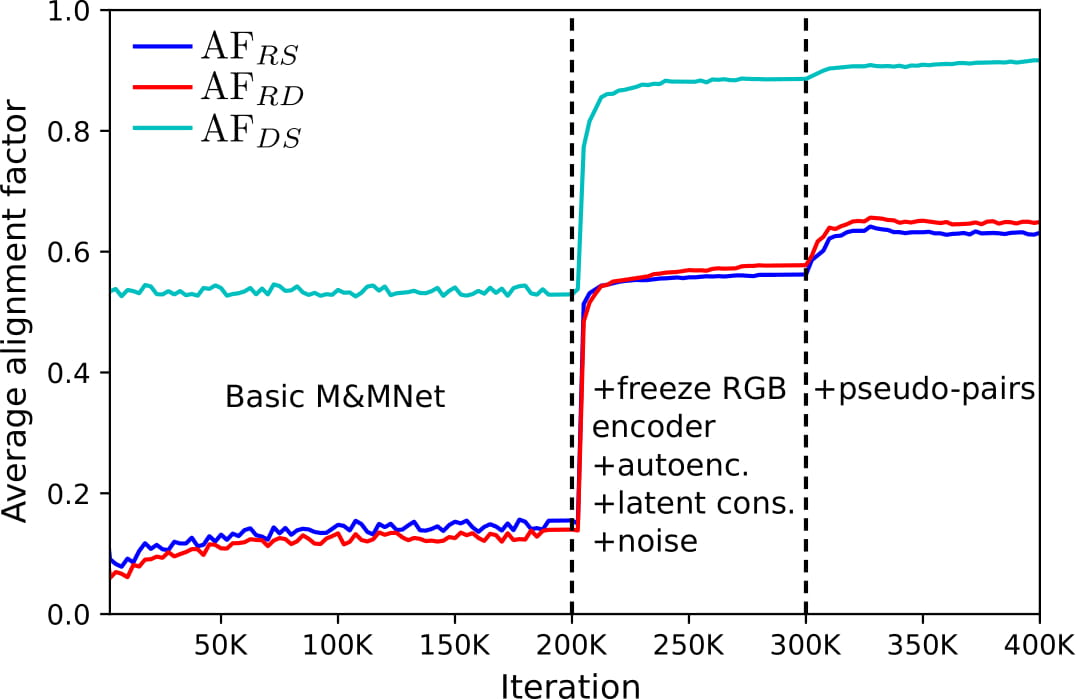}
\caption{\label{fig:latent_space_alignment} 
Monitoring alignment between latent features on SceneNet RGB-D.}
\end{figure}

The main challenge for M\&MNets is to align the different modality-specific bottleneck features, in particular for unseen translations. We measure the alignment between the features extracted from the triplets in the test set $\mathcal{D}_{DS}$ . For each triplet $\left(x,y,z\right)$ (i.e. RGB, segmentation and depth images) we extract the corresponding triplet of latent features $\left ( f_{R}\left ( x \right ), f_{S}\left ( y \right ), f_{D}\left ( z \right ) \right )$  and measure their average pairwise cross-modal alignment. The alignment between RGB and segmentation features is measured using the following alignment factor 
\begin{eqnarray}
\textrm{AF}_{RS}&=&E_{\left ( x, y \right  )\sim {\mathcal{D}_{RS}}} \left [ \frac{f_R\left(x\right)^\intercal f_S\left(y\right)}{\left \| f_R\left(x\right) \right \|\left \|  f_S\left(y\right) \right \|} \right ]
\end{eqnarray}
The other alignment factors $\textrm{AF}_{RD}$ and $\textrm{AF}_{DS}$ between RGB and depth features and between depth and segmentation features are defined analogously. Figure~\ref{fig:latent_space_alignment} shows the evolution of this alignment during training and across the different stages. The three curves follow a similar trend, with the alignment increasing in the first iterations of each stage and then stabilizing. The beginning of stage two shows a dramatic increase in the alignment, with a more moderate increase at stage three. These results are consistent with those of the ablation study of the previous section, showing that better alignment typically leads to better results in unseen translations. Overall, they show that latent space consistency, autoencoders, pseudo-pairs and pooling indices contribute to the effectiveness of M\&MNets to address unseen image translation in the zero-pair setting.

\subsubsection{Comparison with other models}
In this section we compare M\&MNets, and its variant with pseudo-pairs with several  baselines:
\begin{itemize}
\item \textbf{CycleGAN}. We adapt CycleGAN~\citep{zhu2017unpaired} to learn a mapping from depth to semantic segmentation (and vice versa) in a purely unpaired setting. In contrast to M\&MNets, this method only leverages depth and semantic segmentation, ignoring the available RGB data and the corresponding pairs (as shown in Figure~\ref{fig:cross-modal-settings}a).
\item \textbf{2$\times$pix2pix}. We adapt pix2pix~\citep{isola2016image} to learn two cross-modal translations from paired data (i.e. $D \rightarrow R$ and $R \rightarrow S$). The architecture uses skip connections (which are effective in this case since both translations are seen) and the corresponding modality-specific losses. 
We adapt the code from~\citep{isola2016image}. 
In contrast to ours, it requires explicit decoding to RGB, which may degrade the quality of the prediction.
\item \textbf{StarGAN}. We consider two adaptations of the StarGAN~\citep{choi2017stargan}. Both versions share the same network architecture for all modalities except for the first layer of the encoder and the last layer of of decoder which are modality-specific layers. This is required since modalities vary in the number of channels. The first version, called \emph{StarGAN(unpaired)}, uses the losses originally proposed in~\citep{choi2017stargan}. We also implement a version which exploits the paired data, which we call \emph{StarGAN(paired)}.  For this version, we removed the cycle consistency (which is not required for paired modalities). We found this to slightly improve results. 
\item \textbf{$D \rightarrow R \rightarrow S$} is similar to 2$\times$pix2pix but with the architecture  used in M\&MNets. We train a translation model from depth to RGB and from RGB to segmentation, and obtain the transformation depth-to-segmentation by concatenating them. Note that it also requires translating to intermediate RGB images.
\item \textbf{$S \rightarrow R \rightarrow D$} is analogous to the previous baseline.
\item \textbf{M\&MNets} is the original mix and match networks~\citep{wang2018mix}.
\item \textbf{M\&MNets+PP} is the variant of M\&MNets using pseudo-pairs.
\item \textbf{Oracle} is the upper bound obtained by training a translation fully supervised with paired data.
\end{itemize}
\begin{table*}[tb]
\setlength{\tabcolsep}{2.5pt}
\centering
\resizebox{\textwidth}{!}{
\begin{tabular}{=c+c+c|+c+c+c+c+c+c+c+c+c+c+c+c+c|+c+c}
\hline
Method & Conn. & $L_{SEM}$   & \rotatebox{90}{Bed}  &\rotatebox{90}{Book\;}   &\rotatebox{90}{Ceiling\;}   &\rotatebox{90}{Chair\;}   &\rotatebox{90}{Floor}  &\rotatebox{90}{Furniture}  &\rotatebox{90}{Object\;}   &\rotatebox{90}{Picture\;}  &\rotatebox{90}{Sofa\;} &\rotatebox{90}{Table\;} &\rotatebox{90}{TV\;} &\rotatebox{90}{Wall\;}   &\rotatebox{90}{Window\;}   
&\rotatebox{90}{mIoU\;}  &\rotatebox{90}{Global\;} 
\\
 & & &  \crule[u_Bed]{0.2cm}{0.2cm} &  \crule[u_Books]{0.2cm}{0.2cm} & \crule[u_Ceiling]{0.2cm}{0.2cm}& \crule[u_Chair]{0.2cm}{0.2cm} & \crule[u_Floor]{0.2cm}{0.2cm} &  \crule[u_Furniture]{0.2cm}{0.2cm} &  \crule[u_object]{0.2cm}{0.2cm} & \crule[u_Picture]{0.2cm}{0.2cm}& \crule[u_Sofa]{0.2cm}{0.2cm} & \crule[u_Table]{0.2cm}{0.2cm}& \crule[u_Tv]{0.2cm}{0.2cm}& \crule[u_wall]{0.2cm}{0.2cm} & \crule[u_Window]{0.2cm}{0.2cm}& &
\\
\hline
\textbf{Baselines} & & & & & & & & & & &&&&&&& \\
CycleGAN
& SC & CE &{2.79}    &{0.00}   &{16.9}  &{6.81}    &{4.48}    &{0.92}    &{7.43}   &{0.57}  &{9.48}    &{0.92}     &{0.31}    &{17.4}   &{15.1}  &{6.34}     &{14.2}   \\ 
2$\times$pix2pix
&SC &CE &34.6 &1.88 &70.9 &20.9 &63.6 &17.6 &14.1 &0.03 &\textbf{38.4} &10.0 &4.33 &67.7 &20.5  &25.4 &57.6
\\
StarGAN(unpaired)
&{PI} &{CE} &{6.71} &{1.42} &{17.6} &{6.21} &{13.2} &{1.25} &{8.51} &{0.52} &{12.8} &{3.24} &{4.28} &{9.52} &{8.57}  &{7.21} &{10.7}
\\
StarGAN(paired)
&{PI} &{CE} &{9.70} &{2.56} &{18.4} &{5.70} &{15.7} &{0.41} &{9.20} &{1.56} &{14.2} &{5.02} &{3.56} &{14.7} &{11.4}  &{8.62} &{14.1}
\\
M\&MNets $D\rightarrow R \rightarrow S$
& PI & CE &0.02 &0.00 &8.76 &0.10 &2.91 &2.06 &1.65 &0.19 &0.02 &0.28 &0.02 &58.2 &3.30 &5.96 &32.3 
\\ 
M\&MNets $D\rightarrow R \rightarrow S$
& SC & CE &25.4 &0.26 &82.7 &0.44 &56.6 &6.30 &23.6 &5.42 &0.54 &21.9 &10.0 &68.6 &19.6 &24.7 &59.7 
\\ 
\hline
\textbf{Zero-pair} & & & & & & & & &&&&&&& \\
M\&MNets $D \rightarrow S$
& PI & CE &{50.8}    &{18.9}   &{89.8}  &{31.6}   &\textbf{88.7}    &{48.3}    &{44.9}   &{62.1}  &{17.8}    &\textbf{49.9}     &{51.9}    &{86.2}   &{79.2} &55.4    &80.4
\\ 
M\&MNets+PP $D \rightarrow S$
& PI & CE &{52.1}    &{29.0}   &{88.6}  &{32.7}   &{86.9}    &\textbf{66.9}    &{48.4}   &{76.6}  &25.1    &{45.5}     &{58.8}    &{88.5}   &\textbf{82.0} &60.1    &82.2
\\ 
\hline
\textbf{Multi-modal} & & & & & & & & &&&&&&& \\
M\&MNets $\left( R,D\right) \rightarrow S$
& PI & CE &{49.9}    &{25.5}   &{88.2}  &{31.8}    &{86.8}    &{56.0}    &{45.4}   &{70.5}  &{17.4}    &{46.2}     &{57.3}    &{87.9}   &{79.8}  &{57.1}     &{81.2}   
\\
M\&MNets+PP $\left( R,D\right) \rightarrow S$
& PI & CE &\textbf{53.3}    &\textbf{35.7}   &\textbf{89.9}  &\textbf{37.0}    &{88.6}    &{59.3}    &\textbf{55.8}   &\textbf{76.9}  &25.7    &{46.6}     &\textbf{69.6}    &\textbf{89.5}   &{80.0}  &\textbf{62.2}     &\textbf{83.5}   
\\ 
\hline
\textbf{Oracle} & & & & & & & & &&&&&&& \\
$ D \rightarrow S$
& PI & CE &{53.7}    &{31.0}   &{89.1}  &{31.4}    &{88.2}    &{66.8}    &{52.7}   &{78.4}  &{25.7}    &{47.4}     &{59.3}    &{89.7}   &{82.2}  &{61.2}     &{83.4}   
\\
$\left( R,D\right) \rightarrow S$
& PI & CE &58.4    &40.8   &91.3  &41.6    &90.7    &61.5    &57.6   &80.9  &36.8    &51.6     &72.6    &{88.4}   &{83.1}  &{65.7}     &84.0   
\\ 
\hline
\end{tabular}
}
\caption{Zero-pair depth-to-segmentation translation on SceneNet RGB-D. \textbf{SC}: skip connections, \textbf{PI}: pooling indexes, \textbf{CE}: cross-entropy, \textbf{PP}: pseudo-pairs. x}
\label{table:depth2segm}
\end{table*}

Table~\ref{table:depth2segm} shows results for the different methods for depth-to-segmentation translation. CycleGAN is not able to learn a good mapping, showing the difficulty of unpaired translation to solve this complex cross-modal task. 2$\times$pix2pix manages to improve the results by resorting to the anchor modality RGB, although still not satisfactory since this sequence of translations does not enforce explicit alignment between depth and segmentation, and the first translation network may also discard information not relevant for the RGB task, but necessary for reconstructing the segmentation image (like in the ”Chinese whispers”/telephone game). Also, both results for $StarGAN$ show that this approach is unable to learn a good mapping between the unseen modalities.

M\&MNets evaluated on ($D \rightarrow R \rightarrow S$) achieve a similar result as CycleGAN, but significantly worse than 2$\times$pix2pix. However, when we run our architecture with skip connections we obtain results similar to 2$\times$pix2pix. Note that in this setting translations only involve seen encoders and decoders, so skip connections function well.  The direct combination ($D \rightarrow S$) with M\&MNets outperforms all baselines significantly. The performance more than doubles in terms of mIoU. Results improve another $5\%$ in mIoU when adding the pseudo-pairs during training. 

\begin{figure}[tb]
\centering
\includegraphics[width=0.99\columnwidth]{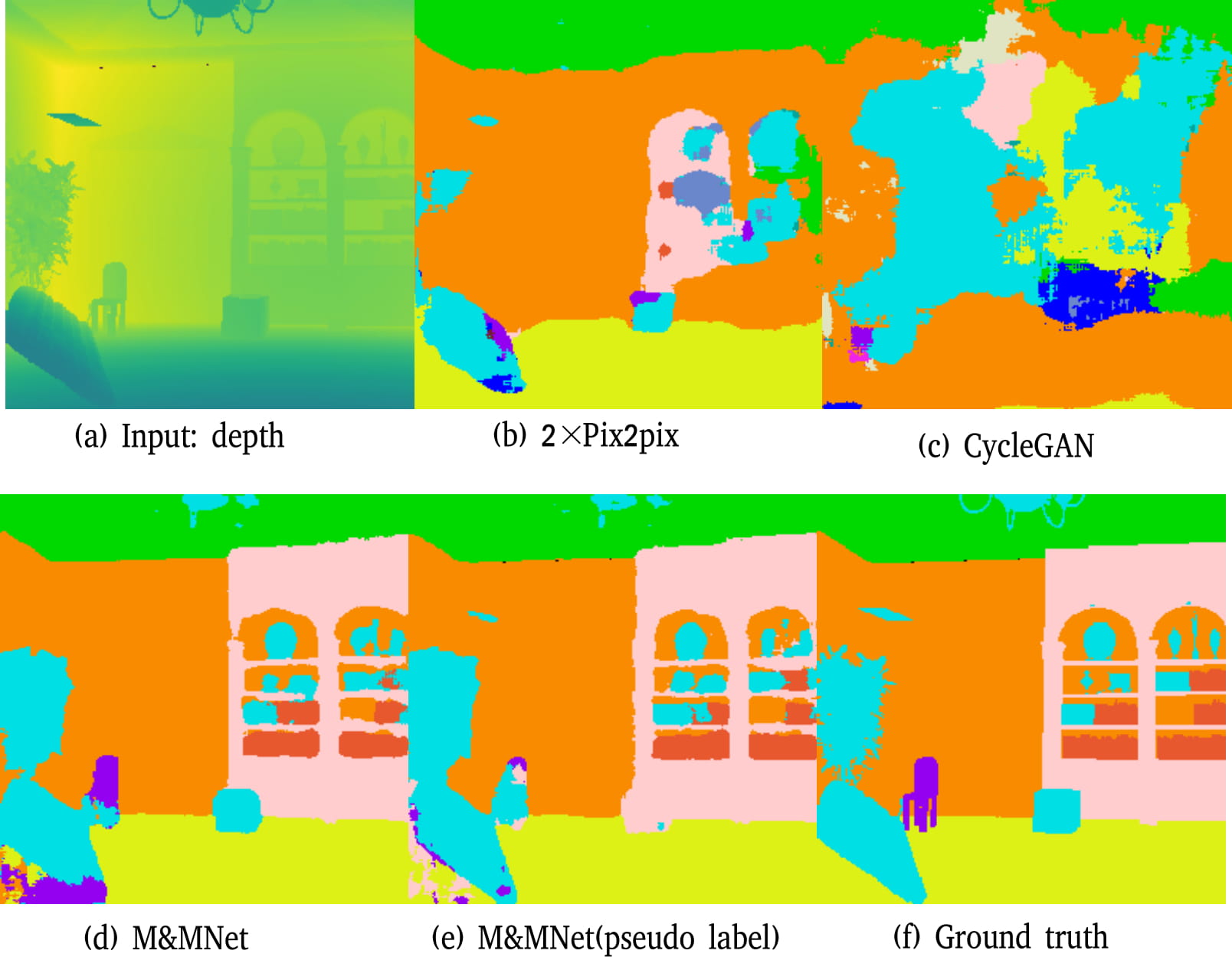}
\caption{
Zero-pair depth-to-segmentation translation on SceneNet RGB-D.}\label{fig:rgbds_comparison}
\end{figure}

Figure~\ref{fig:rgbds_comparison} shows a representative example of the differences between the evaluated methods. CycleGAN fails to recover any meaningful segmentation of the scene, revealing the difficulty to learn cross-modal translations without paired data. 2$\times$pix2pix manages to recover the layout and coarse segmentation, but fails to segment medium and small size objects. M\&MNets are able to obtain finer and more accurate segmentations.

\begin{figure}[tb]
\centering
     \includegraphics[width=0.99\columnwidth]{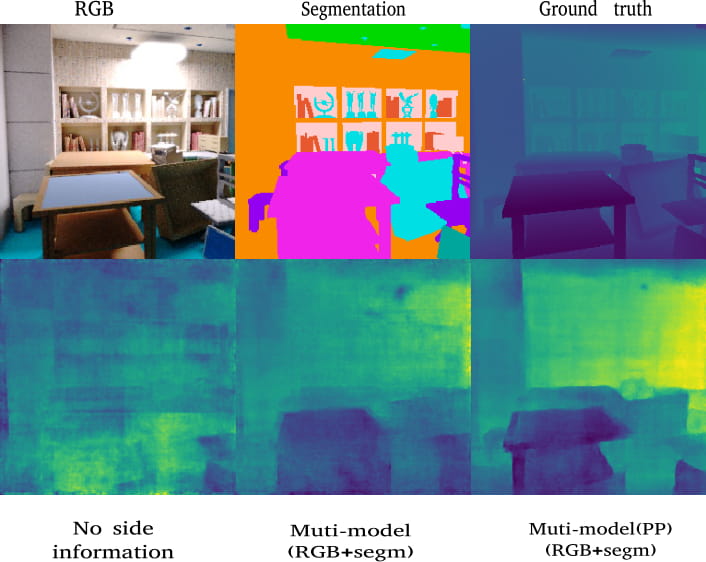}
\caption{
Zero-pair and multimodal segmentation-to-depth on SceneNet RGB-D.}\label{fig:example_segm-to-depth}
\end{figure}

Table~\ref{table:segm2depth} shows results when we test in the opposite direction from semantic segmentation to depth. The conclusions are similar as in previous experiment: M\&MNets outperform both baseline methods on all five evaluation metrics. Figure~\ref{fig:example_segm-to-depth} illustrates this case, showing how pooling indices are also key to obtain good depth images, compared with no side information at all. The variant with pseudo-pairs obtains the best results.

\begin{table}
\setlength{\tabcolsep}{1.5pt}
\resizebox{\columnwidth}{!}{
\begin{tabular}{=c|+c+c+c+c+c}
\hline
\multirow{2}{*}{Method} & \multicolumn{3}{c}{$\delta<$} &RMSE &RMSE \\
 & $1.25$ & $1.25^2$ & $1.25^3$ &(lin) &(log) \\
\hline
\textbf{Baselines} & & & & & \\
CycleGAN&{0.05}    &{0.12}   &{0.20}  &{4.63}    &{1.98}       \\ 
2$\times$pix2pix& {0.14} &{0.31} &{0.46} &{3.14}&{1.28} \\
 StarGAN(unpaired)&{0.05}    &{0.14}   &{0.23}  &{4.60}    &{1.96}\\
StarGAN(paired)&{0.07}    &{0.15}   &{0.26}  &{4.58}    &{1.94}\\
M\&MNets $S\rightarrow R \rightarrow D$ &{0.15}    &{0.30}   &{0.44}  &{3.24}    &{1.24}       \\
\hline
\textbf{Zero-pair} & & & & & \\
M\&MNets $S \rightarrow D$ &{0.33}    &{0.42}   &{0.59}  &{2.80}    &{0.67}       \\
M\&MNets+PP $S \rightarrow D$ &{0.42}    &{0.61}   &{0.79}  &{2.24}    &{0.60}       \\
\hline
\textbf{Multi-modal} & & & & & \\
M\&MNets $\left( R,S\right) \rightarrow D$
&{0.36}    &{0.48}   &{0.65}  &{2.48}    &{0.64}       \\
M\&MNets+PP $\left( R,S\right) \rightarrow D$
&\textbf{0.47}    &\textbf{0.69}   &\textbf{0.81}  &\textbf{1.98}    &\textbf{0.49}       \\
\hline
\textbf{Oracle} & & & & & \\
$ S \rightarrow D$
&{0.49}    &{0.72}   &{0.85}  &{1.94}    &{0.43}       \\
$\left( R,S\right) \rightarrow D$
&{0.51}    &{0.76}   &{0.90}  &{1.79}    &{0.29}       \\
\hline
\end{tabular}
}
  \caption{Zero-pair segmentation-to-depth on SceneNet RGB-D.}\label{table:segm2depth}%
\end{table}

\subsubsection{Multi-modal translation}
Since features from different modalities are aligned, we can also use M\&MNets for multi-modal translation. For instance, in the previous multi-modal setting, given the RGB and depth images of the same scene we can translate to segmentation. We simply combine both modality-specific latent features $x$ and $z$ using a weighted average $y=\left(1-\alpha\right)x+\alpha z$, where $\alpha$ controls the weight of each modality. We set $\alpha=0.2$ and use the pooling indices from the RGB encoder (instead of those from depth encoder). The resulting feature $y$ is then decoded using the segmentation decoder. We proceed analogously to translation from RGB and segmentation to depth. The results in Table~\ref{table:depth2segm} and Table~\ref{table:segm2depth} show that this multi-modal combination further improves the performance of zero-pair translation, as the example in Figure~\ref{fig:example_segm-to-depth} illustrates. 

\subsection{Experiments on SUN RGB-D}
\begin{figure}[t]
\centering
\includegraphics[width=1\columnwidth]{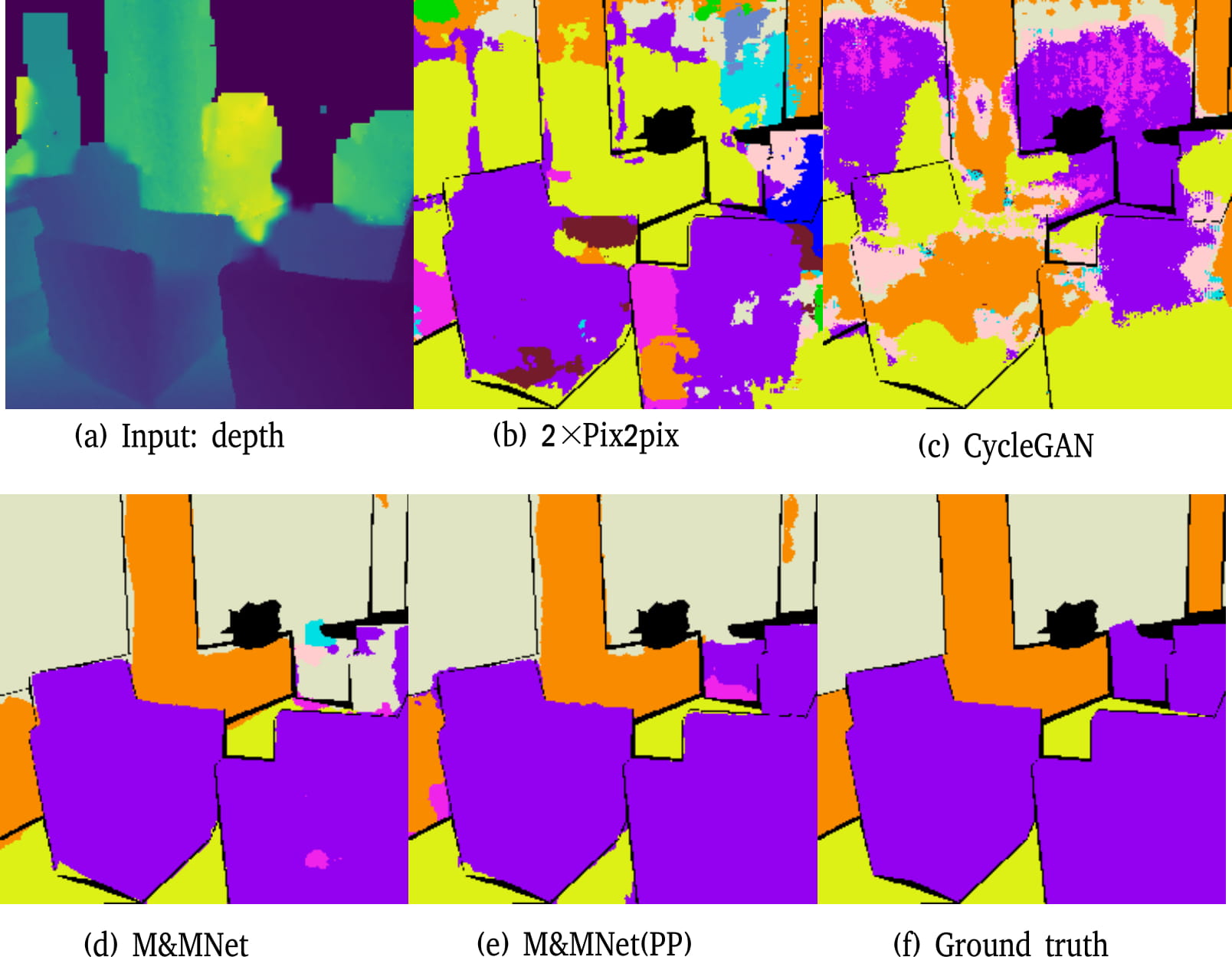}
\caption{\label{fig:example_depth-to-segm_sunrgbd} Example of zero-pair depth-to-segmentation  on SUN RGB-D.}
\end{figure}

\begin{figure}[t]
\centering
\includegraphics[width=1\columnwidth]{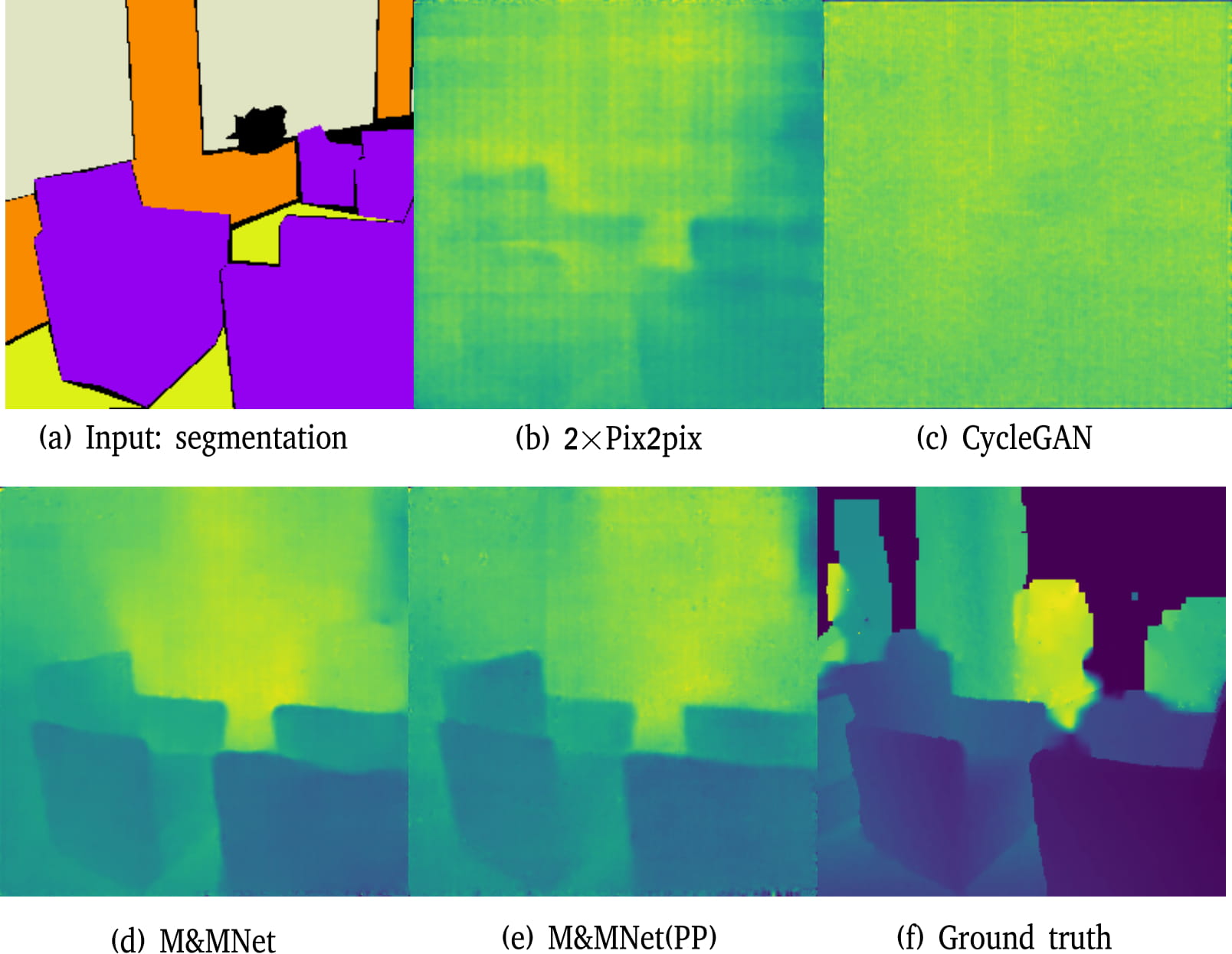}
\caption{\label{fig:example_segm-to-depth_sunrgbd} Example of zero-pair segmentation-to-depth on SUN RGB-D.}
\end{figure}

The previous results were obtained on the SceneNet RGB-D dataset which consists of synthetic images. Here we also show that M\&MNets can be effective for the more challenging dataset SUN RGB-D, which involves real images and more limited data. The results in Table~\ref{table:depth2segm_sunrgbd} and Table~\ref{table:segm2depth_sunrgbd} show that M\&MNets consistently outperform the other baselines in both unseen translation directions, with the new variant with pseudo-pairs obtaining the best performance. Similarly, multi-modal translation further improves the performance. Figures~\ref{fig:example_depth-to-segm_sunrgbd} and \ref{fig:example_segm-to-depth_sunrgbd} illustrate how the proposed methods can reconstruct more reliably the target modality, especially the finer details.

\begin{figure}[t]
\centering
\includegraphics[width=1\columnwidth]{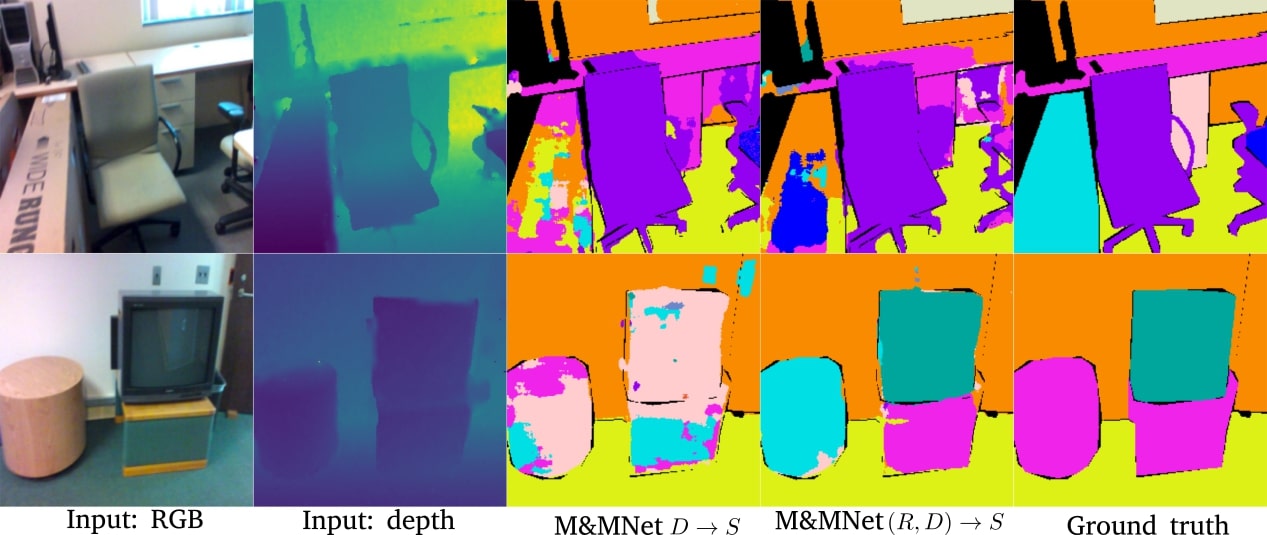}
\caption{\label{fig:example_depth-to-segm_failure_case_sunrgbd} Failure cases of the proposed framework on SUN RGB-D. See text for discussion.}\label{fig_failure}
\end{figure}

The results also show that the depth cue is insufficient to detect some of the classes such as \textit{Book} and \textit{TV}. The oracle results show that this is also the case when you have access to depth-semantic segmentation pairs. The results also show that our multi-modal results are biased towards RGB: this is reflected in the bad results which are obtained for the class \textit{bed} which is well detected in the depth modality but not in the RGB modality, and also not by our multi-modal system. Examples of these cases are provided in Fig.~\ref{fig_failure}.

\begin{table*}[tb]
\setlength{\tabcolsep}{2.5pt}
\centering
\resizebox{\textwidth}{!}{
\begin{tabular}{=c+c+c|+c+c+c+c+c+c+c+c+c+c+c+c+c|+c+c}
\hline
Method & Conn. & $L_{SEM}$   & \rotatebox{90}{Bed}  &\rotatebox{90}{Book\;}   &\rotatebox{90}{Ceiling\;}   &\rotatebox{90}{Chair\;}   &\rotatebox{90}{Floor}  &\rotatebox{90}{Furniture}  &\rotatebox{90}{Object\;}   &\rotatebox{90}{Picture\;}  &\rotatebox{90}{Sofa\;} &\rotatebox{90}{Table\;} &\rotatebox{90}{TV\;} &\rotatebox{90}{Wall\;}   &\rotatebox{90}{Window\;}   
&\rotatebox{90}{mIoU\;}  &\rotatebox{90}{Global\;} 
\\
 & & &  \crule[u_Bed]{0.2cm}{0.2cm} &  \crule[u_Books]{0.2cm}{0.2cm} & \crule[u_Ceiling]{0.2cm}{0.2cm}& \crule[u_Chair]{0.2cm}{0.2cm} & \crule[u_Floor]{0.2cm}{0.2cm} &  \crule[u_Furniture]{0.2cm}{0.2cm} &  \crule[u_object]{0.2cm}{0.2cm} & \crule[u_Picture]{0.2cm}{0.2cm}& \crule[u_Sofa]{0.2cm}{0.2cm} & \crule[u_Table]{0.2cm}{0.2cm}& \crule[u_Tv]{0.2cm}{0.2cm}& \crule[u_wall]{0.2cm}{0.2cm} & \crule[u_Window]{0.2cm}{0.2cm}& &
\\
\hline
\textbf{Baselines} & & & & & & & & & & &&&&&&& \\
CycleGAN
& SC & CE &{0.00}    &{0.00}   &{0.00}  &{17.9}    &{46.9}    &{1.67}    &{4.59}   &{0.00}  &{0.00}    &{18.9}     &{0.00}    &{29.6}   &{25.4}  &{11.1}     &{26.3}   \\ 
2$\times$pix2pix
&SC &CE &{3.88} &{0.00} &{12.4} &{29.6} &{57.1} &{17.2} &{13.0} &{35.4} &{8.07} &{35.1} &{0.00} &{47.0} &{7.73} &{20.5}  &{38.6} \\
StarGAN(unpaired)
&PI &CE &{0.00} &{0.00} &{2.45} &{15.8} &{33.6} &{5.73} &{6.28} &{0.57} &{0.00} &{6.25} &{0.00} &{28.4} &{26.9} &{9.69}  &{20.6}\\
StarGAN(paired)
&PI &CE &{0.00} &{0.00} &{2.01} &{20.2} &{38.9} &{4.12} &{5.78} &{0.31} &{0.00} &{7.30} &{0.00} &{31.5} &{30.7} &{10.8}  &{23.8}\\
M\&MNets $D\rightarrow R \rightarrow S$
& PI & CE &{0.00}    &{0.00}   &{0.00}  &{17.0}    &{39.4}    &{0.52}    &{0.01}   &{0.00}  &{0.01}    &{12.2}     &{0.00}    &{31.0}   &{5.19}  &{8.12}     &{22.8}   \\ 
M\&MNets $D\rightarrow R \rightarrow S$
& SC & CE &\textbf{39.9}    &{0.25}   &{15.2}  &{37.6}    &{58.0}    &{19.0}    &{11.7}   &{2.45}  &{4.82}    &{36.9}     &{0.00}    &{46.8}   &{12.3}  &{21.9}     &{40.6}
\\ 
\hline
\textbf{Zero-pair} & & & & & & & & &&&&&&& \\
M\&MNets $D \rightarrow S$
& PI & CE &{28.4}    &{2.90}   &{22.6}  &{41.9}   &{71.6}    &{14.1}    &{25.1}   &{17.8}  &\textbf{11.8}    &{49.7}     &{0.08}    &{64.2}   &{15.5} &{28.1}    &{51.8}
\\ 
M\&MNets+PP $D \rightarrow S$
& PI & CE &{29.8}    &{4.52}   &\textbf{28.5}  &{44.1}   &{73.3}    &{17.2}    &{27.5}   &{20.1}  &{9.81}    &{53.4}     &{0.14}    &{67.5}   &{17.9} &{30.2}    &{54.2}
\\ 
\hline
\textbf{Multi-modal} & & & & & & & & &&&&&&& \\
M\&MNets $\left( R,D\right) \rightarrow S$
& PI & CE &{0.00}    &{16.6}   &{21.4}  &\textbf{56.0}    &{72.1}    &{24.2}    &{28.3}   &{38.1}  &\textbf{21.7}    &{57.0}     &{64.6}    &{68.0}   &{43.7}  &{39.4}     &{58.8}   
\\ 
M\&MNets+PP $\left( R,D\right) \rightarrow S$
& PI & CE &{0.10}    &\textbf{19.3}   &{25.5}  &{54.6}    &\textbf{74.6}    &\textbf{25.6}    &\textbf{30.1}   &\textbf{42.4}  &{21.0}    &\textbf{58.1}     &\textbf{65.2}    &\textbf{69.0}   &\textbf{49.7}  &\textbf{41.1}     &\textbf{59.8}   
\\ 
\hline
\textbf{Oracle} & & & & & & & & &&&&&&& \\
$ D \rightarrow S$
& PI & CE &{32.6}    &{8.01}   &{36.5}  &{56.8}    &{84.7}    &{20.4}    &{31.4}   &{19.7}  &{8.75}    &{61.7}     &{1.60}    &{72.1}   &{21.2}  &{35.1}     &{62.3}   
\\ 
$\left( R,D\right) \rightarrow S$
& PI & CE &{0.13}    &{21.2}   &{26.4}  &{56.2}    &{78.9}    &{26.9}    &{35.2}   &{44.4}  &{23.2}    &{60.2}     &{67.3}    &{71.2}   &{52.3}  &{43.3}     &{62.5}   
\\ 
\hline
\end{tabular}
}
\caption{Zero-pair depth-to-semantic segmentation on SUN RGB-D. \textbf{SC}: skip connections, \textbf{PI}: pooling indexes, \textbf{CE}: cross-entropy, \textbf{PP}: pseudo-pairs.}
\label{table:depth2segm_sunrgbd}
\end{table*}

\begin{table}
\setlength{\tabcolsep}{2.5pt}
\resizebox{\columnwidth}{!}{
\begin{tabular}{=c|+c+c+c+c+c}
\hline
\multirow{2}{*}{Method} & \multicolumn{3}{c}{$\delta<$} &RMSE &RMSE \\
 & $1.25$ & $1.25^2$ & $1.25^3$ &(lin) &(log) \\
\hline
\textbf{Baselines} & & & & & \\
CycleGAN &{0.06}    &{0.13}   &{0.24}  &{4.80}    &{1.57}       \\ 
2$\times$pix2pix & {0.13} &{0.34} &{0.59} &{3.80}&{1.30} \\
StarGAN(unpaired)&{0.06}    &{0.12}   &{0.22}  &{5.04}    &{1.59}  \\
StarGAN(paired)&{0.07}    &{0.15}   &{0.27}  &{4.60}    &{1.55}  \\
M\&MNets $S\rightarrow R \rightarrow D$ &{0.12}    &{0.35}   &{0.62}  &{3.90}    &{1.36}       \\
\hline
\textbf{Zero-pair} & & & & & \\
M\&MNets $S \rightarrow D$ &{0.45}    &{0.66}   &{0.78}  &{1.75}    &{0.53}       \\
M\&MNets+PP $S \rightarrow D$&{0.49}    &{0.77}   &{0.90}  &{1.42}    &{0.37}        \\
\hline
\textbf{Multi-modal} & & & & & \\
M\&MNets $\left( R,S\right) \rightarrow D$
&{0.53}    &{0.80}   &{0.92}  &{1.63}    &{0.35}       \\
M\&MNets+PP $\left( R,S\right) \rightarrow D$
&\textbf{0.56}    &\textbf{0.83}   &\textbf{0.93}  &\textbf{1.33}    &\textbf{0.34}       \\
\hline
\textbf{Oracle} & & & & & \\
$ S \rightarrow D$
&{0.61}    &{0.88}   &{0.97}  &{1.20}    &{0.30}       \\
$\left( R,S\right) \rightarrow D$
&{0.64}    &{0.92}   &{0.98}  &{0.98}    &{0.27}       \\
\hline
\end{tabular}
}
  \caption{Zero-pair semantic segmentation-to-depth on SUN RGB-D.}\label{table:segm2depth_sunrgbd}%
\end{table}

\subsection{Experiments on four modalities}

\begin{figure}
    \centering
    \includegraphics[width=0.9\columnwidth]{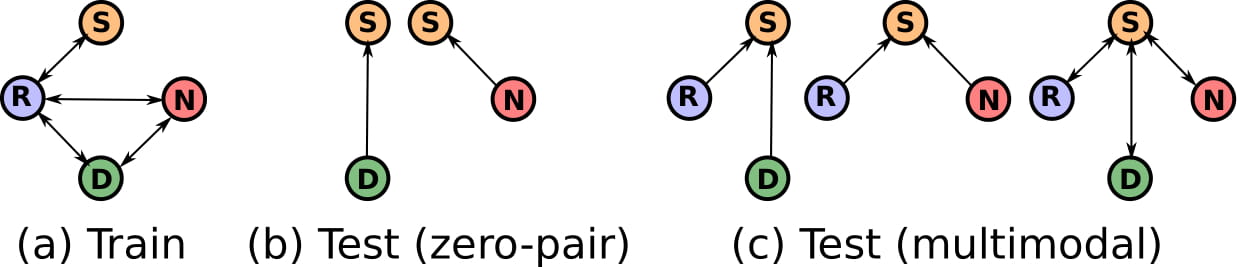}
   \caption{Cross-modal translations in the Freiburg Forest dataset experiment: (a) training, (b) test (zero-shot) and (c) test (multimodal). We show only translations to semantic segmentation for simplicity.}\label{fig:forest_translations}
\end{figure}

As an example of zero-pair translation for an application with more than three modalities we perform experiments on the Freiburg Forest dataset which contains the RGB, depth, NIR and semantic segmentation modalities. For the training we use the settings used in the previous experiments, and add a $Berhu$ loss (see Eq.~\ref{eq_berhu}) for NIR in this experiment.

In the provided dataset all modalities are recorded for all scenes, however we consider that we have pairs for RGB and semantic segmentation, and we have a non-overlapping dataset of triplets for RGB, Depth, and NIR (see Figure~\ref{fig:forest_translations}). This scenario could be considered realistic. It reflects a situation where initially the robot only has an RGB camera, and labellers have provided semantic segmentation maps for these images. Then two additional sensors are added later to the robot, but no segmentation maps are available for this newly obtained multi-modal data. 

\begin{figure}[t]
\includegraphics[width=1\columnwidth]{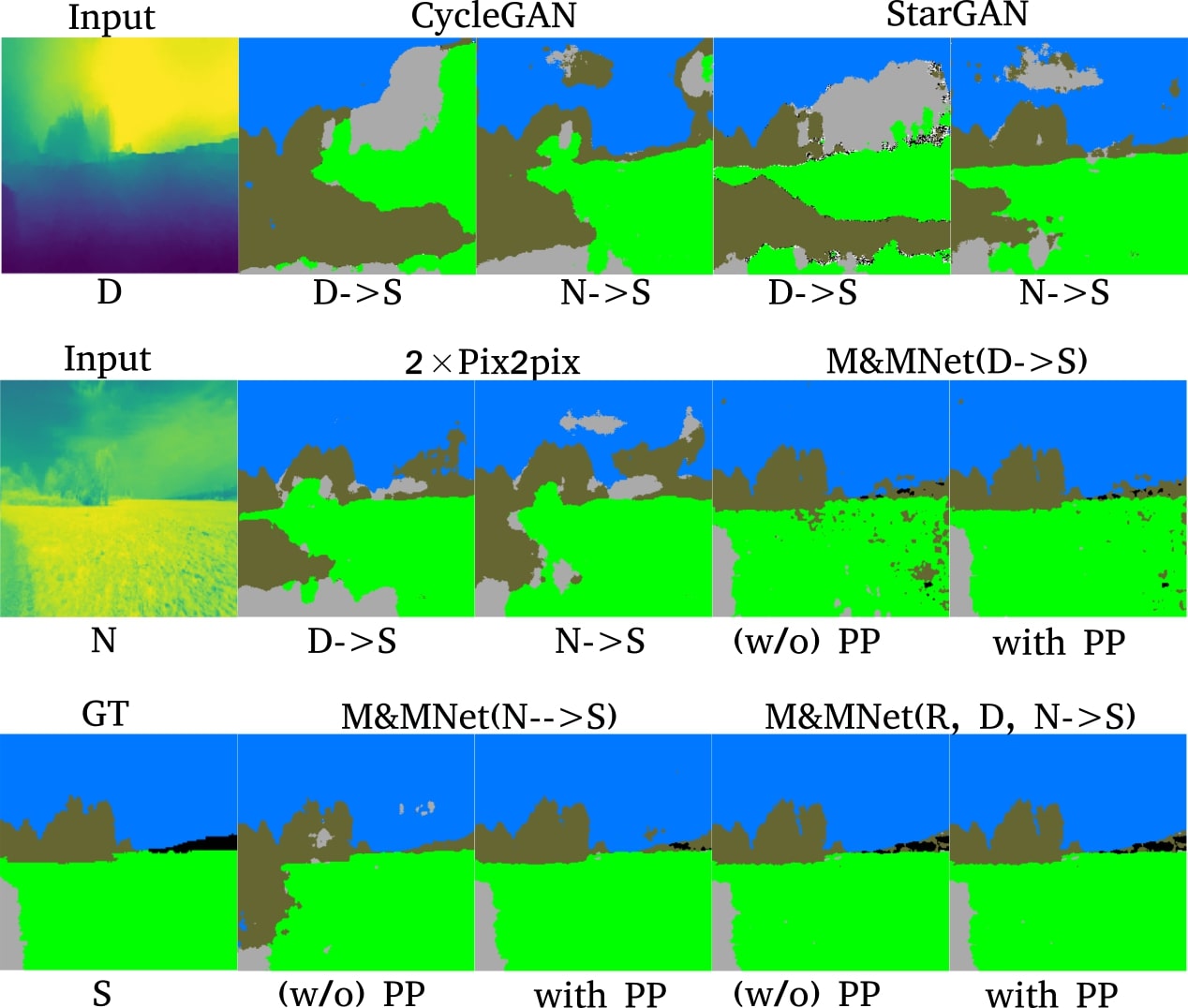}
\caption{\label{fig:example_nir_depth_to_seg_freiburg_forest} Example of zero-pair translations. R: RGB, D: depth, N: NIR, S: semantic segmentation, and PP: pseudo-pairs. }
\end{figure}

As we can see in Table~\ref{table:nir_depth2segm_freiburg_forest}, our method  achieves the best scores. In the case of zero-pair setting (M\&MNets $D \rightarrow S$, M\&MNets $N \rightarrow S$, M\&MNets+PP $D \rightarrow S$ and M\&MNets+PP $N \rightarrow S$) the results obtain a large gap when compared to the baselines, clearly demonstrating the superiority of our method. For example, for  $N \rightarrow S$ we obtain an increase of $22\%$ over $2\times $pix2pix. The multi-modal results show that adding more modalities further increases results. Mainly, the performance on the category \textit{obstacle} increases.   Figure~\ref{fig:example_nir_depth_to_seg_freiburg_forest} shows representative examples of the different methods. The conclusions are similar to previous experiments: we effectively conduct cross-modal translation with zero-pair data and pseudo-labeling further improves the results. 

\section{Conclusions}
We have introduced mix and match networks as a framework to perform image translations between unseen modalities by leveraging the knowledge learned from seen translations with explicit training data. The key challenge lies in aligning the latent representations in the bottlenecks in such a way that any encoder-decoder combination is able to perform effectively its corresponding translation. M\&MNets have advantages in terms of scalability since only seen translations need to be trained. We also introduced zero-pair cross-modal translation, a challenging scenario involving three modalities and paired seen and unseen translations. 
In order to effectively address this problem, we described several tools to enforce the alignment of latent representations, including autoencoders, latent consistency losses, and robust side information. In particular, our results show that side information is critical to perform satisfactory cross-modal translations, but conventional side information such as skip connections may not work properly with unseen translations. We found that pooling indices are more robust and invariant, and provide helpful hints to guide the reconstruction of spatial structure.

We also analyzed a specific limitation of the original M\&MNets~\citep{wang2018mix} in the zero-pair setting, which is that a significant part of the shared features between unseen modalities is not exploited. We proposed a variant that generates pseudo-pairs to enforce the networks to use more information between unseen modalities, even when that information is not shared by seen translations. The effectiveness of M\&MNets with pseudo-pairs has been evaluated in several multi-modal datasets.

A potential limitation of our system is that we work with separate encoder and decoders for each modality. Some recent cross-domain image translators such as StarGAN~\citep{choi2017stargan} and SDIT~\citep{wang2019sdit} use a single shared encoder and a single shared decoder. In that spirit, it could be possible to have partially shared encoders and decoders between different modalities. However,  modality-specific layers would be still required in more challenging cross-modal translation.

\begin{table}[tb]
\setlength{\tabcolsep}{2.5pt}
\centering
\resizebox{\columnwidth}{!}{
\begin{tabular}{cc|ccccc|cc}

\hline
Method & Conn.  & \rotatebox{90}{Sky\;}
&\rotatebox{90}{Trail\;}   &\rotatebox{90}{Grass\;}   &\rotatebox{90}{Vegetation\;}  &\rotatebox{90}{Obstacle\;}  
&\rotatebox{90}{mIoU\;}  &\rotatebox{90}{Global\;}\\
 & &  \crule[f_sky]{0.2cm}{0.2cm} &  \crule[f_trail]{0.2cm}{0.2cm} & \crule[f_grass]{0.2cm}{0.2cm}& \crule[f_vegetation]{0.2cm}{0.2cm} & \crule[f_obstacle]{0.2cm}{0.2cm} & &
\\
\hline
\textbf{Baselines} &  & & & &&&& \\
CycleGAN  $D \rightarrow S$
& SC  &{36.3}    &{31.7}    &{19.2}    &{24.5}   &{5.40}  &{23.4}     &{26.2}   \\ 
CycleGAN  $N \rightarrow S$
& SC  &{37.2}    &{34.1}    &{18.4}    &{29.5}   &{0.41}  &{23.9}     &{28.5}   \\
2$\times$pix2pix$D \rightarrow S$
&SC  &{72.9} &{32.2} &{45.7} &{67.9} &{30.9}  &{49.9} &{59.9} \\
2$\times$pix2pix$N \rightarrow S$ 
&SC  &{78.6} &{43.2} &{53.4} &{74.4} &{18.6}  &{53.6} &{66.8} \\
StarGAN$D \rightarrow S$ 
&PI  &{45.2} &{28.1} &{24.4} &{21.5} &{1.36}  &{24.1} &{28.1} \\
StarGAN  $N \rightarrow S$ 
&PI &{31.2} &{15.1} &{29.4} &{23.2} &{10.7}  &{21.9} &{25.8} \\
M\&MNets $D\rightarrow R \rightarrow S$
& PI &{45.3}    &{19.6}    &{25.4}    &{35.5}   &{25.3}    &{30.0}     &{33.5}   \\ 
M\&MNets $N\rightarrow R \rightarrow S$
& PI   &{58.1}    &{34.1}    &{32.4}    &{42.4}     &{12.3}  &{35.8}     &{42.4}
\\ 
\hline
\textbf{Zero-pair} & & & &&&&& \\
M\&MNets $D \rightarrow S$
& PI   &{89.0}   &{71.8}    &{71.3}    &{82.7}   &{43.7}   &{71.6}    &{80.0}
\\ M\&MNets $N \rightarrow S$
& PI &{88.1}   &{78.1}    &{73.4}    &{83.1}   &{41.0}   &{72.7}    &{81.0}
\\ 
M\&MNets+PP $D \rightarrow S$
& PI   &{89.7}    &{75.4}    &{72.4}   &{83.6}  &{45.7}  &{73.4}    &{81.1}
\\ 
M\&MNets+PP $N \rightarrow S$
& PI   &{89.9}    &{80.1}    &{76.9}   &{85.5}  &{44.2}  &{75.3}    &{83.5}
\\ 
\hline
\textbf{Multi-modal} & & & &&&&& \\
M\&MNets $\left( R,D\right) \rightarrow S$
& PI  &{91.2}    &{84.5}    &{85.4}    &{89.1}   &{50.3}   &{80.1}     &{88.0}   
\\ 
M\&MNets $\left( R,N\right) \rightarrow S$
& PI  &{91.0}    &{83.5}    &{85.3}    &{90.0}   &{52.9}   &{80.5}     &{88.3}   
\\ 
M\&MNets $\left( R,D,N\right) \rightarrow S$
& PI  &{91.2}    &{84.2}    &{85.8}    &{90.1}   &{58.2}   &{81.8}     &{88.5}   
\\ 
M\&MNets+PP $\left( R,D\right) \rightarrow S$
& PI   &{90.9}    &{83.9}    &{85.0}    &{88.7}   &{59.5}   &{81.6}     &{88.1}   
\\
M\&MNets+PP $\left( R,N\right) \rightarrow S$
& PI &\textbf{91.7}    &{85.4}    &{86.1}    &{89.9}   &{58.2}   &{82.2}     &{88.6}   
\\
M\&MNets+PP $\left( R,D,N\right) \rightarrow S$
& PI   &{91.5}    &\textbf{85.8}    &\textbf{86.6}    &\textbf{90.6}   &\textbf{60.3}   &\textbf{83.0}     &\textbf{89.3} 
\\ 
\hline
\textbf{Oracle} & & & &&&&& \\
$D \rightarrow S$
& PI   &{89.5}    &{75.4}    &{80.4}    &{81.2}   &{54.7}   &{76.2}     &{82.2}   
\\
$N \rightarrow S$
& PI   &{90.2}    &{81.5}    &{83.6}    &{85.2}   &{50.4}   &{78.2}     &{85.4}   
\\
$\left( R, D, N\right) \rightarrow S$
& PI  &{{91.9}}    &{85.7}    &{87.9}    &{90.1}   &{64.9}   &{84.1}     &{89.4}   
\\
\hline
\end{tabular}
}
\caption{Zero-pair ({NIR}, depth)-to-semantic segmentation on Freiburg Forest. \textbf{SC}: skip connections, \textbf{PI}: pooling indexes, \textbf{PP}: pseudo-pairs.}
\label{table:nir_depth2segm_freiburg_forest}
\end{table}

\begin{acknowledgements}
The Titan Xp used for this research was donated by the NVIDIA Corporation. We acknowledge  the Spanish projects TIN2016-79717-R and RTI2018-102285-A-I00, the CHISTERA project M2CR (PCIN-2015-251) and the CERCA Programme / Generalitat de Catalunya. Herranz also acknowledges the European Union’s H2020 research under Marie Sklodowska-Curie grant No. 665919. Yaxing Wang acknowledges the Chinese Scholarship Council (CSC) grant No. 201507040048. .

\end{acknowledgements}

\bibliographystyle{spbasic}      %
\bibliography{longstrings,refs}

\begin{thebibliography}{90}
\providecommand{\natexlab}[1]{#1}
\providecommand{\url}[1]{{#1}}
\providecommand{\urlprefix}{URL }
\expandafter\ifx\csname urlstyle\endcsname\relax
  \providecommand{\doi}[1]{DOI~\discretionary{}{}{}#1}\else
  \providecommand{\doi}{DOI~\discretionary{}{}{}\begingroup
  \urlstyle{rm}\Url}\fi
\providecommand{\eprint}[2][]{\url{#2}}

\bibitem[{Akata et~al.(2016)Akata, Perronnin, Harchaoui, and
  Schmid}]{akata2016label}
Akata Z, Perronnin F, Harchaoui Z, Schmid C (2016) Label-embedding for image
  classification. {IEEE} Transactions on Pattern Analysis and Machine
  Intelligence 38(7):1425--1438

\bibitem[{Alharbi et~al.(2019)Alharbi, Smith, and Wonka}]{alharbi2019latent}
Alharbi Y, Smith N, Wonka P (2019) Latent filter scaling for multimodal
  unsupervised image-to-image translation. In: Proceedings of the IEEE
  Conference on Computer Vision and Pattern Recognition, pp 1458--1466

\bibitem[{Almahairi et~al.(2018)Almahairi, Rajeswar, Sordoni, Bachman, and
  Courville}]{almahairi2018augmented}
Almahairi A, Rajeswar S, Sordoni A, Bachman P, Courville A (2018) Augmented
  cyclegan: Learning many-to-many mappings from unpaired data. International
  Conference on Machine Learning

\bibitem[{Amodio and Krishnaswamy(2019)}]{Amodio_2019_CVPR}
Amodio M, Krishnaswamy S (2019) Travelgan: Image-to-image translation by
  transformation vector learning. In: Proceedings of the IEEE Conference on
  Computer Vision and Pattern Recognition

\bibitem[{Anoosheh et~al.(2018)Anoosheh, Agustsson, Timofte, and
  Van~Gool}]{Anoosheh_2018}
Anoosheh A, Agustsson E, Timofte R, Van~Gool L (2018) Combogan: Unrestrained
  scalability for image domain translation. 2018 IEEE/CVF Conference on
  Computer Vision and Pattern Recognition Workshops (CVPRW)
  \doi{10.1109/cvprw.2018.00122},
  \urlprefix\url{http://dx.doi.org/10.1109/CVPRW.2018.00122}

\bibitem[{Badrinarayanan et~al.(2015)Badrinarayanan, Handa, and
  Cipolla}]{badrinarayanan2015segnet}
Badrinarayanan V, Handa A, Cipolla R (2015) Segnet: A deep convolutional
  encoder-decoder architecture for robust semantic pixel-wise labelling.
  Proceedings of the IEEE Conference on Computer Vision and Pattern Recognition

\bibitem[{Cadena et~al.(2016)Cadena, Dick, and Reid}]{cadena2016multi}
Cadena C, Dick AR, Reid ID (2016) Multi-modal auto-encoders as joint estimators
  for robotics scene understanding. In: Robotics: Science and Systems

\bibitem[{Castrejon et~al.(2016)Castrejon, Aytar, Vondrick, Pirsiavash, and
  Torralba}]{castrejon2016learning}
Castrejon L, Aytar Y, Vondrick C, Pirsiavash H, Torralba A (2016) Learning
  aligned cross-modal representations from weakly aligned data. In: Proceedings
  of the IEEE Conference on Computer Vision and Pattern Recognition, pp
  2940--2949

\bibitem[{Chen et~al.(2018)Chen, Papandreou, Kokkinos, Murphy, and
  Yuille}]{chen2018deeplab}
Chen LC, Papandreou G, Kokkinos I, Murphy K, Yuille AL (2018) Deeplab: Semantic
  image segmentation with deep convolutional nets, atrous convolution, and
  fully connected crfs. {IEEE} Transactions on Pattern Analysis and Machine
  Intelligence 40(4):834--848

\bibitem[{Chen and Koltun(2017)}]{chen2017photographic}
Chen Q, Koltun V (2017) Photographic image synthesis with cascaded refinement
  networks. Proceedings of the International Conference on Computer Vision

\bibitem[{Chen et~al.(2017)Chen, Liu, Cheng, and Li}]{chen2017teacher}
Chen Y, Liu Y, Cheng Y, Li VO (2017) A teacher-student framework for
  zero-resource neural machine translation. arXiv preprint arXiv:170500753

\bibitem[{Chen et~al.(2019)Chen, Xu, Tian, and Jia}]{chen2019homomorphic}
Chen YC, Xu X, Tian Z, Jia J (2019) Homomorphic latent space interpolation for
  unpaired image-to-image translation. In: Proceedings of the IEEE Conference
  on Computer Vision and Pattern Recognition, pp 2408--2416

\bibitem[{Cheng et~al.(2016)Cheng, Zhao, Cai, Li, Huang, Rui
  et~al.}]{cheng2016semi}
Cheng Y, Zhao X, Cai R, Li Z, Huang K, Rui Y, et~al. (2016) Semi-supervised
  multimodal deep learning for {RGB-D} object recognition. Proceedings of the
  International Joint Conference on Artificial Intelligence

\bibitem[{Cho et~al.(2019)Cho, Choi, Park, Shin, and Choo}]{Cho_2019_CVPR}
Cho W, Choi S, Park DK, Shin I, Choo J (2019) Image-to-image translation via
  group-wise deep whitening-and-coloring transformation. In: The IEEE
  Conference on Computer Vision and Pattern Recognition (CVPR)

\bibitem[{Choi et~al.(2018)Choi, Choi, Kim, Ha, Kim, and
  Choo}]{choi2017stargan}
Choi Y, Choi M, Kim M, Ha JW, Kim S, Choo J (2018) Stargan: Unified generative
  adversarial networks for multi-domain image-to-image translation. In:
  Proceedings of the IEEE Conference on Computer Vision and Pattern Recognition

\bibitem[{Deng et~al.(2009)Deng, Dong, Socher, Li, Li, and
  Fei-Fei}]{imagenet_cvpr09}
Deng J, Dong W, Socher R, Li LJ, Li K, Fei-Fei L (2009) {ImageNet: A
  Large-Scale Hierarchical Image Database}. In: Proceedings of the IEEE
  Conference on Computer Vision and Pattern Recognition

\bibitem[{Eigen and Fergus(2015)}]{eigen2015predicting}
Eigen D, Fergus R (2015) Predicting depth, surface normals and semantic labels
  with a common multi-scale convolutional architecture. In: Proceedings of the
  International Conference on Computer Vision, pp 2650--2658

\bibitem[{Eitel et~al.(2015)Eitel, Springenberg, Spinello, Riedmiller, and
  Burgard}]{eitel2015multimodal}
Eitel A, Springenberg JT, Spinello L, Riedmiller M, Burgard W (2015) Multimodal
  deep learning for robust rgb-d object recognition. In: Proceedings of the
  {IEEE/RSJ} Conference on Intelligent Robots and Systems, IEEE, pp 681--687

\bibitem[{Fergus et~al.(2010)Fergus, Bernal, Weiss, and
  Torralba}]{fergus2010semantic}
Fergus R, Bernal H, Weiss Y, Torralba A (2010) Semantic label sharing for
  learning with many categories. Proceedings of the European Conference on
  Computer Vision pp 762--775

\bibitem[{Firat et~al.(2016)Firat, Cho, and Bengio}]{firat2016multi}
Firat O, Cho K, Bengio Y (2016) Multi-way, multilingual neural machine
  translation with a shared attention mechanism. arXiv preprint arXiv:160101073

\bibitem[{Fu et~al.(2017)Fu, Xiang, Jiang, Xue, Sigal, and Gong}]{fu2017recent}
Fu Y, Xiang T, Jiang YG, Xue X, Sigal L, Gong S (2017) Recent advances in
  zero-shot recognition. arXiv preprint arXiv:171004837

\bibitem[{Ganin and Lempitsky(2015)}]{ganin2015unsupervised}
Ganin Y, Lempitsky V (2015) Unsupervised domain adaptation by backpropagation.
  In: International Conference on Machine Learning, pp 1180--1189

\bibitem[{Gatys et~al.(2016)Gatys, Ecker, and Bethge}]{gatys2016image}
Gatys LA, Ecker AS, Bethge M (2016) Image style transfer using convolutional
  neural networks. In: Proceedings of the IEEE Conference on Computer Vision
  and Pattern Recognition, pp 2414--2423

\bibitem[{Geusebroek et~al.(2001)Geusebroek, Van~den Boomgaard, Smeulders, and
  Geerts}]{geusebroek2001color}
Geusebroek JM, Van~den Boomgaard R, Smeulders AWM, Geerts H (2001) Color
  invariance. {IEEE} Transactions on Pattern Analysis and Machine Intelligence
  23(12):1338--1350

\bibitem[{Gong et~al.(2012)Gong, Shi, Sha, and Grauman}]{gong2012geodesic}
Gong B, Shi Y, Sha F, Grauman K (2012) Geodesic flow kernel for unsupervised
  domain adaptation. In: Proceedings of the IEEE Conference on Computer Vision
  and Pattern Recognition, IEEE, pp 2066--2073

\bibitem[{Gonzalez-Garcia et~al.(2018)Gonzalez-Garcia, van~de Weijer, and
  Bengio}]{gonzalez2018image}
Gonzalez-Garcia A, van~de Weijer J, Bengio Y (2018) Image-to-image translation
  for cross-domain disentanglement. In: Advances in Neural Information
  Processing Systems, pp 1294--1305

\bibitem[{Goodfellow et~al.(2014)Goodfellow, Pouget-Abadie, Mirza, Xu,
  Warde-Farley, Ozair, Courville, and Bengio}]{goodfellow2014generative}
Goodfellow I, Pouget-Abadie J, Mirza M, Xu B, Warde-Farley D, Ozair S,
  Courville A, Bengio Y (2014) Generative adversarial nets. In: Advances in
  Neural Information Processing Systems, pp 2672--2680

\bibitem[{Gupta et~al.(2016)Gupta, Hoffman, and Malik}]{Gupta_2016_CVPR}
Gupta S, Hoffman J, Malik J (2016) Cross modal distillation for supervision
  transfer. In: Proceedings of the IEEE Conference on Computer Vision and
  Pattern Recognition

\bibitem[{He et~al.(2016)He, Zhang, Ren, and Sun}]{he2016deep}
He K, Zhang X, Ren S, Sun J (2016) Deep residual learning for image
  recognition. In: Proceedings of the IEEE Conference on Computer Vision and
  Pattern Recognition, pp 770--778

\bibitem[{Hoffman et~al.(2016{\natexlab{a}})Hoffman, Gupta, and
  Darrell}]{hoffman2016learning}
Hoffman J, Gupta S, Darrell T (2016{\natexlab{a}}) Learning with side
  information through modality hallucination. In: Proceedings of the IEEE
  Conference on Computer Vision and Pattern Recognition, pp 826--834

\bibitem[{Hoffman et~al.(2016{\natexlab{b}})Hoffman, Gupta, Leong, Guadarrama,
  and Darrell}]{hoffman2016cross}
Hoffman J, Gupta S, Leong J, Guadarrama S, Darrell T (2016{\natexlab{b}})
  Cross-modal adaptation for rgb-d detection. In: Robotics and Automation
  (ICRA), 2016 IEEE International Conference on, IEEE, pp 5032--5039

\bibitem[{Huang et~al.(2018)Huang, Liu, Belongie, and
  Kautz}]{huang2018multimodal}
Huang X, Liu MY, Belongie S, Kautz J (2018) Multimodal unsupervised
  image-to-image translation. In: Proceedings of the European Conference on
  Computer Vision, pp 172--189

\bibitem[{Isola et~al.(2017)Isola, Zhu, Zhou, and Efros}]{isola2016image}
Isola P, Zhu JY, Zhou T, Efros AA (2017) Image-to-image translation with
  conditional adversarial networks. Proceedings of the IEEE Conference on
  Computer Vision and Pattern Recognition

\bibitem[{Jayaraman and Grauman(2014)}]{jayaraman2014zero}
Jayaraman D, Grauman K (2014) Zero-shot recognition with unreliable attributes.
  In: Advances in Neural Information Processing Systems, pp 3464--3472

\bibitem[{Johnson et~al.(2016)Johnson, Schuster, Le, Krikun, Wu, Chen, Thorat,
  Vi{\'e}gas, Wattenberg, Corrado et~al.}]{johnson2016google}
Johnson M, Schuster M, Le QV, Krikun M, Wu Y, Chen Z, Thorat N, Vi{\'e}gas F,
  Wattenberg M, Corrado G, et~al. (2016) Google's multilingual neural machine
  translation system: enabling zero-shot translation. arXiv preprint
  arXiv:161104558

\bibitem[{Kendall et~al.(2018)Kendall, Gal, and Cipolla}]{kendall2017multi}
Kendall A, Gal Y, Cipolla R (2018) Multi-task learning using uncertainty to
  weigh losses for scene geometry and semantics. Proceedings of the IEEE
  Conference on Computer Vision and Pattern Recognition

\bibitem[{Kim et~al.(2016)Kim, Park, Sohn, and Lin}]{kim2016unified}
Kim S, Park K, Sohn K, Lin S (2016) Unified depth prediction and intrinsic
  image decomposition from a single image via joint convolutional neural
  fields. In: Proceedings of the European Conference on Computer Vision,
  Springer, pp 143--159

\bibitem[{Kim et~al.(2017)Kim, Cha, Kim, Lee, and Kim}]{kim2017learning}
Kim T, Cha M, Kim H, Lee J, Kim J (2017) Learning to discover cross-domain
  relations with generative adversarial networks. International Conference on
  Machine Learning

\bibitem[{Kingma and Ba(2014)}]{kingma2014adam}
Kingma D, Ba J (2014) Adam: A method for stochastic optimization. International
  Conference on Learning Representations

\bibitem[{Kuga et~al.(2017)Kuga, Kanezaki, Samejima, Sugano, and
  Matsushita}]{Kuga_2017_ICCV}
Kuga R, Kanezaki A, Samejima M, Sugano Y, Matsushita Y (2017) Multi-task
  learning using multi-modal encoder-decoder networks with shared skip
  connections. In: Proceedings of the International Conference on Computer
  Vision

\bibitem[{Kuznietsov et~al.(2017)Kuznietsov, St{\"u}ckler, and
  Leibe}]{kuznietsov2017semi}
Kuznietsov Y, St{\"u}ckler J, Leibe B (2017) Semi-supervised deep learning for
  monocular depth map prediction. In: Proceedings of the IEEE Conference on
  Computer Vision and Pattern Recognition, pp 6647--6655

\bibitem[{Lai et~al.(2011)Lai, Bo, Ren, and Fox}]{lai2011large}
Lai K, Bo L, Ren X, Fox D (2011) A large-scale hierarchical multi-view rgb-d
  object dataset. In: Proceedings of IEEE International Conference on Robotics
  and Automation, IEEE, pp 1817--1824

\bibitem[{Laina et~al.(2016)Laina, Rupprecht, Belagiannis, Tombari, and
  Navab}]{laina2016deeper}
Laina I, Rupprecht C, Belagiannis V, Tombari F, Navab N (2016) Deeper depth
  prediction with fully convolutional residual networks. In: 3D Vision (3DV),
  2016 Fourth International Conference on, IEEE, pp 239--248

\bibitem[{Lampert et~al.(2014)Lampert, Nickisch, and
  Harmeling}]{lampert2014attribute}
Lampert CH, Nickisch H, Harmeling S (2014) Attribute-based classification for
  zero-shot visual object categorization. {IEEE} Transactions on Pattern
  Analysis and Machine Intelligence 36(3):453--465

\bibitem[{Lee et~al.(2018)Lee, Tseng, Huang, Singh, and Yang}]{lee2018diverse}
Lee HY, Tseng HY, Huang JB, Singh M, Yang MH (2018) Diverse image-to-image
  translation via disentangled representations. In: Proceedings of the European
  Conference on Computer Vision, pp 35--51

\bibitem[{Li et~al.(2018)Li, Liu, Li, Yang, and Kautz}]{li2018closed}
Li Y, Liu MY, Li X, Yang MH, Kautz J (2018) A closed-form solution to
  photorealistic image stylization. In: Proceedings of the European Conference
  on Computer Vision, pp 453--468

\bibitem[{Lin et~al.(2018)Lin, Xia, Qin, Chen, and Liu}]{lin2018conditional}
Lin J, Xia Y, Qin T, Chen Z, Liu TY (2018) Conditional image-to-image
  translation. In: Proceedings of the IEEE Conference on Computer Vision and
  Pattern Recognition, pp 5524--5532

\bibitem[{Liu et~al.(2016)Liu, Shen, Lin, and Reid}]{liu2016learning}
Liu F, Shen C, Lin G, Reid I (2016) Learning depth from single monocular images
  using deep convolutional neural fields. {IEEE} Transactions on Pattern
  Analysis and Machine Intelligence 38(10):2024--2039

\bibitem[{Liu et~al.(2017)Liu, Breuel, and Kautz}]{liu2017unsupervised}
Liu MY, Breuel T, Kautz J (2017) Unsupervised image-to-image translation
  networks. Advances in Neural Information Processing Systems

\bibitem[{Long et~al.(2015)Long, Shelhamer, and Darrell}]{long2015fully}
Long J, Shelhamer E, Darrell T (2015) Fully convolutional networks for semantic
  segmentation. In: Proceedings of the IEEE Conference on Computer Vision and
  Pattern Recognition, pp 3431--3440

\bibitem[{Mao et~al.(2016)Mao, Li, Xie, Lau, and Wang}]{mao2016multi}
Mao X, Li Q, Xie H, Lau RY, Wang Z (2016) Multi-class generative adversarial
  networks with the l2 loss function. arXiv preprint arXiv:161104076

\bibitem[{Mathieu et~al.(2016)Mathieu, Zhao, Zhao, Ramesh, Sprechmann, and
  LeCun}]{mathieu2016disentangling}
Mathieu MF, Zhao JJ, Zhao J, Ramesh A, Sprechmann P, LeCun Y (2016)
  Disentangling factors of variation in deep representation using adversarial
  training. In: Advances in Neural Information Processing Systems, pp
  5040--5048

\bibitem[{McCormac et~al.(2017)McCormac, Handa, Leutenegger, and
  J.Davison}]{McCormac:etal:ICCV2017}
McCormac J, Handa A, Leutenegger S, JDavison A (2017) Scenenet rgb-d: Can 5m
  synthetic images beat generic imagenet pre-training on indoor segmentation?
  Proceedings of the International Conference on Computer Vision

\bibitem[{Mejjati et~al.(2018)Mejjati, Richardt, Tompkin, Cosker, and
  Kim}]{mejjati2018unsupervised}
Mejjati YA, Richardt C, Tompkin J, Cosker D, Kim KI (2018) Unsupervised
  attention-guided image-to-image translation. In: Advances in Neural
  Information Processing Systems, pp 3697--3707

\bibitem[{Mirza and Osindero(2014)}]{mirza2014conditional}
Mirza M, Osindero S (2014) Conditional generative adversarial nets. arXiv
  preprint arXiv:14111784

\bibitem[{Ngiam et~al.(2011)Ngiam, Khosla, Kim, Nam, Lee, and
  Ng}]{ngiam2011multimodal}
Ngiam J, Khosla A, Kim M, Nam J, Lee H, Ng AY (2011) Multimodal deep learning.
  In: International Conference on Machine Learning, pp 689--696

\bibitem[{Nilsback and Zisserman(2008)}]{nilsback2008automated}
Nilsback ME, Zisserman A (2008) Automated flower classification over a large
  number of classes. In: Computer Vision, Graphics \& Image Processing, 2008.
  ICVGIP'08. Sixth Indian Conference on, IEEE, pp 722--729

\bibitem[{Perarnau et~al.(2016)Perarnau, Van De~Weijer, Raducanu, and
  {\'A}lvarez}]{perarnau2016invertible}
Perarnau G, Van De~Weijer J, Raducanu B, {\'A}lvarez JM (2016) Invertible
  conditional gans for image editing. arXiv preprint arXiv:161106355

\bibitem[{Reed et~al.(2016)Reed, Akata, Lee, and Schiele}]{reed2016learning}
Reed S, Akata Z, Lee H, Schiele B (2016) Learning deep representations of
  fine-grained visual descriptions. In: Proceedings of the IEEE Conference on
  Computer Vision and Pattern Recognition, pp 49--58

\bibitem[{Rohrbach et~al.(2011)Rohrbach, Stark, and
  Schiele}]{rohrbach2011evaluating}
Rohrbach M, Stark M, Schiele B (2011) Evaluating knowledge transfer and
  zero-shot learning in a large-scale setting. In: Proceedings of the IEEE
  Conference on Computer Vision and Pattern Recognition, IEEE, pp 1641--1648

\bibitem[{Ronneberger et~al.(2015)Ronneberger, Fischer, and
  Brox}]{ronneberger2015u}
Ronneberger O, Fischer P, Brox T (2015) U-net: Convolutional networks for
  biomedical image segmentation. In: International Conference on Medical image
  computing and computer-assisted intervention, Springer, pp 234--241

\bibitem[{Roy and Todorovic(2016)}]{roy2016monocular}
Roy A, Todorovic S (2016) Monocular depth estimation using neural regression
  forest. In: Proceedings of the IEEE Conference on Computer Vision and Pattern
  Recognition, pp 5506--5514

\bibitem[{Saito et~al.(2017)Saito, Ushiku, and Harada}]{saito2017asymmetric}
Saito K, Ushiku Y, Harada T (2017) Asymmetric tri-training for unsupervised
  domain adaptation. International Conference on Machine Learning

\bibitem[{Silberman et~al.(2012)Silberman, Hoiem, Kohli, and
  Fergus}]{silberman2012indoor}
Silberman N, Hoiem D, Kohli P, Fergus R (2012) Indoor segmentation and support
  inference from {RGBD} images. In: Proceedings of the European Conference on
  Computer Vision, Springer, pp 746--760

\bibitem[{Simonyan and Zisserman(2015)}]{simonyan2014very}
Simonyan K, Zisserman A (2015) Very deep convolutional networks for large-scale
  image recognition. International Conference on Learning Representations

\bibitem[{Song et~al.(2015)Song, Lichtenberg, and Xiao}]{song2015sun}
Song S, Lichtenberg SP, Xiao J (2015) Sun rgb-d: A rgb-d scene understanding
  benchmark suite. In: Proceedings of the IEEE Conference on Computer Vision
  and Pattern Recognition, pp 567--576

\bibitem[{Song et~al.(2017)Song, Herranz, and Jiang}]{song2017depth}
Song X, Herranz L, Jiang S (2017) Depth {CNNs} for {RGB-D} scene recognition:
  learning from scratch better than transferring from rgb-cnns. In: Proceedings
  of the AAAI Conference on Artificial Intelligence

\bibitem[{Taigman et~al.(2017)Taigman, Polyak, and
  Wolf}]{taigman2016unsupervised}
Taigman Y, Polyak A, Wolf L (2017) Unsupervised cross-domain image generation.
  International Conference on Learning Representations

\bibitem[{Tsai et~al.(2018)Tsai, Hung, Schulter, Sohn, Yang, and
  Chandraker}]{tsai2018learning}
Tsai YH, Hung WC, Schulter S, Sohn K, Yang MH, Chandraker M (2018) Learning to
  adapt structured output space for semantic segmentation. Proceedings of the
  IEEE Conference on Computer Vision and Pattern Recognition

\bibitem[{Valada et~al.(2016)Valada, Oliveira, Brox, and
  Burgard}]{valada2016deep}
Valada A, Oliveira GL, Brox T, Burgard W (2016) Deep multispectral semantic
  scene understanding of forested environments using multimodal fusion. In:
  International Symposium on Experimental Robotics, Springer, pp 465--477

\bibitem[{Wang et~al.(2015)Wang, Shen, Lin, Cohen, Price, and
  Yuille}]{wang2015towards}
Wang P, Shen X, Lin Z, Cohen S, Price B, Yuille AL (2015) Towards unified depth
  and semantic prediction from a single image. In: Proceedings of the IEEE
  Conference on Computer Vision and Pattern Recognition, pp 2800--2809

\bibitem[{Wang et~al.(2018{\natexlab{a}})Wang, Liu, Zhu, Tao, Kautz, and
  Catanzaro}]{wang2018high}
Wang TC, Liu MY, Zhu JY, Tao A, Kautz J, Catanzaro B (2018{\natexlab{a}})
  High-resolution image synthesis and semantic manipulation with conditional
  gans. In: Proceedings of the IEEE Conference on Computer Vision and Pattern
  Recognition, pp 8798--8807

\bibitem[{Wang and Neumann(2018)}]{wang2018depth}
Wang W, Neumann U (2018) Depth-aware {CNN} for {RGB-D} segmentation. In:
  Proceedings of the European Conference on Computer Vision, pp 135--150

\bibitem[{Wang et~al.(2018{\natexlab{b}})Wang, van~de Weijer, and
  Herranz}]{wang2018mix}
Wang Y, van~de Weijer J, Herranz L (2018{\natexlab{b}}) Mix and match networks:
  encoder-decoder alignment for zero-pair image translation. In: Proceedings of
  the IEEE Conference on Computer Vision and Pattern Recognition, pp 5467--5476

\bibitem[{Wang et~al.(2019)Wang, Gonzalez-Garcia, van~de Weijer, and
  Herranz}]{wang2019sdit}
Wang Y, Gonzalez-Garcia A, van~de Weijer J, Herranz L (2019) Sdit: Scalable and
  diverse cross-domain image translation. arXiv preprint arXiv:190806881

\bibitem[{Wu et~al.(2019)Wu, Cao, Li, Qian, and Loy}]{Wu_2019_CVPR}
Wu W, Cao K, Li C, Qian C, Loy CC (2019) Transgaga: Geometry-aware unsupervised
  image-to-image translation. In: The IEEE Conference on Computer Vision and
  Pattern Recognition (CVPR)

\bibitem[{Wu et~al.(2018)Wu, Han, Lin, Uzunbas, Goldstein, Lim, and
  Davis}]{wu2018dcan}
Wu Z, Han X, Lin YL, Uzunbas MG, Goldstein T, Lim SN, Davis LS (2018) Dcan:
  Dual channel-wise alignment networks for unsupervised scene adaptation. In:
  Proceedings of the European Conference on Computer Vision

\bibitem[{Xian et~al.(2018{\natexlab{a}})Xian, Lampert, Schiele, and
  Akata}]{xian2018zero}
Xian Y, Lampert CH, Schiele B, Akata Z (2018{\natexlab{a}}) Zero-shot
  learning-a comprehensive evaluation of the good, the bad and the ugly. {IEEE}
  Transactions on Pattern Analysis and Machine Intelligence

\bibitem[{Xian et~al.(2018{\natexlab{b}})Xian, Lorenz, Schiele, and
  Akata}]{xian2018feature}
Xian Y, Lorenz T, Schiele B, Akata Z (2018{\natexlab{b}}) Feature generating
  networks for zero-shot learning. In: Proceedings of the IEEE Conference on
  Computer Vision and Pattern Recognition, pp 5542--5551

\bibitem[{Xu et~al.(2017)Xu, Ouyang, Ricci, Wang, and Sebe}]{xu2017learning}
Xu D, Ouyang W, Ricci E, Wang X, Sebe N (2017) Learning cross-modal deep
  representations for robust pedestrian detection. In: Proceedings of the IEEE
  Conference on Computer Vision and Pattern Recognition, pp 5363--5371

\bibitem[{Yi et~al.(2017)Yi, Zhang, Gong et~al.}]{yi2017dualgan}
Yi Z, Zhang H, Gong PT, et~al. (2017) Dualgan: Unsupervised dual learning for
  image-to-image translation. In: Proceedings of the International Conference
  on Computer Vision

\bibitem[{Yu and Koltun(2016)}]{yu2015multi}
Yu F, Koltun V (2016) Multi-scale context aggregation by dilated convolutions.
  International Conference on Learning Representations

\bibitem[{Yu et~al.(2018)Yu, Zhang, van~de Weijer, Khan, Cheng, and
  Parraga}]{yu2018beyond}
Yu L, Zhang L, van~de Weijer J, Khan FS, Cheng Y, Parraga CA (2018) Beyond
  eleven color names for image understanding. Machine Vision and Applications
  29(2):361--373

\bibitem[{Zhang et~al.(2019)Zhang, Gonzalez-Garcia, van~de Weijer, Danelljan,
  and Khan}]{zhang2019synthetic}
Zhang L, Gonzalez-Garcia A, van~de Weijer J, Danelljan M, Khan FS (2019)
  Synthetic data generation for end-to-end thermal infrared tracking. {IEEE}
  Transactions on Image Processing 28(4):1837--1850

\bibitem[{Zhang et~al.(2016)Zhang, Isola, and Efros}]{zhang2016colorful}
Zhang R, Isola P, Efros AA (2016) Colorful image colorization. In: Proceedings
  of the European Conference on Computer Vision, Springer, pp 649--666

\bibitem[{Zhao et~al.(2017)Zhao, Shi, Qi, Wang, and Jia}]{zhao2017pyramid}
Zhao H, Shi J, Qi X, Wang X, Jia J (2017) Pyramid scene parsing network. In:
  Proceedings of the IEEE Conference on Computer Vision and Pattern
  Recognition, pp 2881--2890

\bibitem[{Zheng et~al.(2017)Zheng, Cheng, and Liu}]{zheng2017maximum}
Zheng H, Cheng Y, Liu Y (2017) Maximum expected likelihood estimation for
  zero-resource neural machine translation. In: Proceedings of the
  International Joint Conference on Artificial Intelligence

\bibitem[{Zhu et~al.(2017{\natexlab{a}})Zhu, Park, Isola, and
  Efros}]{zhu2017unpaired}
Zhu JY, Park T, Isola P, Efros AA (2017{\natexlab{a}}) Unpaired image-to-image
  translation using cycle-consistent adversarial networks. In: Proceedings of
  the International Conference on Computer Vision

\bibitem[{Zhu et~al.(2017{\natexlab{b}})Zhu, Zhang, Pathak, Darrell, Efros,
  Wang, and Shechtman}]{zhu2017toward}
Zhu JY, Zhang R, Pathak D, Darrell T, Efros AA, Wang O, Shechtman E
  (2017{\natexlab{b}}) Toward multimodal image-to-image translation. In:
  Advances in Neural Information Processing Systems, pp 465--476

\bibitem[{Zou et~al.(2018)Zou, Yu, Vijaya~Kumar, and Wang}]{zou2018domain}
Zou Y, Yu Z, Vijaya~Kumar B, Wang J (2018) Unsupervised domain adaptation for
  semantic segmentation via class-balanced self-training. In: Proceedings of
  the European Conference on Computer Vision

\end{thebibliography}

\appendix

\section{Appendix: Network architecture on RGB-D or RGB-D-NIR dataset}
\label{Appendix_rgbd}

Table~\ref{table:encoders} shows the architecture (convolutional and pooling layers) of the encoders used in the cross-modal experiment. Tables~\ref{table:decoders} and \ref{table:rgb_decoder} show the corresponding decoders. Table~\ref{table:rgb_discriminator} shows the discriminator used for RGB. Every convolutional layer of the encoders, decoders and the discriminator is followed by a batch normalization layer and a ReLU layer  (LeakyReLU for the discriminator). The only exception is the RGB encoder, which is initialized with weights from the VGG16 model pretrained on imageNet~\citep{simonyan2014very} and does not use batch normalization. The used abbreviations are shown in Table~\ref{table:Abbreviation name}.

\begin{table}[h]
\setlength{\tabcolsep}{2.5pt}
\centering
\resizebox{\columnwidth}{!}{
\begin{tabular}{cc|cc}
\hline
Layer &Input $\rightarrow $Output    &Kernel, stride\\ 
\hline  
conv1 (RGB)   & [6,256,256,3] $\rightarrow$ [6,256,256,64] & [3,3], 1\\ 
conv1 (Depth)  & $[6,256,256,1] \rightarrow [6,256,256,64]$ & [3,3], 1 \\
conv1 (NIR)  & $[6,256,256,1] \rightarrow [6,256,256,64]$ & [3,3], 1 \\
conv1 (Segm.)  &  [6,256,256,14] $\rightarrow$ [6,256,256,64]& [3,3], 1 \\
conv2     & [6,256,256,64] $\rightarrow$ [6,256,256,64] & [3,3], 1\\ 
pool2 (max) & [6,256,256,64] $\rightarrow$ [6,128,128,64]+indices2& [2,2], 2 \\
\hline
conv3     & [6,128,128,64] $\rightarrow$ [6,128,128,128] & [3,3], 1\\ 
conv4     & [6,128,128,128] $\rightarrow$ [6,128,128,128] & [3,3], 1\\ 
pool4 (max) & [6,128,128,128] $\rightarrow$ [6,64,64,128]+indices4 & [2,2], 2 \\
\hline
conv5    & [6,64,64,128] $\rightarrow$ [6,64,64,256] & [3,3], 1\\ 
conv6     & [6,64,64,256] $\rightarrow$ [6,64,64,256] & [3,3], 1\\ 
conv7     & [6,64,64,256] $\rightarrow$ [6,64,64,256] & [3,3], 1\\ 
pool7 (max) & [6,64,64,256] $\rightarrow$ [6,32,32,256]+indices7& [2,2], 2 \\
\hline
conv8     &  [6,32,32,256] $\rightarrow$ [6,32,32,512] & [3,3], 1\\ 
conv9     & [6,32,32,512] $\rightarrow$ [6,32,32,512] & [3,3], 1\\ 
con10     & [6,32,32,512] $\rightarrow$ [6,32,32,512] & [3,3], 1\\ 
pool10 (max) & [6,32,32,512] $\rightarrow$ [6,16,16,512]+indices10& [2,2], 2 \\
\hline
conv11     & [6,16,16,512] $\rightarrow$ [6,16,16,512] & [3,3], 1\\ 
conv12     & [6,16,16,512] $\rightarrow$ [6,16,16,512] & [3,3], 1\\ 
conv13     & [6,16,16,512] $\rightarrow$ [6,16,16,512] & [3,3], 1\\ 
relu13     & [6,16,16,512] $\rightarrow$ [6,16,16,512] & -, -\\ 
pool13 (max) & [6,16,16,512] $\rightarrow$ [6,8,8,512]+indices13& [2,2], 2 \\
\hline
\end{tabular}
}
\caption{The architecture of the encoder of RGB, depth, NIR and semantic segmentation.}
\label{table:encoders}
\end{table}

\begin{table}[h]
\setlength{\tabcolsep}{2.5pt}
\centering
\resizebox{\columnwidth}{!}{
\begin{tabular}{cc|cc}
\hline
layer  &Input $\rightarrow $Output    &Kernel, stride\\ 
\hline
unpool1      &indices13 + [6,8,8,512] $\rightarrow [6,16,16,512] $ & [2, 2], 2 \\
conv1     & [6,16,16,512] $\rightarrow [6,16,16,512]$ & [3,3], 1\\ 
BN1     & [6,16,16,512] $\rightarrow [6,16,16,512]$ & -, -\\ 
relu1     & [6,16,16,512] $\rightarrow [6,16,16,512]$ & -, -\\
conv2     & [6,16,16,512] $\rightarrow [6,16,16,512]$ & [3,3], 1\\ 
BN2     & [6,16,16,512] $\rightarrow [6,16,16,512]$ & -, -\\ 
relu2     & [6,16,16,512] $\rightarrow [6,16,16,512]$ & -, -\\
conv3     & [6,16,16,512] $\rightarrow [6,16,16,512]$ & [3,3], 1\\
BN3     & [6,16,16,512] $\rightarrow [6,16,16,512]$ & -, -\\ 
relu3     & [6,16,16,512] $\rightarrow [6,16,16,512]$ & -, -\\
\hline
unpool4  &indices10 + [6,16,16,512] $\rightarrow [6,32,32,512]$& [2, 2], 2 \\
conv4     & [6,32,32,512] $\rightarrow [6,32,32,512]$ & [3,3], 1\\ 
BN4     & [6,32,32,512] $\rightarrow [6,32,32,512]$ & -, -\\ 
relu4     & [6,32,32,512] $\rightarrow [6,32,32,512]$ & -, -\\
conv5    & [6,32,32,512] $\rightarrow [6,32,32,512]$ & [3,3], 1\\ BN5     & [6,32,32,512] $\rightarrow [6,32,32,512]$ & -, -\\ 
relu5     & [6,32,32,512] $\rightarrow [6,32,32,512]$ & -, -\\
conv6     & [6,32,32,512] $\rightarrow [6,32,32,256]$ & [3,3], 1\\ 
BN6     & [6,32,32,512] $\rightarrow [6,32,32,512]$ & -, -\\ 
relu6     & [6,32,32,512] $\rightarrow [6,32,32,512]$ & -, -\\
\hline
unpool7  & indices7 + [6,32,32,256] $\rightarrow [6,64,64,256]$& [2, 2], 2 \\
conv7     & [6,64,64,256] $\rightarrow [6,64,64,256]$ & [3,3], 1\\ 
BN7     & [6,64,64,256] $\rightarrow [6,64,64,256]$ & -, -\\ 
relu7     & [6,64,64,256] $\rightarrow [6,64,64,256]$ & -, -\\
conv8     &  [6,64,64,256]$\rightarrow [6,64,64,256]$ & [3,3], 1\\ 
BN8     & [6,64,64,256] $\rightarrow [6,64,64,256]$ & -, -\\ 
relu8     & [6,64,64,256] $\rightarrow [6,64,64,256]$ & -, -\\
conv9    & [6,64,64,256]$\rightarrow [6,64,64,128]$ & [3,3], 1\\ 
BN9     & [6,64,64,256] $\rightarrow [6,64,64,256]$ & -, -\\ 
relu9     & [6,64,64,256] $\rightarrow [6,64,64,256]$ & -, -\\
\hline
unpool10  &indices4 + [6,64,64,128] $\rightarrow [6,128,128,128]$& [2, 2], 2 \\
conv10     & [6,128,128,128] $\rightarrow [6,128,128,128]$ & [3,3], 1\\ 
BN10     & [6,128,128,128] $\rightarrow [6,128,128,128]$ & -, -\\ 
relu10     & [6,128,128,128] $\rightarrow [6,128,128,128]$ & -, -\\
conv11     & [6,128,128,128] $\rightarrow [6,128,128,64]$ & [3,3], 1\\
BN11     & [6,128,128,128] $\rightarrow [6,128,128,128]$ & -, -\\ 
relu11     & [6,128,128,128] $\rightarrow [6,128,128,128]$ & -, -\\
\hline
unpool12  & indices2 + [6,128,128,64] $\rightarrow [6,256,256,64]$& [2, 2], 2 \\
conv12     & [6,256,256,64] $\rightarrow [6,256,256,64]$ & [3,3], 1\\ 
conv13 (Depth)     & [6,256,256,64] $\rightarrow [6,256,256,1]$ & [3,3], 1\\ 
conv13 (NIR)     & [6,256,256,64] $\rightarrow [6,256,256,5]$ & [3,3], 1\\ 
conv13 (Segm.)  & [6,256,256,64] $\rightarrow [6,256,256,14]$ & [3,3], 1\\ 

\hline
\end{tabular}
}
\caption{The architecture of the decoder of depth, NIR and semantic segmentation.}
\label{table:decoders}
\end{table}

\begin{table}[h]
\centering
\setlength{\arrayrulewidth}{0.4\arrayrulewidth}
\resizebox{\columnwidth}{!}{
\begin{tabular}{cc|cc}
\hline
layer  &Input $\rightarrow $Output    &Kernel, stride \\
\hline
conv1    & [6,8,8,512] $\rightarrow [6,16,16,512]$ & [3, 3], 1\\ 
BN1    & [6,16,16,512]$\rightarrow [6,16,16,512]$ & -, -\\
relu1    & [6,16,16,512]$\rightarrow [6,16,16,512]$ & -, -\\
conv2    & [6,16,16,512] $\rightarrow [6,32,32,256]$ & [3, 3], 1\\ 
BN2    &  [6,32,32,256]$\rightarrow  [6,32,32,256]$ & -, -\\
relu2    &  [6,32,32,256]$\rightarrow  [6,32,32,256]$ & -, -\\
conv3    & [6,32,32,256] $\rightarrow [6,64,64,128]$ & [3, 3], 1\\ 
BN3    &  [6,64,64,128]$\rightarrow [6,64,64,128]$ & -, -\\
relu3    & [6,64,64,128]$\rightarrow [6,64,64,128]$ & -, -\\
conv4    & [6,64,64,128] $\rightarrow [6,128,128,64]$ & [3, 3], 1\\ 
BN4    & [6,128,128,64]$\rightarrow [6,128,128,64]$ & -, -\\
relu4    & [6,128,128,64]$\rightarrow [6,128,128,64]$ & -, -\\
conv5    & [6,128,128,64]$ \rightarrow [6,256,256,3]$ & [3, 3], 1\\ 
\hline
\hline
\end{tabular}
}
\caption{The architecture of the decoder of RGB}
\label{table:rgb_decoder}
\end{table}

\begin{table}[h]
\centering
\setlength{\arrayrulewidth}{0.4\arrayrulewidth}%
\resizebox{\columnwidth}{!}{
\begin{tabular}{cc|cc}

\hline
layer  &Input $\rightarrow $Output    &Kernel, stride \\
\hline
deconv1    &$ [6,256,256,3] \rightarrow [6,128,128,64]$ & [5, 5], 2\\ 
lrelu1    &$ [6,128,128,64] \rightarrow [6,128,128,64]$ & -, -\\ 
deconv2     &$ [6,128,128,64] \rightarrow [6,64,64,128]$ & [5, 5], 2\\ 
lrelu2    &$ [6,64,64,128] \rightarrow [6,64,64,128]$ & -, -\\
deconv3     &$ [6,64,64,128] \rightarrow [6,32,32,256]$ & [5,5], 2\\ 
lrelu3    &$ [6,32,32,256] \rightarrow [6,32,32,256]$ & -, -\\
deconv4     &$ [6,32,32,256] \rightarrow [6,16,16,512]$ & [5,5], 2\\ 
\hline
\hline
\end{tabular}
}
\caption{RGB discriminator.}
\label{table:rgb_discriminator}
\end{table}

\section{Appendix: Network architecture on the color dataset and the artworks dataset}
\label{Appendix_color_artworks}
We use several datasets to verify the generality of our method, including  object (Color) and scenes (Artworks).

\textbf{Color dataset}~\citep{yu2018beyond}. We consider the object dataset for color which is collected by~\cite{yu2018beyond}, which includes 11 color labels, each category containing 1000 images. We resize all images to $128 \times 128$. 

\textbf{Artworks}~\citep{zhu2017unpaired}. We also illustrate M$\&$MNet in an artwork setting. This includes real images (\textit{photo}) and four artistic styles (\textit{Monet},  \textit{van Gogh}, \textit{Ukiyo-e} and  \textit{Cezanne}). The the set contains 3000 (photo), 800 (Ukiyo-e), 500 (van Gogh), 600 (Cezanne) and 1200 (Monet) images. All images are resized to $256 \times 256$.

We consider Adam~\citep{kingma2014adam} with a batch size of 4, using a learning rate of 0.0002. The network is initialized using a Gaussian distribution with zero mean and a standard deviation of 0.5.  We only use adversarial loss to train our model.

Tables~\ref{table:encoders of content enconder}-\ref{table:the image discriminator} show the architectures of the encoder, image decoder and discriminator used in the cross-modal experiment. The following tables only show the image size of $128 \times 128$, while for artworks dataset it is same architecture except for image resolution. The used abbreviations are shown in Table~\ref{table:Abbreviation name}.
\begin{table}[h]
\setlength{\arrayrulewidth}{\arrayrulewidth}
\centering
\resizebox{\columnwidth}{!}{
\begin{tabular}{cc|ccc}

\hline
Layer &Input $\rightarrow $Output    &Kernel, stride, pad\\ 
\hline  
conv1   & [4,128, 128,3] $\rightarrow$ [4,128, 128, 64] & [7,7], 1, 3\\ 
IN1   & [4,128, 128, 64] $\rightarrow$ [4,128, 128, 64] &  -, -, -\\ 
pool1 (max) & [4,128, 128, 64] $\rightarrow$[4,64, 64, 64]+indices1 & [2,2], 2, - \\
\hline
conv2   & [4,64, 64,64] $\rightarrow$ [4,64, 64,128] & [7,7], 1, 3\\ 
IN2   & [4,64, 64,128] $\rightarrow$ [4,64, 64,128] &  -, -, -\\ 
pool2 (max) & [4,64, 64,128] $\rightarrow$[4,32, 32,128]+indices2 & [2,2], 2, - \\
\hline
conv3   & [4,32, 32,128] $\rightarrow$ [4,32, 32,256] & [7,7], 1, 3\\ 
IN3   &  [4,32, 32,256] $\rightarrow$ [4,32, 32,256] &  -, -, -\\ 
pool3 (max) &  [4,32, 32,256] $\rightarrow$ [4,16, 16,256]+indices3 & [2,2], 2, - \\
\hline
RB(IN)4-9   & [4,16, 16,256] $\rightarrow$ [4,16, 16,256] & [7,7], 1, 3\\ 

\hline
\end{tabular}
}
\caption{The architecture of the encoder for $128\times128$ input.}
\label{table:encoders of content enconder}
\end{table}


\begin{table}[h]
\setlength{\arrayrulewidth}{1\arrayrulewidth}
\centering
\resizebox{\columnwidth}{!}{
\begin{tabular}{cc|ccc}

\hline
Layer &Input $\rightarrow $Output    &Kernel, stride, pad\\ 
\hline  
RB(IN)1-6    &  [4,16, 16,256] $\rightarrow$ [4,16, 16,256] & [7,7], 1, 3\\ 

\hline
unpool1   & indices3 + [4,16, 16,256] $\rightarrow$ [4,32, 32,256] & [2, 2], 2, -\\ 
conv1   & [4,32, 32,256] $\rightarrow$ [4,32, 32,128] & [7,7], 1, 3\\ 
IN1   & [4,32, 32,128] $\rightarrow$ [4,32, 32,128] &  -, -, -\\ 
\hline
unpool2   & indices2 + [4,32, 32,128] $\rightarrow$ [4, 64, 64,128] & [2, 2], 2, -\\ 
conv2   & [4, 64, 64,128] $\rightarrow$  [4, 64, 64,64] & [7,7], 1, 3\\ 
IN2   & [4, 64, 64,64]$\rightarrow$   [4, 64, 64,64] &  -, -, -\\ 
\hline
unpool3   & indices1 + [4, 64, 64,64] $\rightarrow$ [4, 128, 128,64] & [2, 2], 2, -\\ 
conv3   & [4, 128, 128,64] $\rightarrow$  [4, 128, 128,3] & [7,7], 1, 3\\ 
\hline
\end{tabular}
}
\caption{The architecture of the decoder for $128\times128$ output.}
\label{table:the image generator}
\end{table}

\begin{table}[h]
\centering
\setlength{\arrayrulewidth}{0.9\arrayrulewidth}
\resizebox{\columnwidth}{!}{
\begin{tabular}{cc|ccc}

\hline
Layer &Input $\rightarrow $Output    &Kernel, stride, pad\\ 
\hline  
conv1    &  [4,128, 128,3] $\rightarrow$ [4,64, 64,64] & [4,4], 2, 1\\ 
lrelu1    &  [4,64, 64,64] $\rightarrow$ [4,64, 64,64] & -, -, -\\
\hline
conv2    &  [4,64, 64,64] $\rightarrow$ [4,32, 32,128] & [4,4], 2, 1\\ 
lrelu2    &  [4,32, 32,128] $\rightarrow$ [4,32, 32,128] & -, -, -\\
\hline
conv3    &   [4,32, 32,128]  $\rightarrow$   [4,16, 16,256]  & [4,4], 2, 1\\ 
lrelu3    &  [4,16, 16,256] $\rightarrow$ [4,16, 16,256] & -, -, -\\
\hline
conv4    &   [4,16, 16,256]  $\rightarrow$   [4,8, 8,512]  & [4,4], 2, 1\\ 
lrelu4    &[4,8, 8,512]  $\rightarrow$[4,8, 8,512]  & -, -, -\\
\hline
conv5    &[4,8, 8,512]  $\rightarrow$[4,8, 8,1]  & [1,1], 1, 0\\ 
\hline

\end{tabular}
}
\caption{Architecture for the discriminator Loss specification for $128\times128$ input.}
\label{table:the image discriminator}
\end{table}

\begin{table}[h]
\centering
\resizebox{\columnwidth}{!}{
\begin{tabular}{cc}
\hline
Abbreviation & Name   \\ 

\hline  
pool    &  pooling layer \\ 
\hline
unpool    &  unpooling layer \\ 
\hline
lrelu    &  leaky relu layer \\ 
\hline
conv    &  convolutional layer\\ 
\hline 
linear    &  fully connection layer \\ 
\hline
BN    &   batch normalization layer\\ 
\hline
IN    &  instance normalization layer\\ 
\hline  
RB(IN)    &  residual block layer using instance normalization \\ 
\hline

\end{tabular}
}
        \caption{Abbreviations used in other tables.}
\label{table:Abbreviation name}

\end{table}

\section{Appendix: Network architecture for the Flower dataset}
\label{Appendix_flower}

\textbf{Flower dataset}~\citep{nilsback2008automated}. The Flower dataset consists of 102 categories. We consider 10 categories(\textit{passionflower}, \textit{petunia}, \textit{rose}, \textit{wallflower}, \textit{watercress}, \textit{waterlily}, \textit{cyclamen}, \textit{foxglove}, \textit{frangipani}, \textit{hibiscus}). Each category includes between 100 and 258 images. we resize the image to $128 \times 128$. 

Similarly, we optimize our model by means of using Adam~\citep{kingma2014adam}, the batch size of 4 and a learning rate of 0.0002. We initialize  hyperparameters using a Gaussian distribution with zero mean and a standard deviation of 0.5.  We  use adversarial loss  and $L2$ to train $ \Theta _3$, and only $L2$ for $ \Theta _1$ and $ \Theta _2$.

Tables~\ref{table:encoder_flower_domain1_2} and \ref{table:decoder_flower} detail the architecture of the encoder and decoder, respectively, of the two single channel modalities $ \Theta _1$ and $ \Theta _2$. The encoder and decoder for the third modality $ \Theta _3$ are analogous, just adapted to three input and output channels, respectively. For $ \Theta _3$ we also use the discriminator detailed in Table~\ref{table:the image discriminator}.

\begin{table}[h]
\setlength{\arrayrulewidth}{\arrayrulewidth}
\centering
\resizebox{\columnwidth}{!}{
\begin{tabular}{cc|ccc}
\hline
Layer &Input $\rightarrow $Output    &Kernel, stride, pad\\ 
\hline  
conv1   & [4,128, 128,1] $\rightarrow$ [4,128, 128, 64] & [7,7], 1, 3\\ 
IN1   & [4,128, 128, 64] $\rightarrow$ [4,128, 128, 64] &  -, -, -\\ 
pool1 (max) & [4,128, 128, 64] $\rightarrow$[4,64, 64, 64]+indices1 & [2,2], 2, - \\
\hline
conv2   & [4,64, 64,64] $\rightarrow$ [4,64, 64,128] & [7,7], 1, 3\\ 
IN2   & [4,64, 64,128] $\rightarrow$ [4,64, 64,128] &  -, -, -\\ 
pool2 (max) & [4,64, 64,128] $\rightarrow$[4,32, 32,128]+indices2 & [2,2], 2, - \\
\hline
conv3   & [4,32, 32,128] $\rightarrow$ [4,32, 32,256] & [7,7], 1, 3\\ 
IN3   &  [4,32, 32,256] $\rightarrow$ [4,32, 32,256] &  -, -, -\\ 
pool3 (max) &  [4,32, 32,256] $\rightarrow$ [4,16, 16,256]+indices3 & [2,2], 2, - \\
\hline
RB(IN)4-9   & [4,16, 16,256] $\rightarrow$ [4,16, 16,256] & [7,7], 1, 3\\ 

\hline
\end{tabular}
}
\caption{The architecture of the encoder of $\Theta _1$ and $ \Theta _2$.}
\label{table:encoder_flower_domain1_2}
\end{table}


\begin{table}[h]
\setlength{\arrayrulewidth}{1\arrayrulewidth}
\centering
\resizebox{\columnwidth}{!}{
\begin{tabular}{cc|ccc}

\hline
Layer &Input $\rightarrow $Output    &Kernel, stride, pad\\ 
\hline  
RB(IN)1-6    &  [4,16, 16,256] $\rightarrow$ [4,16, 16,256] & [7,7], 1, 3\\ 

\hline
unpool1   & indices3 + [4,16, 16,256] $\rightarrow$ [4,32, 32,256] & [2, 2], 2, -\\ 
conv1   & [4,32, 32,256] $\rightarrow$ [4,32, 32,128] & [7,7], 1, 3\\ 
IN1   & [4,32, 32,128] $\rightarrow$ [4,32, 32,128] &  -, -, -\\ 
\hline
unpool2   & indices2 + [4,32, 32,128] $\rightarrow$ [4, 64, 64,128] & [2, 2], 2, -\\ 
conv2   & [4, 64, 64,128] $\rightarrow$  [4, 64, 64,64] & [7,7], 1, 3\\ 
IN2   & [4, 64, 64,64]$\rightarrow$   [4, 64, 64,64] &  -, -, -\\ 
\hline
unpool3   & indices1 + [4, 64, 64,64] $\rightarrow$ [4, 128, 128,64] & [2, 2], 2, -\\ 
conv3   & [4, 128, 128,64] $\rightarrow$  [4, 128, 128,1] & [7,7], 1, 3\\ 
\hline
\end{tabular}
}
\caption{The architecture of the decoder for $\Theta _1$ and $ \Theta _2$.}
\label{table:decoder_flower}
\end{table}

\end{document}